\documentclass[mnsc]{informs3}

\renewcommand{\theARTICLETOP}{}
\OneAndAHalfSpacedXII

\usepackage{endnotes}
\let\footnote=\endnote
\let\enotesize=\normalsize
\def\notesname{Endnotes}%
\def\makeenmark{\hbox to1.275em{\theenmark.\enskip\hss}}
\def\enoteformat{\rightskip0pt\leftskip0pt\parindent=1.275em
\leavevmode\llap{\makeenmark}}

\usepackage{natbib}
\bibpunct[, ]{(}{)}{,}{a}{}{,}%
\def\bibfont{\small}%

\newcommand{\qedwhite}{\hfill \ensuremath{\Box}}
\usepackage{subfigure}
\usepackage{custom_informs}
\usepackage{booktabs}
\usepackage{multirow}

\usepackage{algorithm}
\usepackage{algorithmic}

\usepackage{afterpage}

\usepackage{enumitem}
\usepackage{bbm}

\usepackage{colortbl}
\newcommand{\graycell}{\cellcolor{black!15}}

\usepackage{tikz}
\usetikzlibrary{shapes,arrows,shadows,positioning,decorations.pathreplacing,trees,calc,patterns, fit}
\usepackage{amsfonts}
\usepackage{amsmath}
\usepackage{rotating}
\usepackage{amssymb}

\tikzstyle{arrow} = [line width=0.75mm,->,>=stealth]

\tikzstyle{data} = [rectangle, rounded corners, minimum width=1.5cm, minimum height=1.5cm,text width=1.5cm,text centered, draw=black,]
\tikzstyle{dataVert} = [rectangle, rounded corners, minimum width=1.5cm, minimum height=7cm,text width=1.5cm,text centered, draw=black, fill=red!20!white]
\tikzstyle{noBox} = [rectangle, rounded corners, minimum width=2cm, minimum height=1.5cm] 
\tikzstyle{noBoxWide} = [rectangle, rounded corners, minimum width=2cm, minimum height=1.5cm,text width=2cm, text centered]
\tikzstyle{process} = [rectangle, ultra thick, rounded corners, minimum width=2.25cm, minimum height=1.cm,  draw=black,fill=black!15!white]
\tikzstyle{output} = [rectangle, rounded corners, minimum width=1cm, minimum height=.5cm,text width=1cm,text centered, draw=black, fill=red!40!white]
\tikzstyle{IOBox} = [rectangle, rounded corners, minimum width=2cm, minimum height=0.5cm,text width=1cm,text centered, draw=black]

\tikzstyle{block} = [rectangle, draw, fill=green!20, thick, text width=2em,align=center, minimum height=2.5em]
\tikzstyle{bigblock} = [rectangle, draw, fill=green!20, thick, text width=6em,align=center, minimum height=4em]

\tikzstyle{datapoint} = [circle, fill, inner sep=1.5]
\tikzstyle{datasample} = [rectangle, draw, ultra thick, inner sep=3]
\tikzstyle{datasampleDiamond} = [diamond, draw, ultra thick, inner sep=2.6]
\tikzstyle{datasampleBd} = [circle, draw, ultra thick, inner sep=2.6]

\makeatletter
\newcommand{\vast}{\bBigg@{4}}
\newcommand{\Vast}{\bBigg@{5}}
\makeatother

\newcommand{\pfeas}{\field{P}_{(\bx,\bu)}}
\newcommand{\pinfeas}{\bar{\field{P}}_{(\bx,\bu)}}
\newcommand{\pparam}{\field{P}_{\bu}}

\newcommand{\EX}{\mathbb{E}}
\newcommand{\neighbourhood}[1]{\set{N}_\delta \left(#1\right)}
\newcommand{\neighbourhoodsub}[2]{\set{N}_{#2} \left(#1\right)}
\newcommand{\Xfeas}{\set{X}}

\newcommand{\Xrel}{\set{P}}

\newcommand{\Feasfunc}{\Psi(\bx, \bu)}

\newcommand{\OP}{\mathbf{OP}}
\newcommand{\BP}{\mathbf{BP}}
\newcommand{\GBP}{\mathbf{GBP}}
\newcommand{\FCP}{\mathbf{FCP}}
\newcommand{\SFCP}{\mathbf{S}\text{--}\mathbf{FCP}}

\newcommand{\dataset}{\set{D}}
\newcommand{\datasetinfeas}{\bar{\set{D}}}
\newcommand{\datasetparam}{\hat{\set{U}}}

\newcommand{\Niter}{J}
\newcommand{\Nx}{{N_{\bx}}}
\newcommand{\Nbx}{{\bar{N}_{\bx}}}
\newcommand{\Nu}{{N_{\bu}}}

\newcommand{\barr}{B_\delta}
\newcommand{\btest}{B^\mathrm{test}}
\newcommand{\bxl}{\bx^{\lambda}}
\newcommand{\bxlj}{\bx^{\lambda_j}}
\newcommand{\bxrel}{\bx^{\set{P}}}

\newcommand{\lambdau}{\tlambda}
\newcommand{\lambdal}{\tlambda'}

\newcommand{\Rad}{\mathfrak{R}_\Nu}

\newcommand{\hRad}{\hat{\mathfrak{R}}_\Nu}

\newcommand{\Proj}{\mathrm{proj}}
\newcommand{\card}{\mathrm{card}}

\newcommand{\RT}{\mathbf{RTP}}
\newcommand{\PP}{\mathbf{PPP}}

\newcommand{\softmax}{\mathrm{softmax}}
\newcommand{\convblk}{\texttt{conv3d}}
\newcommand{\deconvblk}{\texttt{deconv3d}}
\newcommand{\batchnorm}{\texttt{BN}}
\newcommand{\leakyrelu}{\texttt{LR}}
\newcommand{\dropout}{\texttt{D}}
\newcommand{\bnlr}{\batchnorm-\leakyrelu}
\newcommand{\bnr}{\batchnorm-\texttt{R}}
\newcommand{\bndr}{\batchnorm-\dropout-\texttt{R}}
\newcommand{\avgpool}{\texttt{AP}}
\newcommand{\ReLU}{\mathrm{ReLU}}

\newif\ifproofinbody
\proofinbodytrue
\newif\ifproofinappendix
\proofinappendixfalse

\TheoremsNumberedThrough     
\ECRepeatTheorems

\EquationsNumberedThrough    

\begin{document}


\RUNAUTHOR{Babier et al.}

\RUNTITLE{
    Learning to Optimize Contextually Constrained Problems
}

\TITLE{
    Learning to Optimize Contextually Constrained Problems for Real-Time Decision-Generation
}

\ARTICLEAUTHORS{%
        \AUTHOR{Aaron Babier$^1$, Timothy C. Y. Chan$^1$, Adam Diamant$^2$, Rafid Mahmood$^1$}
        \AFF{
        $^1$Mechanical \& Industrial Engineering, University of Toronto, Toronto, Canada, \EMAIL{\{ababier, tcychan, rmahmood\}@mie.utoronto.ca} \\
        $^2$Schulich School of Business, York University, Toronto, Canada, \EMAIL{adiamant@schulich.yorku.ca}
        }
} 

\ABSTRACT{%
        The topic of learning to solve optimization problems has received interest from both the operations research and machine learning communities. In this work, we combine techniques from both fields to address the problem of learning to generate decisions to instances of continuous optimization problems where the feasible set varies with contextual features. We propose a novel framework for training a generative model to estimate optimal decisions by combining interior point methods and adversarial learning, which we further embed within a data generation algorithm. Decisions generated by our model satisfy in-sample and out-of-sample optimality guarantees. Finally, we investigate case studies in portfolio optimization and personalized treatment design, demonstrating that our approach yields advantages over predict-then-optimize and supervised deep learning techniques, respectively. 
}%



\maketitle

%


\section{Introduction}

Advances in machine learning have led to a growing interest in solving contextual optimization problems whose objective and/or constraints depend on input context features that vary from instance to instance. For example, consider a medical decision making application where the context is a patient's medical history, current diagnosis, and the results of recent clinical tests, all of which influence the design of a personalized treatment. In this paper, we study contextual optimization problems characterized by three distinct attributes: (i) a large universe of contexts where the relationship between each context and the feasible set requires estimation;
(ii) a set of complicating constraints that are controlled by the contexts and make the problem challenging to solve; and (iii) decision-makers or end-users that require high-quality solutions to be generated quickly for incoming contexts. 
Below, we provide two concrete examples.

\textbf{Portfolio Optimization.} 
Automated brokers are increasingly used to recommend personalized investment portfolios over web or mobile interfaces~\citep{capponi:2021personalized}. (i) A user of the service possesses unique demographic characteristics and risk tolerances that affect the types of portfolios that can be recommended~\citep{bansal:2019special, li:2016inverse}. 
(ii) The feasible set for each user's optimization problem may be complex, e.g., containing non-convex cardinality constraints to restrict the number and type of investments that can be held~\citep{kim:2019gan}. (iii) Recommended investment portfolios must be of certifiable quality~\citep{ban:2018machine} since users receiving unsuitable recommendations may churn. Finally, to be deemed user-responsive, modern web services require low latencies when producing recommendations. For instance, Google estimates each 500 ms delay corresponds to a $20\%$ traffic loss for their mobile applications~\citep{wang:2012far}.

\textbf{Personalized Cancer Treatment.} 
Radiation therapy is a widely used cancer treatment modality where treatment quality is typically evaluated on whether it meets target thresholds for a set of clinical performance metrics or criteria~\citep{delaney:2005role}. 
Since not all criteria can be satisfied simultaneously in general, oncologists aim to design treatments that satisfy a personalized subset of them. 
(i) Each patient's clinical history and medical images are used as context to determine which criteria should be used as constraints in treatment design~\citep{Kearney:2018aa, Babier:2019aa}. 
(ii) The most commonly used clinical criteria are akin to Value-at-Risk and thus, induce non-convex constraints. 
(iii) High-quality treatments are essential in achieving good clinical outcomes. Additionally, treatment design needs to be efficient to minimize the delay until treatment initiation, since poor initial designs undergo time-consuming re-planning iterations.

In contextual optimization problems, contexts are often integrated following the `\emph{predict-then-optimize}' paradigm, which uses machine learning to map a context to an input parameter of an optimization model (see~\citealt{chan:2012optimizing, angalakudati:2014business, ferreira:2015analytics} for examples and~\citealt{mivsic:2020data} for a modern survey). However, predict-then-optimize frameworks may be challenging to solve due to the presence of parametric complicating constraints that inhibit efficient solution methods~\citep{campbell:2004efficient, conejo:2006decomposition, boyd2007notes}. Given the large universe of contexts, solving such optimization problems on-demand can be slow and sensitive to poor estimates of the constraint parameters~\citep{bertsimas:2011theory}. An alternative is to use deep neural networks that learn from context-solution pairs to directly generate solutions to contextual optimization problems without having to solve mathematical programs (see~\citealt{bengio:2018machine} and \citealt{mazyavkina2020reinforcement} for recent surveys). Such approaches benefit from fast decision generation and the ability to estimate complex relationships between contexts and the feasible sets of the corresponding optimization problems, but they also require large data sets for training and often do not provide theoretical guarantees on solution quality~\citep{kotary:2021learning}.

\begin{figure}[t]
    \centering
        \includegraphics[width=0.9\linewidth]{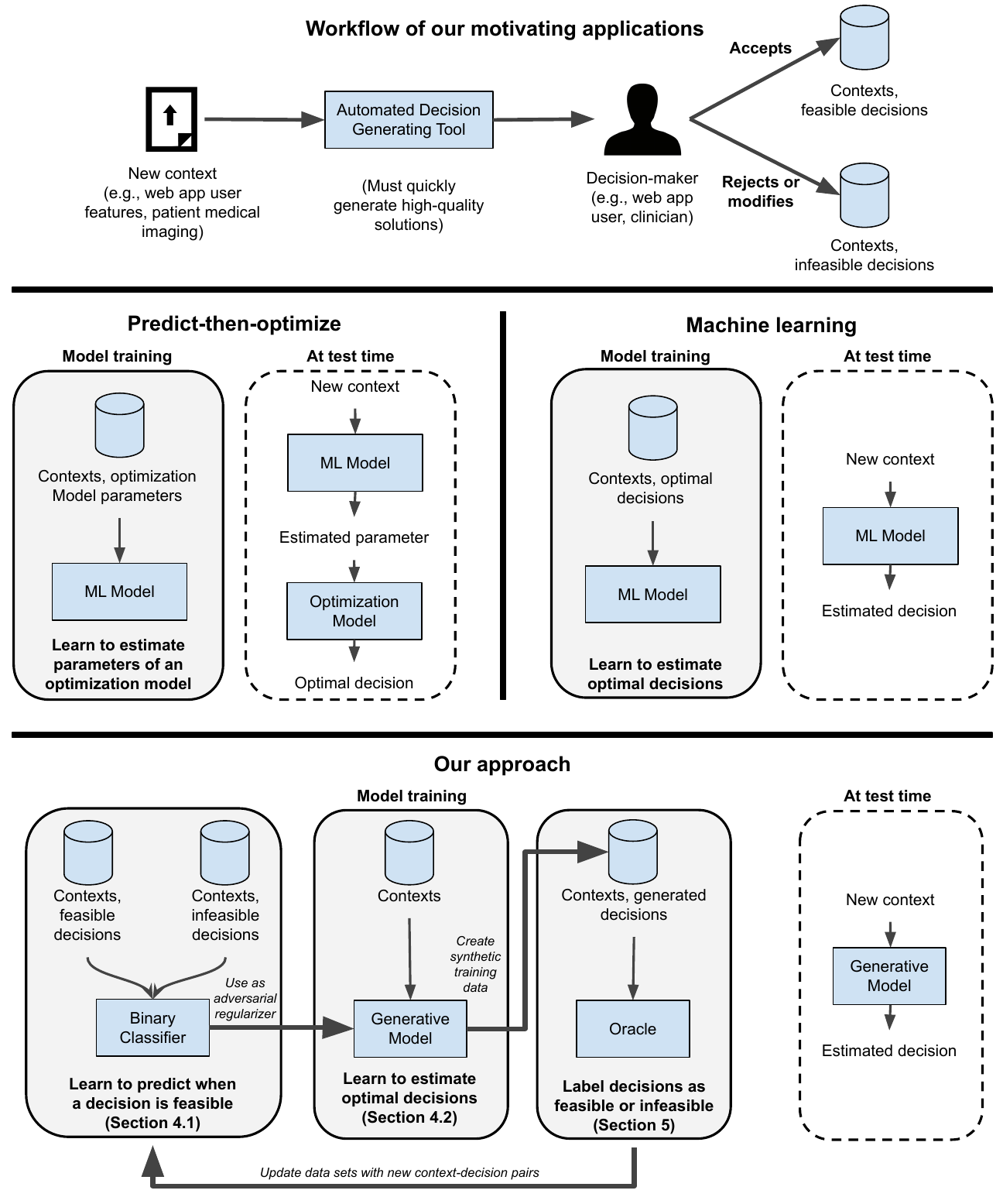}
    \caption{
    Top: In applications that require quickly generating high-quality contextual solutions, an automated tool may produce erroneous decisions, which decision makers review and reject if infeasible. We collect this data for training.
    Left: In predict-then-optimize methods, the decision generating tool is an optimization model, which may approximate the problem in order to solve quickly.
    Right: In machine learning methods, the tool is a predictive model trained to estimate optimal decisions.
    Bottom: Our final tool is a generative model, trained iteratively using a binary classifier and feasibility oracle.  
    }
    \label{fig:flowchart}
\end{figure}

In this paper, we develop an alternative learning-based methodology to solve contextual optimization problems with a linear objective and context-dependent complicating constraints. 
To accommodate complex relationships between the large number of contexts and their corresponding constraint sets, as well as the need to produce decisions in near real-time, we employ the machine learning paradigm of generating solutions directly from contexts. Our approach consists of:
\begin{itemize}
    \item A binary classification model (\emph{classifier}) that, given an input decision and context vector, predicts whether that decision is feasible or infeasible.
    
    \item A generative model (\emph{generator}) that outputs optimal decisions given an input context that is trained using the classifier as a regularizer to penalize the generation of infeasible decisions.

    \item An active learning loop that uses the outputs of the generative model, which are labeled via a feasibility oracle, to iteratively refine the classifier and generator during training.
\end{itemize}
Figure~\ref{fig:flowchart} describes our motivating applications, the existing predict-then-optimize and machine learning paradigms, as well as our proposed learning algorithm.

In our approach, we first train a classifier using data on infeasible and feasible solutions to learn a mapping between a contextual input and the feasible region of the corresponding optimization problem. We then use this trained classifier as a regularizer when training a generative model to learn to generate optimal decisions when given a context vector as input. Because learning high-dimensional feasible sets may require large amounts of data, we augment our learning approach by actively generating synthetic decision data using the contexts available in our training set. This synthetic training data is evaluated and labeled as either feasible or infeasible using an ``oracle of feasibility", which can be a look-up-table that uses explicit information on the constraint structure and past feasible decisions as a reference, human decision-makers in-the-loop that manually assess the outputs of the model during training, or a model-based algorithmic labeler. Thus, the classifier and generator are embedded in an active learning loop that ensures the generator learns to produce progressively better solutions for all contexts. 
After training, our framework will have learned a data-driven representation of the feasible region and how it varies with the contextual input; furthermore, the generator will then be able to quickly generate optimal decisions.

The learning algorithm used to train the generator is analogous to those used for Interior Point Methods (IPMs) in that the classifier acts as a barrier function that penalizes infeasibility. Because the classifier is data-driven, it may not be a perfect representation of the feasible region, which means that classical results on the quality of IPM-generated solutions do not apply. Consequently, we first derive new results in IPM theory by considering barrier functions that are defined over a relaxation of a feasible set; we call these barrier functions $\delta$-barriers, where $\delta$ refers to the approximate size of this relaxed feasible set. We then develop a new $(\delta, \epsilon)$-optimality guarantee for optimization problems that employ $\delta$-barriers and generalize key properties of IPM algorithms to this setting. Because the classifier is used as a substitute for the $\delta$-barrier function when training the generator, we prove that the generator inherits in-sample guarantees on prediction accuracy associated with this $(\delta, \epsilon)$-optimality property and provide a generalization bound on the out-of-sample $(\delta,\epsilon)$-optimality gap of our framework. Finally, we show that our active learning-based algorithm (which we call Interior Point Methods with Adversarial Networks, or IPMAN) produces progressively tighter $(\delta,\epsilon)$-optimal solutions by iteratively growing the training data set.

To illustrate the effectiveness of IPMAN, we apply this method to two case studies representing the examples above. In the portfolio optimization application, we collect asset data from multiple exchanges and create a synthetic data set of the demographic characteristics, preferences, and feasible/infeasible decisions made by retail investors. In practice, this set would be collected by deploying a baseline model for a trial group of investors and then observing whether they use the baseline recommendations or reject them. The recorded preferences are also used to create an oracle that labels the decisions generated during training within the active learning loop. We compare IPMAN to an idealized predict-then-optimize baseline, where we predict investor preferences with perfect accuracy and then optimize the portfolio. Our experiments show that IPMAN generates portfolio recommendations with competitive objective function values in an order-of-magnitude less time than a commercial optimization solver. Further, we show that IPMAN is more robust to parameter estimation error compared to the predict-then-optimize baseline.

In the personalized treatment design application, real imaging data and dose distributions are obtained for 217 patients with head-and-neck cancer. 
By cross-referencing which clinical targets were met by delivered treatments versus those that our models generate during training, we can define a look-up-table-based oracle to evaluate the quality of the generated treatment plans. We compare against two state-of-the-art neural network dose prediction models as baselines and show that IPMAN learns to generate radiation therapy doses that better satisfy clinical constraints as compared to the baseline. We also show that IPMAN can adapt to out-of-sample settings without data by demonstrating that it continues to perform well if the clinical criteria used to evaluate treatment quality change. This analysis models the situation of domain shift where IPMAN is trained on one clinic's data but applied to patients from a new clinic.

The paper is organized as follows. In Section 3, we present an IPM-based theory for relaxed barrier functions which we call $\delta$-barriers. In Section 4, we develop our classification and generative models, and establish bounds on the quality of in-sample solutions produced by our generator when using a classifier-based $\delta$-barrier as a regularizer. Section 5 describes the active learning framework and how feasible and infeasible data are synthesized to iteratively improve the performance of the classifier and generator. We prove a generalization bound on out-of-sample performance in Section 6 and numerically validate IPMAN in Section 7 on the two examples discussed above. We conclude in Section 8 by discussing the nuances of our approach and propose several potential extensions.

\subsection{Relationship to Prior Work}
In the optimization literature, complicating constraints are typically addressed via decomposition techniques (e.g., Langrangian relaxation, Benders algorithms). Given the generality of the effect that a context may have on the formulation of an optimization problem, there is no guarantee that an optimal solution can be obtained in real-time.
Alternatively in our data-driven framework, the feasible set is learned and provably optimal solutions are generated quickly. Active learning iteratively creates new data points which allows the algorithm to explore the feasible region~\citep{settles:2009active}. 
Thus, our approach relaxes the mathematical representation of the constraint set and replaces it with a data-driven model trained on past feasible and infeasible decisions.

Our framework uses two predictive models, one to evaluate and one to generate decisions; this is common in reinforcement learning (e.g., actor-critic methods,~\citealp{konda:2000actor}) and deep learning (e.g., generative adversarial networks,~\citealp{goodfellow:2014generative}). 
The objective used to train the generative model is derived from IPMs~\citep{nemirovski:1994interior} and our learning guarantees extend previous Rademacher complexity results~\citep{bartlett:2002rademacher} for data-driven optimization. Finally, the design of a model to generate decisions is related to the estimation of distribution algorithms commonly used in the black-box optimization literature~\citep{pelikan:2002survey}. 
We summarize the most relevant related work below.

\textbf{Interior Point Methods.~}
These algorithms are among the most popular techniques for solving constrained optimization problems~\citep{nemirovski:1994interior}. 
IPMs have been primarily applied to linear and quadratic optimization problems where they quickly converge to optimal solutions~\citep{gondzio:2012interior}. Recent results on barriers for arbitrary convex sets have renewed interest in IPMs for general convex optimization~\citep{badenbroek:2018aa, bubeck:2014entropic}. IPMs also exist for non-convex problems~\citep{vanderbei:1999interior, benson:2004interior, hinder:2018one}. These papers all assume access to explicit constraints or a barrier that is well-defined for the entire feasible set. Here, we construct a barrier function (i.e., our classifier) that approximates a relaxation of the feasible set and develop an IPM theory for this setting.

\textbf{Contextual Optimization.~}
The most common approach is the `predict-then-optimize' paradigm, i.e., to construct a parametric optimization model of the decision-making problem and use machine learning to predict parameters from context-dependent inputs~\citep{angalakudati:2014business, ferreira:2015analytics, elmachtoub:2017smart}. 
A non-parametric alternative is to directly estimate the effect of the context in terms of a conditional stochastic optimization model. For example, a stochastic objective that is conditioned on the context may be characterized by embedding a machine learning model to estimate probability weights on the objective~\citep{kao:2009directed, hannah:2010nonparametric, ban:2018big, bertsimas:2014predictive, bertsimas:2018optimization}.

Our work is most similar to~\citet{ban:2018big} who use Empirical Risk Minimization (ERM) to construct a predictor for the optimal solution to a Newsvendor problem. 
Because their optimization model is only constrained by the non-negativity of the decision variables, we generalize their result to an arbitrary set of restrictions by incorporating constraint satisfaction via a binary classifier.~\citet{bertsimas:2014predictive} remark on the challenges of constraint satisfaction when using ERM, and instead, propose a weighted learning framework that estimates the weights (i.e., conditional probability terms) in a sample-average optimization problem. They also prove several generalization results that arise from ERM theory. We extend their generalization bounds to out-of-sample $\epsilon$-optimality guarantees, and in particular, to problems where the feasible set depends on the contexts via a relationship that must be estimated.

\textbf{Deep Learning for Constrained Optimization.~}
Recent advances have spurred interest in using neural networks to solve constrained optimization problems~\citep{bengio:2018machine}. 
These task-specific models are trained via customized learning algorithms and include supervised learning with a data set of contexts and optimal solutions~\citep{vinyals:2015pointer, Larsen:2018aa}, reinforcement learning~\citep{bello:2017neural, kool:2018attention}, and task-based learning~\citep{donti:2017task}.

Generating a solution via the function call of a neural network is typically faster than solving an optimization problem~\citep[see][for numerical studies]{nair:2018, kool:2018attention}. However, a trained model may not guarantee that predicted solutions can satisfy every constraint. There are several potential approaches that address this issue. For instance, training a supervised learning technique on a large data set may help make predictions more accurate~\citep{Larsen:2018aa}. An increasingly popular approach is to use a customized class of neural network layers which can directly learn to satisfy first-order KKT conditions for solving a convex problem~\citep{amos:2017optnet, agrawal:2019differentiable}. 
Finally, the loss function may be customized to encourage constraint satisfaction~\citep{donti:2017task}. 
We train a prediction model using a loss function motivated by IPM theory. This allows us to prove optimality guarantees on the generated solutions. 
We also relax the assumption that training data must solely consist of optimal solutions for different contexts.

\textbf{Active Learning.~}
Many operational applications are characterized by data that is scarce or expensive to obtain~\citep{gupta:2020small}. 
However, training models with high-dimensional data (e.g., decision vectors in $\field{R}^n$) requires large data sets~\citep{sun:2017revisiting}. 
An increasingly popular approach for learning with limited data is to synthetically generate and label new data~\citep{bastani:2018interpreting, zhang:21stylegan3d, zhang:21dataset}; this is also known as active learning by membership query synthesis~\citep{angluin:1988queries, wang:2015active}. 
For classification problems, given a means of generating artificial unlabeled data and a labeling oracle, one can iteratively synthesize a training set and re-train a model until it meets a desired performance level~\citep{settles:2009active}.
In our setting, the unlabeled data are context-decision tuples and the labels are feasible/infeasible.

In practice, the oracle can be a heuristic~\citep{zhang:21dataset}, a black-box~\citep{bastani:2018interpreting}, or even human decision-makers in-the-training-loop~\citep{castrejon:2017annotating, emmanouilidis2019enabling}. Note that our use of oracles or humans in-the-loop differs from conventional decision-making settings~\citep{melanccon:2021machine}, where they guide an algorithm when solving an optimization problem for a new context. In active learning, oracles are used in model training; at test time, we directly generate a context-dependent decision. 
Since oracles must label up to thousands of data points during training, we automate rule-based oracles by leveraging prior optimization structure. 

\section{Problem Definition}
\label{sec:problem_definition}

In this paper, vectors are denoted in bold and sets in calligraphic. The interior, boundary, and closure of a set are $\interior(\set{X})$, $\boundary(\set{X})$, and $\closure(\set{X})$, respectively. The exclusion of $\set{X}_1$ from $\set{X}_2 \supseteq \set{X}_1$ is denoted $\set{X}_2 \setminus \set{X}_1$ while $\left[x\right]^+ := \max\{x,0\}$. We denote probability distributions with $\field{P}$ and their supports with $\support(\field{P})$. Finally, $\norm{\cdot}$ denotes the $l_2$-norm.

Let $\bx \in \field{R}^n$ be a decision vector, $\bc \in \field{R}^n$ be a cost vector (we assume without loss of generality that $\norm{\bc} = 1$), and $\bu \in \set{U}$ be a context vector (e.g., customer demographics, patient information, decision-maker preferences) from a context space $\set{U} \subset \field{R}^p$. We consider the problem 
\begin{align*}
    \OP(\bu): \quad \min_{\bx} \left\{ \bc^\tpose \bx \;\Big|\; \bx \in \Xfeas(\bu) \right\},
\end{align*}
where $\Xfeas(\bu)$ is a feasible set that depends on the context vector $\bu$ via some relationship that must be estimated.
Solving this optimization problem is challenging because: (i) the relationship between $\bu$ and $\Xfeas(\bu)$ must be learned; and (ii) the corresponding  
context-dependent constraints $\Xfeas(\bu)$ may induce an optimization problem that is difficult to efficiently solve.

Let $\bx^*(\bu)$ be an optimal solution to $\OP(\bu)$. Our goal is to train a machine learning model $F: \set{U} \rightarrow \field{R}^n$ such that given a context vector $\bu$, it can generate a decision satisfying 
\begin{align}\label{eq:objective_func_error}
    \Big| \bc^\tpose F(\bu) - \bc^\tpose \bx^*(\bu) \Big| < \epsilon
\end{align}
for some $\epsilon > 0$. To facilitate this learning, we assume access to the following: 
\begin{itemize}
    \item A data set of context vectors $\datasetparam := {\left\{ \bhu_i \right\}}_{i=1}^\Nu$.

    \item A training data set of feasible decisions $\dataset := \{(\bhx_i, \bhu_i)\}_{i=1}^{\Nx}$, where $\bhu_i \in \datasetparam$ and $\bhx_i \in \Xfeas(\bhu_i)$ for all $i \in \{1, \dots, \Nx\}$. This data consists of decisions that were implemented in the past by domain experts or recommended by heuristic models and may include multiple decisions for each $\bhu_i$. 
    
    \item A training data set of infeasible decisions $\datasetinfeas := \{(\bbx_{\bar{i}}, \bhu_{\bar{i}})\}_{\bar{i}=1}^\Nbx$, where $\bbx_{\bar{i}} \in \field{R}^n \setminus \Xfeas(\bhu_{\bar{i}})$. 
    This data could consist of decisions that were previously proposed but not implemented.
    
    \item A polyhedron $\Xrel := \left\{ \bx \;\big|\; \ba_m^\tpose \bx \leq b_m,\; m=1,\dots,M \right\}$ whose interior bounds all context-dependent feasible sets, i.e., $\Xfeas(\bu) \subset \interior(\Xrel)$ for all $\bu \in \set{U}$. We refer to $\Xrel$ as \emph{bounding constraints}.  
\end{itemize}
Knowledge of $\Xrel$, which is a relaxation of any feasible set, does not sacrifice the generality of problem $\OP(\bu)$.
For instance, one can often define a bounding box such that $\Xrel := \left\{ \bx \;\big|\; -C \leq x_i \leq C, \; i = 1, \dots, n \right\} \supset \Xfeas(\bu)$ for all $\bu$ if $C$ is sufficiently large. A tighter $\Xrel$ may improve the learning rate of the algorithm.  
Finally, we make the following assumption on $\Xrel$ and $\set{U}$, to ensure that $\OP(\bu)$ will always have an optimal solution $\bx^*(\bu)$.
 
\begin{assumption}[Compactness] \label{ass:structure_of_op}
For all $\bu \in \set{U}$, the feasible set $\Xfeas(\bu)$ is compact and has a non-empty interior. Furthermore, $\left\{ (\bx, \bu) \;|\; \bu \in \set{U},\; \bx \in \Xfeas(\bu) \right\}$ is compact and has a non-empty interior.
\end{assumption}

\section{Interior Point Methods with Approximate Feasible Sets}
\label{sec:main_theory}

Before addressing our learning problem, 
we first develop an optimization theory for solving $\OP(\bu)$ by extending IPMs to the case where the barrier function may not exactly characterize the feasible region.  
Consider the problem $\OP(\bu)$ for a single context vector $\bu$. 
We define a barrier function that approximates the  feasible region with some error and then propose a barrier optimization problem that produces a solution to $\OP(\bu)$ with an optimality guarantee. We use this result to provide in-sample and out-of-sample optimality guarantees for our learning approach in Section~\ref{sec:ipman}.

For intuition, suppose first that $\Xfeas(\bu) = \Xrel$.
Here, the feasible set is is a fully-specified polyhedron and we can construct a canonical log-barrier, i.e., $\log B(\bx) = \sum_{m=1}^M \log\left(b_m - \ba_m^\tpose \bx\right)$, to solve $\OP(\bu)$ via an IPM~\citep[see][]{nemirovski:1994interior}. However, when $\Xfeas(\bu) \subset \Xrel$, this log-barrier may incorrectly return finite values for $\bx \in \Xrel \setminus \Xfeas(\bu)$. 
Thus, to generalize the concept of a barrier over a relaxation of the feasible set, we define a class of functions, which we call \emph{$\delta$-barriers}, that are strictly positive for all $\bx \in \Xfeas(\bu)$ and are zero for all $\bx$ that are sufficiently far from $\Xfeas(\bu)$.  
\begin{definition}
        For some $\delta > 0$, let $\neighbourhood{\Xfeas(\bu)} = \left\{ \bx + \bepsilon\;|\;\bx \in \Xfeas(\bu),\;  \norm{\bepsilon} < \delta \right\}$ be a neighborhood of $\Xfeas(\bu)$. A \emph{$\delta$-barrier} $\barr: \field{R}^n \times \set{U} \rightarrow [0,1)$ is a continuous function that satisfies
        \begin{align*}
                \Xfeas(\bu) \subset \left\{ \bx \;\big|\; \barr(\bx,\bu) > 0 \right\} \subseteq \neighbourhood{\Xfeas(\bu)}.
        \end{align*}
\end{definition}
%


A $\delta$-barrier is a function whose support is a bounded superset of $\Xfeas(\bu)$, where $\delta$ relates to the size of this superset. We visualize this and the below properties in Figure~\ref{fig:delta_barrier}~(Left). 
\begin{remark}
        The $\delta$ term is an upper bound on the Hausdorff distance $d_H(\cdot, \cdot)$ between the feasible set $\Xfeas(\bu)$ and the support of a function $B(\bx, \bu)$, where this support is a superset of $\Xfeas(\bu)$,
        %
        \begin{align}\label{eq:delta_barrier_haussdorf}
            d_H\left(\Xfeas(\bu), \left\{\bx\;\big|\; B(\bx, \bu) > 0 \right\}\right) := \min_{\xi \geq 0} \left\{ \xi \;\Big|\;\left\{ \bx\;\big|\;B(\bx, \bu) >0 \right\} \subseteq \neighbourhoodsub{\Xfeas(\bu)}{\xi} \right\} \leq \delta.
        \end{align}
\end{remark}
\begin{remark}
        Given the set $\Xrel$, we can always construct a $\delta$-barrier for $\OP(\bu)$. For instance, let $C^{\Xrel} > \max_{m\in\{1,\dots,M\}, \bx \in \Xrel} \left\{ b_m - \ba_m^\tpose \bx \right\}$ be a normalization factor and consider the function 
        \begin{align}\label{eq:canonical_barrier}
            B^{\Xrel}(\bx) := \prod_{m=1}^M \left[ \frac{b_m - \ba_m^\tpose \bx}{C^{\Xrel}}\right]^+.
        \end{align}
        This function is always bounded between $[0, 1)$ and has support equal to $\Xrel$, which is a bounded superset of $\Xfeas(\bu)$. Thus, $B^{\Xrel}(\bx)$ is a $\delta$-barrier where $\delta = d_H(\Xfeas(\bu), \Xrel)$.  
\end{remark}

Suppose we have a $\delta$-barrier $\barr(\bx, \bu)$ for some $\delta$. Let $\lambda > 0$ be a constant corresponding to a regularization parameter and consider the unconstrained barrier optimization problem
%
%
\begin{align}\label{eq:fwd_barrier_prob}
        \BP(\bu, \barr, \lambda):\quad \min_{\bx}\Big\{ \bc^\tpose \bx - \lambda \log{\barr(\bx, \bu)} \Big\}.
\end{align}
We now show that the optimal value of $\OP(\bu)$ is bounded by the optimal value of $\BP(\bu, \barr, \lambda)$. 
\begin{theorem}
        \label{thm:optimal_barrier}
        For any $\lambda > 0$, $\BP(\bu, \barr, \lambda)$ has an optimal solution $\bxl(\bu)$. Furthermore, this solution is $(\delta, \epsilon)$-optimal for $\OP(\bu)$:
        \begin{align}\label{eq:epsilon_optimality}
                \bc^\tpose \bxl(\bu) - \epsilon < \bc^\tpose \bx^*(\bu) < \bc^\tpose \bxl(\bu) + \delta,
        \end{align}
        where $\epsilon = C \lambda$ with $C$ being a positive constant.
\end{theorem}

\begin{figure}[t]
    \centering
        \resizebox{0.45\textwidth}{!}{%
            \begin{tikzpicture}
                \path [draw, ultra thick] (4, 0) -- (1.26, 3.79) -- (-3.20, 2.39) -- (-3.28, -2.28) -- (1.13, -3.83) -- (4, 0);
                \draw [thick, dotted] plot [smooth cycle] coordinates{(3, 0) (0.9, 2.8) (-2.4,1.8) (-2.46, -1.7) (0.8, -2.9)};
                \draw [thick, dotted] plot [smooth cycle] coordinates{(2, 0) (0.6, 1.89) (-1.6,1.2) (-1.65, -1.1) (0.5, -1.9)};
                \draw [thick, dotted] plot [smooth cycle] coordinates{(1, 0) (0.3, 0.9) (-0.8,0.6) (-0.82, -0.6) (0.3, -0.95)};
                \draw [ultra thick, dashed, fill=black!30, fill opacity=0.5] plot [smooth cycle] coordinates{(2.2, 0) (1.8, 1.5) (1.0, 2.5) (-1, 1) (-2.5, 0) (-1, -2) (1.5, -1.5)};
                \draw [ultra thick, dashed, fill=black!30, fill opacity=0., scale=2.8] plot [smooth cycle] coordinates{(2.2, 0) (1.8, 1.5) (1.0, 2.5) (-1, 1) (-2.5, 0) (-1, -2) (1.5, -1.5)};
                \draw (0, 0) node {\LARGE $\Xfeas(\bu)$};
                \draw (-1.8, 2.) node {\LARGE $\{\bx\;|\;B^{\Xrel}(\bx) > 0\}$};
                \draw (3.6, 4.4) node {\LARGE $\neighbourhood{\Xfeas(\bu)}$};
                \draw [thick, ->] (2.2, 0) -- (6.1, 0.4) node[midway, label=above:{\LARGE $\delta$}] {};
            \end{tikzpicture}
        }
        \resizebox{0.45\textwidth}{!}{%
            \begin{tikzpicture}
                \path [draw, ultra thick] (4, 0) -- (1.26, 3.79) -- (-3.20, 2.39) -- (-3.28, -2.28) -- (1.13, -3.83) -- (4, 0);
                \draw [thick, dotted] plot [smooth cycle] coordinates{(3, 0) (0.9, 2.8) (-2.4,1.8) (-2.46, -1.7) (0.8, -2.9)};
                \draw [thick, dotted] plot [smooth cycle] coordinates{(2, 0) (0.6, 1.89) (-1.6,1.2) (-1.65, -1.1) (0.5, -1.9)};
                \draw [thick, dotted] plot [smooth cycle] coordinates{(1, 0) (0.3, 0.9) (-0.8,0.6) (-0.82, -0.6) (0.3, -0.95)};
                \draw [ultra thick, dashed, fill=black!30, fill opacity=0.5] plot [smooth cycle] coordinates{(2.2, 0) (1.8, 1.5) (1.0, 2.5) (-1, 1) (-2.5, 0) (-1, -2) (1.5, -1.5)};
                \draw [->, ultra thick] (3.0, -2.6) -- (3.9, -3.5) node[midway, label=above:{$\bc$}] {};
                \draw (-2.5, 0) node[datapoint, label=right:$\bx^*(\bu)$] (xstar) {};
                \draw (0.2, -0.2) node[datapoint, label=above right:$\bx^{\lambda_1}(\bu)$] (x0) {};
                \draw (-1.2, -0.) node[datapoint, label=below right:$\bx^{\lambda_2}(\bu)$] (x1) {};
                \draw (-2., 0.4) node[datapoint, label=right:$\bx^{\lambda_3}(\bu)$] (x2) {};
                \draw (-2.4, 0.8) node[datapoint, label=above right:$\bx^{\lambda_4}(\bu)$] (x3) {};
                \draw (-2.8, 1.4) node[datapoint, label=above right:$\bx^{\lambda_5}(\bu)$] (x4) {};
                \draw [->, thick] plot [smooth] coordinates{(x0) (x1) (x2) (x3) (x4)};
                \draw [color=white, ultra thick, dashed, fill=black!30, fill opacity=0., scale=2.] plot [smooth cycle] coordinates{(2.2, 0) (1.8, 1.5) (1.0, 2.5) (-1, 1) (-2.5, 0) (-1, -2) (1.5, -1.5)};
            \end{tikzpicture}
        }
    \vspace{-1em}
    \caption{
    The polyhedron is $\Xrel$ and the filled shape is $\Xfeas(\bu)$. The dotted lines are level sets for $B^{\Xrel}(\bx)$.
    Left: The support of $B^{\Xrel}(\bx)$ is bounded within the neighbourhood $\neighbourhood{\Xfeas(\bu)}$, meaning that this function is a $\delta$-barrier where $\delta$ is equal to the Hausdorff distance.
    Right: A central path of solutions to $\BP(\bu, B^{\Xrel}, \lambda)$ for a decreasing sequence $\lambda_0  > \lambda_1 > \lambda_2 > \lambda_3$ (solid line). Solutions start where $B^{\Xrel}$ is largest and move to the boundaries. 
    }
    \label{fig:delta_barrier}
\end{figure}

The $(\delta, \epsilon)$-optimality inequalities in Theorem~\ref{thm:optimal_barrier} generalize the classical $\epsilon$-optimality~\citep{nemirovski:1994interior} to problems where we only possess a relaxation of the feasible set. When $\delta=0$, the bounding set is tight and \eqref{eq:epsilon_optimality} reduces to the classical $\epsilon$-optimality bound $\bc^\tpose \bxl(\bu) - \epsilon < \bc^\tpose \bx^*(\bu) < \bc^\tpose \bxl(\bu)$. The proof of Theorem~\ref{thm:optimal_barrier} does not require the classical IPM assumptions of coerciveness and self-concordance; the former is replaced with continuity and a compact feasible set, while the latter is only required for polynomial-time optimality guarantees. Similar to classical IPMs, the $(\delta, \epsilon)$-optimality of solutions to $\BP(\bu, \barr, \lambda)$ can be controlled by tuning $\lambda$. Because $\epsilon = C \lambda$ for a fixed $C$, as $\lambda$ goes to $0$, so does $\epsilon$. However, $\delta$ is a property characterizing the approximation error of the barrier function and does not change with $\lambda$.

Figure~\ref{fig:delta_barrier} (Right) shows a sequence of solutions $\bxl(\bu)$ for a decreasing set of regularization values $\lambda_1 > \lambda_2 > \cdots > \lambda_5$. Intuitively, we may expect that when $\lambda$ is large (i.e., $\epsilon$ is also large), $\bxl(\bu)$ may also have a large objective function value and be sub-optimal for $\OP(\bu)$. When $\lambda$ and $\epsilon$ are small, $\bxl(\bu)$ may have a small objective function value and could be infeasible for $\OP(\bu)$. Although counterexamples may exist if the $\delta$-barrier has a pathological structure, we show in Section~\ref{sec:ipman_delta_algorithm} that given a regularity assumption on the $\delta$-barrier function, we can guarantee that $\bxl(\bu) \in \Xfeas(\bu)$ is feasible for a sufficiently large $\lambda$ and that as $\lambda$ decreases, $\bxl(\bu)$ becomes infeasible. This leads to an interior point algorithm using $\delta$-barriers that is analogous to classical IPMs.

\section{Learning to Optimize with Data-Driven Barrier Functions}
\label{sec:ipman}

The previous section assumes that (i) we have access to a $\delta$-barrier; and (ii) we optimize $\OP(\bu)$ over a single context vector $\bu$. We now relax both assumptions in order to address our core problem of learning to generate $\bx^*(\bu)$ for any input $\bu$. 
We first develop a function, i.e., a binary classifier, that serves as a data-driven $\delta$-barrier for any given context vector $\bu$.
We then develop a generative model, or generator, that uses this trained classifier to learn to generate $(\delta, \epsilon)$-optimal solutions to $\OP(\bu)$ for any given $\bu$.

\subsection{Learning a $\delta$-Barrier Using Classification}
\label{sec:ipman_step1}

Suppose we possess a data set of feasible $(\bhx_i, \bhu_i) \in \dataset$ and infeasible $(\bhx_{\bar{i}}, \bhu_{\bar{i}}) \in \datasetinfeas$ decisions. To create a function that can act as a data-driven $\delta$-barrier for any $\bu$, we train a classifier $B(\bx, \bu)$ to predict $1$ whenever the input decision-context tuples are feasible, i.e., $\bhx_i \in \Xfeas(\bhu_i)$, and $0$ otherwise i.e., $\bhx_{\bar{i}} \notin \Xfeas(\bhu_{\bar{i}})$. 
After training, the classifier will output larger values when $\bx \in \Xfeas(\bu)$ and smaller values when $\bx \in \field{R}^n \setminus \Xfeas(\bu)$. 
If for a fixed $\bhu$, the trained classifier outputs positive values for all $\bx \in \Xfeas(\bhu)$, and zero for all $\bx$ sufficiently far from $\Xfeas(\bhu)$, then this function is supported over a relaxation of $\Xfeas(\bhu)$ and is a $\delta$-barrier for $\OP(\bhu)$. 
Figure~\ref{fig:steps_of_ipman} (Left) visualizes this intuition.

There exists a plethora of models and learning algorithms for developing binary classifiers including logistic regression, decision trees, random forests, and neural networks. We note two main considerations for our task. First, we want the classifier to be differentiable, since later in Section~\ref{sec:ipman_step2}, we will use it as a $\delta$-barrier to learn to solve a barrier problem via gradient descent. Second, to produce good approximations of the $\delta$-barrier for any $\bu \in \set{U}$, we require a model class $\set{B}$ that can express complex shapes.
Given these needs, we employ a neural network classifier. 

The most popular approach to training neural network classifiers is by gradient descent and minimizing the Binary Cross Entropy (BCE) loss function~\citep{goodfellow:2016deep}. 
Therefore, we define the \emph{Feasibility Classification Problem} as a BCE optimization problem wherein we train $B \in \set{B}$ to predict $(\bhx_{i}, \bhu_{i}) \in \dataset$ as feasible and $(\bhx_{\bar{i}}, \bhu_{\bar{i}}) \in \datasetinfeas$ as infeasible:
\begin{align}\label{eq:ipman_step1}
    \FCP(\dataset, \datasetinfeas):\quad\max_{B \in \set{B}}~\left\{\frac{1}{\Nx} \sum_{i=1}^\Nx \log{B(\bhx_i, \bhu_i)} + \frac{1}{\Nbx} \sum_{\bar{i}=1}^\Nbx \log{(1 - B(\bhx_{\bar{i}}, \bhu_{\bar{i}}))} \right\}.
\end{align}
In \eqref{eq:ipman_step1}, the classifier output (i.e., the estimated likelihood of a decision being feasible) is compared to whether the decision is actually feasible. The objective falls to $-\infty$ if the model incorrectly predicts $B(\bhx_i, \bhu_i) = 0$ or $B(\bhx_{\bar{i}}, \bhu_{\bar{i}}) = 1$ for any point in $\dataset$ or $\datasetinfeas$, respectively, but it is maximized when $B(\bhx_i, \bhu_i) = 1$ and $B(\bhx_{\bar{i}}, \bhu_{\bar{i}}) = 0$ for all points in $\dataset$ and $\datasetinfeas$, respectively.
Let $B^*(\bx, \bu)$ be a classifier trained by solving $\FCP(\dataset, \datasetinfeas)$.

In addition to BCE being an easy-to-implement approach to training a neural network classifier, we also provide an expectation-level argument as to why this loss function is appropriate. Let $\pfeas$ be a distribution of feasible pairs (i.e., $(\bx, \bu)$ where $\bx \in \Xfeas(\bu)$) and let $\pinfeas$ be a distribution of infeasible pairs (i.e., $(\bx, \bu)$ where $\bx \notin \Xfeas(\bu)$) corresponding to empirical data distributions. 
Then, consider the stochastic version of the Feasibility Classification Problem: 
%
\begin{align*}
    \SFCP(\pfeas, \pinfeas):\quad\max_{B \in \set{B}}~\left\{ \EX_{\bx,\bu\sim\pfeas} \Big[ \log{B(\bx, \bu)} \Big] + \EX_{\bx,\bu\sim\pinfeas} \Big[ \log\big(1-B(\bx,\bu)\big) \Big] \right\}.
\end{align*}
\begin{lemma}\label{lem:discriminator_infeas_obj}
    Assume that $\pfeas$ and $\pinfeas$ have closed and disjoint supports and that $\set{B}$ includes all continuous functions mapping $\field{R}^n \times \set{U} \rightarrow [0, 1]$. 
    \begin{enumerate}
        \item There exists a continuous function $B^*(\bx, \bu)$ of $\SFCP(\pfeas, \pinfeas)$ where $B^*(\bx, \bu) = 1$ for all $(\bx, \bu) \in \support(\pfeas)$ and $B^*(\bx, \bu) = 0$ for all $(\bx, \bu) \in \support(\pinfeas)$ that achieves an optimal value of $0$ for $\SFCP(\pfeas, \pinfeas)$.
        
        \item If $\support(\pfeas) = \{ (\bx, \bu) \;|\; \bx \in \Xfeas(\bu), \bu \in \set{U} \}$, then for any $\bu \in \set{U}$, the product $B^*(\bx, \bu) B^{\Xrel}(\bx)$ is a $\delta$-barrier with $\delta = d_H(\Xfeas(\bu), \Xrel)$.
    \end{enumerate}
\end{lemma}

\begin{figure}[t]
    \centering
        \resizebox{0.4\textwidth}{!}{%
        \begin{tikzpicture}
            \path [draw, color=white] (-3.4, -3.4) -- (-3.4, 3.4) -- (3.4, 3.4) -- (3.4, -3.4) -- (-3.4, -3.4); 
            \draw [ultra thick] plot [smooth cycle] coordinates{(2.9, 0) (1.4, 2.8) (-1.1, 2.4) (-2.5, 1.6) (-2.6, -0.8) (-1.6, -2.1) (-0.3, -2.6) (1.8, -2)}; 
            \draw [ultra thick, dashed, fill=black!30, fill opacity=0.5] plot [smooth cycle] coordinates{(2.2, 0) (1.8, 1.5) (1.0, 2.5) (-1, 1) (-2.5, 0) (-1, -2) (1.5, -1.5)};
            \draw (-2., 0) node[datasampleDiamond] (x1) {};
            \draw (-1.2, -1.2) node[datasampleDiamond] (x2) {};
            \draw (0.2, -1.4) node[datasampleDiamond] (x3) {};
            \draw (1.3, -0.8) node[datasampleDiamond] (x4) {};
            \draw (1.3, 0.8) node[datasampleDiamond] (x5) {};
            \draw (-0.2, 0.6) node[datasampleDiamond] (x6) {};
            \draw (0.6, 1.4) node[datasampleDiamond] (x7) {};
            \draw (-2., 0) node[datasampleDiamond] (x8) {};
            \draw (-3.2, 0.4) node[datasample] (x0) {};
            \draw (-1.8, -2.6) node[datasample] (x0) {};
            \draw (1.4, -2.8) node[datasample] (x0) {};
            \draw (2.7, -1.6) node[datasample] (x0) {};
            \draw (2.8, 1.8) node[datasample] (x0) {};
            \draw (-0.7, 3.0) node[datasample] (x0) {};
            \draw (-1.5, 1.5) node {$B(\bx, \bhu_i) > 0$};
            \draw (-2.3, 2.5) node {$B(\bx, \bhu_i) = 0$};
        \end{tikzpicture}
        }
        \resizebox{0.4\textwidth}{!}{%
        \begin{tikzpicture}
            \path [draw, color=white] (-3.4, -3.4) -- (-3.4, 3.4) -- (3.4, 3.4) -- (3.4, -3.4) -- (-3.4, -3.4); 
            \draw [ultra thick] plot [smooth cycle] coordinates{(2.9, 0) (1.4, 2.8) (-1.1, 2.4) (-2.5, 1.6) (-2.6, -0.8) (-1.6, -2.1) (-0.3, -2.6) (1.8, -2)}; 
            \draw [ultra thick, dashed, fill=black!30, fill opacity=0.5] plot [smooth cycle] coordinates{(2.2, 0) (1.8, 1.5) (1.0, 2.5) (-1, 1) (-2.5, 0) (-1, -2) (1.5, -1.5)};
            \draw (-2., 0) node[datasampleDiamond] (x1) {};
            \draw (-1.2, -1.2) node[datasampleDiamond] (x2) {};
            \draw (0.2, -1.4) node[datasampleDiamond] (x3) {};
            \draw (1.3, -0.8) node[datasampleDiamond] (x4) {};
            \draw (1.3, 0.8) node[datasampleDiamond] (x5) {};
            \draw (-0.2, 0.6) node[datasampleDiamond] (x6) {};
            \draw (0.6, 1.4) node[datasampleDiamond] (x7) {};
            \draw (-2., 0) node[datasampleDiamond] (x8) {};
            \draw (-3.2, 0.4) node[datasample] (x0) {};
            \draw (-1.8, -2.6) node[datasample] (x0) {};
            \draw (1.4, -2.8) node[datasample] (x0) {};
            \draw (2.7, -1.6) node[datasample] (x0) {};
            \draw (2.8, 1.8) node[datasample] (x0) {};
            \draw (-0.7, 3.0) node[datasample] (x0) {};
            \draw (0.2, -0.2) node[datasampleDiamond, fill, label=above right:$F^{(1)}(\bhu_i)$] (x0) {};
            \draw (-1.2, -0.) node[datasampleDiamond, fill, label=below:$F^{(2)}(\bhu_i)$] (x1) {};
            \draw (-2., 0.4) node[datasampleDiamond, fill, label=right:$F^{(3)}(\bhu_i)$] (x2) {};
            \draw (-2.4, 0.8) node[datasample, fill, label=above right:$F^{(4)}(\bhu_i)$] (x3) {};
            \draw (-2.4, 1.4) node[datasample, fill, label=above right:$F^{(5)}(\bhu_i)$] (x4) {};
            \draw [->, thick] plot [smooth] coordinates{(x0) (x1) (x2) (x3) (x4)};
        \end{tikzpicture}
        }
    \vspace{-1em}
    \caption{
    Given a training sample $\bhu_i$, the filled shape is $\Xfeas(\bhu_i)$. $\Diamond$ and $\Box$ represent feasible $\dataset$ and infeasible $\datasetinfeas$ decisions, respectively. 
    Left: The support of a classifier $B^*(\bx, \bhu_i)$ (black line) approximates the feasible set.
    Right: The sequence of solutions generated by $F^{(j)}(\bhu_i)$ (filled) approximate a sequence of $(\delta, \epsilon)$-optimal solutions.
    }
    \label{fig:steps_of_ipman}
\end{figure}

The first statement is a variation of a result by~\citet[Theorem 2.1]{arjovsky:2017towards}, which demonstrates that 
an optimal binary classifier trained by minimizing BCE will produce a continuous function that is equal to $1$ and $0$ over the distributions for each respective class.
The second statement indicates that a product of classifiers can be used as a $\delta$-barrier when trained with a probability distribution over feasible decisions.

Although the Lemma is stated in terms of expected risks over true data distributions, the assumption of closed and disjoint supports is satisfied with empirical data distributions and thus, the lemma holds without loss of generality. Further, Lemma~\ref{lem:discriminator_infeas_obj} assumes that the model class $\set{B}$ contains all continuous mappings of $\field{R}^n \times \set{U} \rightarrow [0, 1]$. This assumption is a sufficient rather than a necessary condition. Finally, although we theoretically require neural networks of infinite size in order to approximate any continuous function classifier (see~\citealt{hornik:1991}), in practice, we simply train a large but finite neural network with a large data set of different contexts and multiple feasible/infeasible decisions per context. 
Our numerical results show that neural network models with sufficiently high capacity
can be trained as high-quality classifiers to approximate $\delta$-barriers.

\subsection{Learning to Generate Solutions to the Data-Driven Barrier Problem}
\label{sec:ipman_step2}

Given a classifier that approximates a $\delta$-barrier for $\OP(\bu)$, we now develop an algorithm for training a generator to learns to output $(\delta, \epsilon)$-optimal solutions for any context $\bu$. Let $\set{F} := \{ F : \set{U} \rightarrow \field{R}^n \}$ denote a model class of generators. Let $j \in \{1, \dots, J\}$ denot a series of steps and let $\lambda_j$ be a regularization parameter where $\lambda_{j+1} < \lambda_j$ for all $j$. Recall that solving $\BP(\bu, \barr, \lambda_j)$ for a decreasing sequence of $\lambda_j$ can produce a sequence of solutions $\bx^{\lambda_j}(\bu)$ that slowly improve the objective function value until they become infeasible for sufficiently small values of $\lambda_j$. In an analogous fashion, we train a set of generators $F^{(j)}(\bu)$ over a decreasing sequence of $\lambda_j$. The sequence of generators $F^{(j)}(\bu)$ for $j=1,...,J$ should progressively produce decisions with lower objective function values, but for a sufficiently small value of $\lambda_j$, the generator may produce infeasible decisions, i.e., $F^{(j)}(\bu) \notin \Xfeas(\bu)$. Figure~\ref{fig:steps_of_ipman} (Right) visualizes this intuition.

Let $B^*(\bx, \bu)$ be the classifier trained in Section~\ref{sec:ipman_step1} and $B^{\Xrel}(\bx)$ be the canonical barrier from~\eqref{eq:canonical_barrier}. 
We now define the \emph{Generative Barrier Problem} as an empirical risk minimization problem: 
%
\begin{align}\label{eq:ipman_step2a}
    \GBP(\datasetparam, B^*, \lambda_j): \quad \min_{F \in \set{F}}\left\{ \frac{1}{\Nu} \sum_{i=1}^\Nu \bc^\tpose F(\bhu_i) - \lambda_j \log{\Big( B^*\big(F(\bhu_i), \bhu_i\big) B^{\Xrel}\big(F(\bhu_i)\big) \Big)} \right\},
\end{align}
The above equation is the machine learning analogue of~\eqref{eq:fwd_barrier_prob}. 
The first term is a linear cost with respect to training set decisions outputted by the generator, i.e., the objective of $\OP(\bu)$. The second term is the data-driven $\delta$-barrier, as determined by~\eqref{eq:ipman_step1}. Similarly to the log-barrier term in~\eqref{eq:fwd_barrier_prob}, it penalizes the objective by discouraging solutions that are far from the feasible set. Thus, the generator learns to produce low-cost decisions that the classifier would predict to be feasible.


We solve $\GBP(\datasetparam, B^*, \lambda_j)$ for $J$ steps to produce $J$ different generators similar to IPMs, where we would solve a barrier problem over different steps to obtain solutions with different optimality bounds (see~\ref{sec:ipman_delta_algorithm} for details on the optimization theory). However, we only need a single model to generate decisions at test time. Once training is complete, we select a particular generator $F^{(j*)}(\bu)$ that most often predicts feasible decisions with low objective function values by using a decision maker as an oracle to cross-validate the feasibility of decisions outputted by each model for a hold-out set of context vectors. Thus, $\lambda_j$ can be interpreted as a tuneable regularization parameter.

For each $j$, let $F^{(j)}(\bu)$ be an optimal generator trained by minimizing $\GBP(\datasetparam, B^*, \lambda_j)$. 
To further simplify the above equation, denote $B^{*, \Xrel}(\bx, \bu) := B^*(\bx, \bu) B^{\Xrel}(\bx)$ as a $\delta$-barrier defined as the product of two barrier functions. Note that the support of $B^{*, \Xrel}(\bx, \bu)$ is bounded since $\Xrel$ is assumed to be compact. This property is needed to prove an in-sample optimality bound below.

To justify training these models via empirical risk minimization of the barrier problem, we show that for a fixed $\lambda_j$, optimizing $\GBP(\datasetparam, B^*, \lambda_j)$ ensures that for all in-sample context vectors $\bhu_i \in \datasetparam$, the error between the solution generated by $F^{(j)}(\bhu_i)$
and the optimal value of $\OP(\bhu_i)$ is bounded. That is, the generative model outputs decisions satisfying the optimality guarantee given by~\eqref{eq:objective_func_error}. 
%
\begin{theorem}
        \label{thm:delta_epsilon_insample}
        For $\lambda_j >0$, let $F^{(j)}(\bx, \bu)$ and $B^{*, \Xrel}(\bx,\bu) = B^*(\bx, \bu) B^{\Xrel}(\bx)$ be the trained generator and product barrier, respectively. For any $\bhu_i \in \datasetparam$, let $\bxlj(\bhu_i)$ be an optimal solution to $\BP(\bhu_i, B^{*, \Xrel}, \lambda_j)$. Then, there exists $\delta, \epsilon > 0$ such that
        \begin{align}\label{eq:delta_epsilon_insample}
            \left| \bc^\tpose F^{(j)}(\bhu_i) - \bc^\tpose \bx^*(\bhu_i) \right| < \left| \bc^\tpose F^{(j)}(\bhu_i) - \bc^\tpose\bxlj(\bhu_i) \right| + \max(\delta, \epsilon).
        \end{align}
        %
\end{theorem}
%

Theorem~\ref{thm:delta_epsilon_insample} ensures that for any generated decision from an in-sample context vector $\bhu_i \in \datasetparam$, the optimality gap is bounded above by the sum of two terms: (i) the empirical error between the optimal solution to $\BP(\bhu_i, B^{*, \Xrel}, \lambda_j)$ and $\OP(\bhu_i)$ as given in Theorem~\ref{thm:optimal_barrier}; and (ii) a $(\delta, \epsilon)$-optimality bound. 
The value of this bound, and hence the quality of the trained generator $F^{(j)}(\bu)$, depends on the quality of the classifier. 
Recall that $\lambda_j$ is effectively a regularization parameter for training $F^{(j)}(\bu)$. If the classifier is a $\delta$-barrier for $\bhu_i$, then selecting an appropriate $\lambda_j$ ensures that the trained generator produces a solution that is arbitrarily close to the optimal value for $\OP(\bhu_i)$.

Note that Theorem~\ref{thm:delta_epsilon_insample} holds independently of Lemma~\ref{lem:discriminator_infeas_obj}, meaning that we do not necessarily need a perfect $\delta$-barrier to achieve a specified bound. However, as the proof of Theorem~\ref{thm:delta_epsilon_insample} demonstrates, a classifier that satisfies Lemma~\ref{lem:discriminator_infeas_obj} can reduce the size of the $\max(\delta, \epsilon)$ term. The proof also demonstrates that, similar to Theorem~\ref{thm:optimal_barrier}, $\epsilon$ can be calculated using known quantities. However, $\delta$ is a property of the classifier and the feasible set $\Xfeas(\bhu_i)$ which may not be fully characterized by the barrier function. 
As a consequence, we may not know the exact value of $\delta$. The next result demonstrates that we can bound this value as a function of the data set.

\begin{corollary}\label{cor:lb_on_delta}
        For any $\bhu_i \in \datasetparam$, $\delta$ in~\eqref{eq:delta_epsilon_insample} is bounded by $\delta \leq d_H\left( \left\{ \bhx \;|\; (\bhx, \bhu_i) \in \dataset \right\},\; \Xrel \right)$.
\end{corollary}

Corollary~\ref{cor:lb_on_delta} reveals the key challenge of our learning approach. While we may generate decisions that bound the optimal value by $\max(\delta, \epsilon)$, the size of $\delta$ is dependent on the available training data.
Indeed, in order to appropriately train a classifier to be used as a $\delta$-barrier, we require a significant amount of data that includes both feasible $\dataset$ and infeasible $\datasetinfeas$ decisions. While this setup is typical of many deep learning applications (e.g., the FaceNet system described in \citet{schroff2015facenet} uses more than $400,000$ samples to train a neural network to verify and recognize faces), 
many operational applications do not have data sets of this magnitude. Thus, in the next section, we embed the classifier and generator in an active learning framework, which allows us to augment the data used to train our machine learning models.

\section{Active Learning with Data-Driven Barrier Functions}
\label{sec:ipman_with_oracle}

The quality of the generator depends on the tightness of $\delta$, and therefore, on the quality of the classifier $B(\bx, \bu)$. 
Given data sets of limited size, we can improve our classifier using active learning by iteratively generating synthetic training data, labeling it, and then re-training the model~\citep{zhang:21dataset}. In our problem, the unlabeled data consists of decision-context pairs $(\bhx, \bhu)$. Thus, we must be able to (i) easily label incoming decisions (as either feasible or infeasible decisions for their corresponding context vectors); and (ii) synthesize new decision-context pairs.

Creating and labeling synthetic data is a challenging problem in general. To simplify our task, we leverage the available context vectors $\datasetparam$ and only synthesize new decision vectors $\bhx_i$ using the existing generators $F^{(j)}(\bhu_i)$ trained in Section~\ref{sec:ipman_step2}. Thus, our synthetic data consists of pairs $(F^{(j)}(\bhu_i), \bhu_i)$ for each $\bhu_i \in \datasetparam$ and we need only to determine whether a decision generated for a context is feasible. We first discuss how to create a labeling oracle by leveraging a priori knowledge of $\Xfeas(\bu)$.
We then present our active learning-based algorithm previewed in Figure~\ref{fig:flowchart} (Bottom).

\subsection{Constructing a Labeling Oracle}
\label{sec:labelingProcedure}

Given a trained generator $F(\bu)$ and $\bhu_i \in \datasetparam$, we require a means of determining whether $F(\bhu_i)$ is feasible or infeasible, i.e., a function $\Feasfunc$ that gives $\Psi(F(\bx, \bhu_i) = 1$ if $\bx \in \Xfeas(\bhu_i)$ and $0$ otherwise for any $\bhu_i \in \datasetparam$ and $\bx = F(\bhu_i)$. 
From the active learning literature, the most intuitive choice for $\Feasfunc$ is to employ human labelers. For example, AI companies building image classifiers will use Mechanical Turks to label image data~\citep{liao:2021towards}. 
However, due to the amount of data that is labeled in active learning, it is often advantageous to use automated labeling techniques via heuristics or black-box functions~\citep{wu:2015deep, zhang:21dataset, bastani:2018interpreting}.

While our learning approach 
does not assume any specific structure for $\Xfeas(\bu)$, we can leverage structural knowledge of the constraints and feasible decision data set to construct an automated, data-driven labeling function.
We discuss two examples, which are later used in our case studies, that assume $\Xfeas(\bu) := \{ \bx \;|\; g_r(\bx) \leq h_r(\bu), \; \forall r \in \{1, \dots, R\} \}$ is composed of a set of inequality constraints, where the left-hand-side, $g_r(\bx)$, are a priori known functions of the decision $\bx$, but the right-hand-side, $h_r(\bu)$, are functions of the context $\bu$ that must be estimated. Note that both $g_r(\bx)$ and $h_r(\bu)$ may be non-convex functions of $\bx$ and $\bu$, respectively. 
In our portfolio optimization case study (Section~\ref{sec:portfolio}), $g_r(\bx)$ is the expected risk of a portfolio $\bx$, while $h_r(\bu)$ represents the risk tolerance of an investor characterized by demographic features $\bu$. 

\begin{example}[Look-Up-Table Constraints]\label{ex:oracle_lt}
    To collect $\dataset$, $\datasetinfeas$, and $\datasetparam$, we may deploy a baseline generator for a small trial group of users and observe whether they accept or reject the produced decisions (see Figure~\ref{fig:flowchart} Top). 
    At the end of the trial period, we query these users to collect their specific values for $h_r(\bhu_i)$ with respect to the constraint functions $g_r(\bx)$. 
    Then during model training, we can label generated $F(\bhu_i)$ for $\bhu_i \in \datasetparam$ via the following look-up table:
    %
    \begin{align*}
        \Psi(\bx, \bhu_i) = \begin{cases}
            1 & \displaystyle\text{if } g_r(\bx) \leq h_r(\bhu_i), \; \forall r \in \{1,\dots,R\} \\
            0 & \displaystyle\text{otherwise}
        \end{cases},
        \qquad\qquad \forall \bhu_i \in \datasetparam.
    \end{align*}
\end{example}

Alternatively, in our personalized cancer treatment case study (Section~\ref{sec:rt}), 
$g_r(\bx)$ can be a clinical performance measure or criterion of a treatment protocol, while $h_r(\bu)$ represents how important the criterion is given the particular patient context. 
\begin{example}[On-off constraints]\label{ex:oracle_onoff}
    Suppose we know a priori that $h_r(\bu) \in \{h_{r, 1}, h_{r, 2}\}$ for all $\bu \in \set{U}$, where $h_{r, 1} < h_{r, 2}$. For instance, this can represent scenarios where a constraint is either ``on'' ($h_r(\bu) = h_{r, 1}$) or ``off'' ($h_r(\bu) = h_{r, 2}$) depending on the context. Given a context-decision pair $(\bhx_i, \bhu_i) \in \dataset$, we can estimate $h_r(\bhu_i)$ by evaluating whether $\bhx_i$ satisfies $h_{r, 1}$. That is, for any $\bhu_i$ such that there exists $(\bhx_i, \bhu_i) \in \dataset$, a labeling oracle can determine feasibility for a new $\bx$ by evaluating
    \begin{align*}
        \Psi(\bx, \bhu_i) = 
        \begin{cases}
            1  & \displaystyle\text{if } g_r(\bx) \leq h_{r, 1} \text{ and } \max_{(\bhx, \bhu_i) \in \dataset} g_r(\bhx) \leq h_{r, 1}, \; \forall r \in \{1,\dots, R\}  \\
            1  & \displaystyle\text{if } g_r(\bx) \leq h_{r, 2} \text{ and } h_{r, 1} \leq \max_{(\bhx, \bhu_i) \in \dataset} g_r(\bhx) \leq h_{r, 2}, \; \forall r \in \{1,\dots, R\}  \\
            0  & \text{otherwise}
        \end{cases},\qquad\qquad \forall \bhu_i \in \datasetparam.
    \end{align*}
\end{example}

The oracles presented above leverage the general structure of $\Xfeas(\bu)$ to automatically determine whether a decision is feasible. The primary advantage of using this approach is that because oracles can be called a large number of times during training, labeling can be performed efficiently. One disadvantage is that these oracles may mislabel the synthetic data points.
While we develop theory assuming a perfect feasibility oracle, we demonstrate in our numerical experiments that our approach is robust to erroneous or noisy oracles that may misclassify artifically generated points. 

\subsection{Interior Point Methods with Adversarial Networks (IPMAN)}
\label{sec:iterative_training}

We now present the complete learning algorithm shown in Figure~\ref{fig:flowchart} (Bottom), which leverages active learning in conjunction with the Feasibility Classification and Generative Barrier Problems. In each iteration of the IPMAN algorithm, we train a classifier and a generator in succession as described in Section~\ref{sec:ipman}. Then, for each context $\bhu_i \in \datasetparam$, we use the trained generator to produce decisions. These new data points are labeled and then added to the existing data sets of feasible and infeasible decisions. Let $k \in \{1, \dots, K\}$ index the $k$-th iteration of training and let superscript $(k)$ index the model and data sets after the conclusion of the $k$-th iteration (see Algorithm~\ref{alg:ipman_summary}).

\begin{algorithm}[t]
    \caption{Interior Point Methods with Adversarial Networks (IPMAN)}
    \label{alg:ipman_summary}
    \begin{algorithmic}[1]
            \renewcommand{\algorithmicrequire}{\textbf{Input:}}
            \renewcommand{\algorithmicensure}{\textbf{Output:}}
            \REQUIRE Number of IPMAN iterations $K$ and IPM steps $J$; Dual regularizers $\{ \lambda_j \}_{j=1}^J$; Initial data sets $\dataset^{(1)}$, $\datasetinfeas^{(1)}$, and $\datasetparam$.
            \ENSURE  Final generative models $F^{(j,K)}$ for $j \in \{1, \dots, J\}$
            \FOR{$k = 1$ \TO $K$}
                \STATE Train classifier $B^{(k)}$ via $\FCP(\dataset^{(k)}, \datasetinfeas^{(k)})$.
                \FOR{$j = 1$ \TO $J$}
                    \STATE Train generator $F^{(j, k)}$ via $\GBP(\datasetparam, B^{(k)}, \lambda_j)$.
                    \STATE Update $\dataset^{(k+1)} = \dataset^{(k)} \cup \set{Q}$, where $\set{Q}:= \left\{ (F^{(j, k)}(\bhu_i), \bhu_i) \;\Big|\; \Psi(F^{(j, k)}(\bhu_i), \bhu_i) = 1, \bhu_i \in \datasetparam\right\}$.
                    \STATE Update $\datasetinfeas^{(k+1)} = \datasetinfeas^{(k)} \cup \bar{\set{Q}}$, where $\bar{\set{Q}} := \left\{ (F^{(j, k)}(\bhu_i), \bhu_i) \;\Big|\;  \Psi(F^{(j, k)}(\bhu_i), \bhu_i) = 0, \bhu_i \in \datasetparam \right\}$.
                \ENDFOR
            \ENDFOR
            \RETURN $F^{(j,K)}$ for $j \in \{1, \dots, J\}$
    \end{algorithmic}
\end{algorithm}

In the $(k+1)$-th iteration, the classifier $B^{(k+1)}(\bx, \bu)$ is trained with feasible and infeasible data sets $\dataset^{(k+1)}$ and $\datasetinfeas^{(k+1)}$, respectively, which are larger than the data sets $\dataset^{(k)}$ and $\datasetinfeas^{(k)}$ due to the labeling procedure described in Section~\ref{sec:labelingProcedure}. Because of this data augmentation step, the optimal solution set of the Feasibility Classification Problem can be shown to contract after each iteration, i.e., $B^{(k+1)}(\bx, \bu)$ represents a tighter approximation to $\Xfeas(\bu)$ than $B^{(k)}(\bx, \bu)$.

\begin{proposition}\label{propn:shrinking_optimal_set}
Assume that $\set{B}$ contains all continuous functions mapping $\set{U} \times \field{R}^n \rightarrow [0, 1]$. For any $k$, let $\set{B}^{(k)}$ be the optimal solution set of $\FCP(\dataset^{(k)}, \datasetinfeas^{(k)})$. Then, $\set{B}^{(k+1)} \subset \set{B}^{(k)}$.
\end{proposition}
%

Proposition~\ref{propn:shrinking_optimal_set} demonstrates that by iteratively increasing the amount of available data using the oracle, we can contract the optimal solution set. Because the sets are augmented with correctly labeled data (assuming a perfect oracle), data augmentation removes classifiers whose supports are looser approximations to the different $\Xfeas(\bu)$. Intuitively, it allows the classifier to correct regions it had previously mislabeled as feasible and reinforce regions it had correctly labeled so that it does not incorrectly mislabel them in future iterations. Figure~\ref{fig:ipman_data_augmentation} visualizes this behavior; the black classifier is a tighter approximation of $\Xfeas(\bu)$ than the gray classifier due to the additional points.


\begin{figure}[t]
    \centering
        \resizebox{0.4\textwidth}{!}{%
            \begin{tikzpicture}
                \draw [ultra thick, opacity=0.2] plot [smooth cycle] coordinates{(2.9, 0) (1.4, 2.8) (-1.1, 2.4) (-2.5, 1.6) (-2.6, -0.8) (-1.6, -2.1) (-0.3, -2.6) (1.8, -2)}; 
                \draw [ultra thick] plot [smooth cycle] coordinates{(2.9, 0) (1.4, 2.8) (-1.1, 2.2) (-1.8, 1.6) (-2.4, 0.6) (-2.7, 0) (-2.6, -0.8) (-1.6, -2.1) (-0.3, -2.6) (1.8, -2)};
                \draw [ultra thick, dashed, fill=black!30, fill opacity=0.5] plot [smooth cycle] coordinates{(2.2, 0) (1.8, 1.5) (1.0, 2.5) (-1, 1) (-2.5, 0) (-1, -2) (1.5, -1.5)};
                \draw (-2., 0) node[datasampleDiamond] (x1) {};
                \draw (-1.2, -1.2) node[datasampleDiamond] (x2) {};
                \draw (0.2, -1.4) node[datasampleDiamond] (x3) {};
                \draw (1.3, -0.8) node[datasampleDiamond] (x4) {};
                \draw (1.3, 0.8) node[datasampleDiamond] (x5) {};
                \draw (-0.2, 0.6) node[datasampleDiamond] (x6) {};
                \draw (0.6, 1.4) node[datasampleDiamond] (x7) {};
                \draw (-2., 0) node[datasampleDiamond] (x8) {};
                \draw (-3.2, 0.4) node[datasample] (x0) {};
                \draw (-1.8, -2.6) node[datasample] (x0) {};
                \draw (1.4, -2.8) node[datasample] (x0) {};
                \draw (2.7, -1.6) node[datasample] (x0) {};
                \draw (2.8, 1.8) node[datasample] (x0) {};
                \draw (-0.7, 3.0) node[datasample] (x0) {};
                \draw (0.2, -0.2) node[datasampleDiamond, fill] (x0) {};
                \draw (-1.2, -0.) node[datasampleDiamond, fill] (x1) {};
                \draw (-2., 0.4) node[datasampleDiamond, fill] (x2) {};
                \draw (-2.4, 0.8) node[datasample, fill] (x3) {};
                \draw (-2.4, 1.4) node[datasample, fill] (x4) {};
                \draw (-0.2, 1.7) node {$B^{(k+1)}(\bx, \bhu_i) > 0$};
            \end{tikzpicture}
        }
    \caption{The trained classifier after an iteration of data augmentation. $\Diamond$ and $\Box$ represent points in $\dataset^{(k+1)}$ and $\datasetinfeas^{(k+1)}$, respectively, with the filled points denoting the augmented points. The filled region is $\Xfeas(\bhu_i)$. The gray line shows the support of $B^{(k)}(\bx,\bhu_i)$ and the black line shows the support of $B^{(k+1)}(\bx,\bhu_i)$.}
    \label{fig:ipman_data_augmentation}
\end{figure}


Proposition~\ref{propn:shrinking_optimal_set} assumes that all synthetic data is correctly labeled (i.e., a perfect oracle of feasibility), whereas it is possible that oracles may mislabel new data points. If $\Feasfunc$ is a noisy oracle, our data sets $\dataset$ and $\datasetinfeas$ may contain noisy labels for classification. We briefly remark that we can use algorithms from the deep learning literature on classification with noisy labels, which modify the BCE loss in $\FCP(\dataset, \datasetinfeas)$ to account for this~\citep{natarajan:2013learning, li:2017learning, han:2019deep}. 
Furthermore, our numerical experiments evaluate of the effect of a noisy oracle to show that our generative model is robust to a modest degree of misclassification.

\section{Generalization of $(\delta, \epsilon)$-Optimality to Out-of-sample Instances}
\label{sec:generalization_bound}

In this section, we analyze the potential of IPMAN, or any machine learning model used to generate solutions to contextually constrained optimization problems, to obtain $(\delta, \epsilon)$-optimal solutions when given an out-of-sample context vector $\bu$. To this end, we assume there exists a probability distribution over context vectors $\pparam$ and that the data set of context vectors $\datasetparam$ is sampled i.i.d. according to $\pparam$. We then use Rademacher complexity theory to obtain a probabilistic bound on the empirical error from an out-of-sample input~\citep{bartlett:2002rademacher}.

\begin{definition}[\citet{bertsimas:2014predictive}]
        Let $\set{F} \subset \{ F(\bu) : \set{U} \rightarrow \field{R}^n \}$ be a function class and $\datasetparam \sim \pparam$ be an i.i.d.\ data set. The empirical multivariate Rademacher complexity of $\set{F}$ is
        \begin{align*}
                \hRad(\set{F},\datasetparam)  = \EX_{\bsigma \sim p_{\bsigma}} \Bigg[ \frac{2}{\Nu} \sup_{F \in \set{F}}\sum_{i=1}^\Nu \bsigma_{i}^\tpose F(\bhu_i) \;\Bigg|\; \datasetparam = {\{\bhu_i\}}_{i=1}^\Nu \Bigg],
        \end{align*}
        where $\bsigma_{i} \sim p_{\bsigma}$ is an $n$-dimensional vector of i.i.d. Rademacher variables. The multivariate Rademacher complexity of $\set{F}$ is $\Rad(\set{F}) =  \EX_{\datasetparam \sim \pparam} \left[ \hRad(\set{F}, \datasetparam ) \right]$.  
\end{definition}

In statistical learning theory, Rademacher complexities are used to generate risk bounds that are dependent on the class $\set{F}$ of learning model that is used and the training data~\citep{bartlett:2002rademacher}. \citet{bertsimas:2014predictive} apply the theory to develop generalization bounds for producing decisions to problems with a conditional stochastic optimization objective. We extend their work by providing a probabilistic bound on $(\delta, \epsilon)$-optimality when the feasible set is estimated. Typically, more complex model classes are required to learn more complex relationships. The IPMAN algorithm is agnostic to the class of model used. However, if we need a tight generalization bound, we should use a model class $\set{F}$ with a tightly bounded Rademacher complexity.
%
\begin{remark}
        Although the literature mostly focuses on the univariate Rademacher complexity,~\citet{maurer:2016vector} and~\citet{bertsimas:2014predictive} prove bounds for linear multivariate classes (e.g., $\set{F}_R = \{ \bW \bu \mid \norm{\bW} \leq R \}$). In general, if $F(\bu) = (F_1(\bu),\dots,F_n(\bu))$, then $\set{F} \subset \times_{\ell=1}^n \set{F}_\ell$, where $\set{F}_\ell = \{ {F(\bu)}^\tpose \bolde_\ell \;|\; F \in \set{F} \}$ and $\bolde_\ell$ is the $\ell$-th identity vector. Then, $\hRad(\set{F}, \datasetparam) \leq \sum_{\ell=1}^n \hRad(\set{F}_\ell, \datasetparam)$ decomposes to a sum of univariate complexities. We refer to~\citet{bartlett:2002rademacher} for bounds on linear and tree models and~\citet{bartlett:2002rademacher, neyshabur:2015norm} and \citet{foster:2018uniform} for neural networks. 
\end{remark}

For the ease of theory, we assume that our generator is trained by minimizing $\GBP(\datasetparam, \barr, \lambda)$ with a $\delta$-barrier $\barr(\bx, \bu)$ for some $\delta > 0$. Note that since we are using a fixed data set $\datasetparam$, there is no guarantee that $F(\bu)$ will be feasible or even satisfy $\barr(F(\bu), \bu) > 0$ for an arbitrary $\bu$. Because $\Xrel$ is available, however, we can always project any generated solution to the polyhedron to ensure that the generated decisions are bounded.
\begin{assumption}\label{ass:well_behaved_generator}
        Consider a $\delta$-barrier $\barr(\bx, \bu)$ and generator $F(\bu)$ trained via $\GBP(\datasetparam, \barr, \lambda)$. We use projections $F^*(\bu) = \argmin_{\bx} \left\{ \norm{\bx - F(\bu) } \;|\; \bx \in \Xrel \right\}$ at test time. 
\end{assumption}

Our generalization bound follows from~\citet{bertsimas:2014predictive}. While they derive an empirical risk bound on an unconstrained stochastic optimization problem, we focus on $(\delta, \epsilon)$-optimality for a constrained continuous optimization problem (see~\ref{sec:ec_proof_of_prob_bound} for the proof and discussion).

\begin{theorem}\label{thm:prob_bound_generative_model}
        Let $F^*$ satisfy Assumption~\ref{ass:well_behaved_generator}, $K$ and $L_\infty$ be sufficiently large positive constants, and fix $\beta \in (0, 1)$. Then, for any $\gamma > 0$, the following inequality holds 
        \begin{align*}
                &\pparam\Big\{ \bc^\tpose F^*(\bu) - \epsilon - \gamma< \bc^\tpose \bx^*(\bu) < \bc^\tpose F^*(\bu) + \delta + \gamma \Big\} \\
                &\qquad \geq 1 - \frac{\displaystyle\frac{1}{\Nu}\sum_{i=1}^\Nu \left| \bc^\tpose F^*(\bhu_i) - \bc^\tpose \bxl(\bhu_i) \right| + K \sqrt{\frac{\log(1/\beta)}{2\Nu}} + \sqrt{2n} L_\infty \Rad(\set{F}) }{\gamma},
        \end{align*}
        with probability at least $1-\beta$ with respect to the sampling of $\datasetparam \sim \pparam$.
\end{theorem}
%

Theorem~\ref{thm:prob_bound_generative_model} bounds the $(\delta,\epsilon)$-optimality of a random context vector. Given a $\gamma$ and $F^*(\bu)$, it computes the probability that the model will predict a $(\delta + \gamma, \epsilon + \gamma)$-optimal solution for an out-of-sample context vector $\bu_{\Nu+1} \sim \pparam$. The first term $(\sum_{i=1}^\Nu | \bc^\tpose F^*(\bhu_i) - \bc^\tpose \bxl(\bhu_i) |)/\Nu$, is the empirical error associated with solving $\GBP(\datasetparam, \barr, \lambda)$ versus $\BP(\bhu_i)$ for all $\bhu_i \in \datasetparam$. It effectively measures how well the model performs in-sample. The second term is dependent on the constants $K$ and $1/\beta$ and scales with $O(1/\Nu)$. Thus, as the size of the data set increases, the smaller this term becomes. The third term is dependent the Rademacher complexity of $\set{F}$, where the greater the representative capacity of the learning model, the larger its Rademacher complexity. The summation of these terms is a lower bound on the probability of producing optimal decisions for unseen instances and holds with probability at least $1-\beta$. Thus, in order to obtain a tight bound, we must balance the trade-off between a model class with high complexity versus obtaining a model with low error.

\section{Case Studies}
\label{sec:case_studies}

In this section, we perform two numerical case studies where we compare IPMAN to state-of-the-art optimization and deep learning baselines, respectively. In the first set of experiments, we 
consider an online automated investment service whose users require investment strategies with personalized risk tolerances and investment quotas~\citep{vielma:2017extended, swezey:2018large, li:2016inverse, bertsimas:2022scalable}. First-time investors, who represent the target market of these services, may not be able to accurately identify their ideal investment preferences and incorporate them as parameters into a portfolio optimization problem. The conventional solution is to estimate the parameters from user profiles or questionnaires 
and then solve an optimization problem. However, online services typically require latencies of less than 500 milliseconds~\citep{wang:2012far, crankshaw:2014missing, zhao:2018leveraging}, and a mobile service employing mathematical programming for real-time user interfacing will likely be too slow, especially if the optimization problem has complicating constraints. Instead, we train IPMAN to generate decisions that replicate the 
performance of an industrial-strength optimization solver such as Gurobi~\citep{gurobi}. This model can then be deployed by an online service to produce high-quality solutions to a contextual, non-convex optimization problem in milliseconds. We find that:
\begin{enumerate}
    \item IPMAN-trained models generate portfolios in approximately 130 ms, while a predict-then-optimize baseline often requires up to 100 s. Moreover, on average, the portfolios generated by IPMAN are within a $17\%$ optimality gap of a predict-then-optimize baseline that perfectly predicts parameters for every context. Thus, we find that IPMAN is competitive with optimistic optimization baselines that are tuned to quickly generate solutions. 
    
    \item The quality of decisions produced by predict-then-optimize approaches are sensitive to estimation errors in the prediction of the constraint parameters. Those solutions are often infeasible with respect to the true constraints even for small deviations from the true parameters. However, even with an similarly noisy labeling oracle, we demonstrate that IPMAN can generate decisions that are more robust to labeling errors.
    
\end{enumerate}

In the second case study, we investigate a treatment design problem. 
Radiation therapy (RT) is a cancer treatment procedure in which high-energy x-rays deliver radiation to a patient to destroy a tumor
~\citep{delaney:2005role}. 
Treatment plans are evaluated using a set of non-convex, pass-fail constraints that vary from clinic to clinic. Since not all constraints can be simultaneously satisfied in general, treatment planners manually tune parameters in the treatment optimization software and re-solve the problem until an oncologist approves the treatment plan for delivery. Modern treatment planning methods have moved towards machine learning to automate the planning process and reduce manual re-planning iterations~\citep{kaderka:2021wide}.  
This involves generating a dose distribution that satisfies a relevant subset of the pass-fail constraints that an oncologist would have deemed acceptable for the patient. The current state-of-the-art uses deep learning models trained on medical images from previous patients with approved treatment plans~\citep{Kearney:2018aa, Babier:2019aa}. Instead, by treating the constraints satisfied by previous oncologist-approved doses as a set of patient-specific complicating constraints,
IPMAN can generate doses that better meet contextually relevant clinical constraints while simultaneously delivering lower radiation to healthy tissue. 
We find that:
\begin{enumerate}
    \item Compared to supervised learning baselines, IPMAN-trained models yield dose distributions that are feasible $7\%$ more often and result in $7\%$ less radiation to healthy tissue. 
    
    \item IPMAN demonstrates better generalization. That is, by encoding clinical policies as constraints, IPMAN-trained models can produce treatments that better meet a new clinic's metrics when trained using approved treatments from another clinic. This suggests that IPMAN can leverage mathematical structure to generate decisions that learn from data whose covariate contexts are shifted from the target environment.
\end{enumerate}

\subsection{Automated Portfolio Optimization: 
A Comparison with Optimization}
\label{sec:portfolio}

We consider a non-convex, cardinality-constrained portfolio optimization model \citep[see][]{kim:2019gan} with $n=100$ stocks and $p=10$ contextual features per user. 
These contextual features map to a personalized risk tolerance, as well as minimum and maximum constraints on the number of stocks each user is willing to purchase, which are parameters that would typically require estimation. 
Motivated by recent successes of predict-then-optimize frameworks in estimating parameters for portfolio optimization~\citep[e.g.,][]{ban:2018machine, elmachtoub:2017smart, ferber:2020mipaal, yu:2020learning}, we develop several optimization baselines for comparison. 
Below, we highlight the setup and main results. We include details on the decision-making problem, data generation process, baselines, and algorithms, as well as additional ablations in  Section~\ref{sec:ec_portfolio}.

\subsubsection{Data and Model.}

We collect daily prices of $100$ stocks sampled from the Nasdaq, NYSE, and NYSE American over a four week period from August 1, 2020 to construct a return vector $\bmu \in \field{R}^n_+$ and a covariance matrix $\bSigma \in \field{S}^{n\times n}_+$. Then, we create a framework for generating artificial data that simulates investor behavior by sampling three quantities, $\bb_r, \bb_c, \bb_\Delta \in \field{R}^p_+$, that represent the ``true relationship'' of an user's (i) risk tolerance $r(\bu) := \bb_r^\tpose \bu $, (ii) minimum cardinality constraint $c(\bu) := \bb_c^\tpose \bu + 1$, and (iii) cardinality bandwidth requirement $\Delta c(\bu) := \bb_\Delta^\tpose \bu + 1$. Without loss of generality, these relationships are generated so that the average user desires a low-risk portfolio containing five to ten stocks, making the optimization problems sufficiently difficult.

Given the above user data generation model, we create a data set of i.i.d. randomly sampled context vectors $\bu \in \field{R}^p_+$ for $14,000$ users and then randomly split the data into three groups of size $10,000$, $2,000$, and $2,000$ for training, validation, and hold-out testing, respectively. User parameters are used as input to the following personalized portfolio optimization problem
\begin{subequations}\label{eq:PO_problem}
\begin{align*}
    \PP(\bu):\quad\minimize{\bx}\quad  & -\bmu^\tpose \bx  \\ 
    \subjectto \quad  
    & \bx^\tpose \bSigma \bx \leq r(\bu) \\ 
    & \lfloor c(\bu) \rfloor \leq \card(\bx) \leq \lfloor c(\bu) \rfloor + \lceil \Delta c(\bu) \rceil \\
    & \bx^\tpose \bone = 1, \quad \bx \geq \bzero,  
\end{align*}
\end{subequations}
where $\Xfeas(r(\bu), c(\bu), \Delta c(\bu)) := \{ \bx  \;|\; \bx^\tpose \bSigma \bx \leq r(\bu) , \; \lfloor c(\bu) \rfloor \leq \card(\bx) \leq \lfloor c(\bu) \rfloor + \lceil \Delta c(\bu) \rceil \}$ represents context-dependent feasible sets and the simplex $\Xrel := \left\{ \bx \;|\; \bx^\tpose \bone = 1, \; \bx \geq \bzero \right\}$ are the bounding constraints. 
Note that our data-driven portfolio optimization problem differs from the common setting where $\bmu$ and/or $\bSigma$ are estimated. In our case, we assume $\bmu$ and $\bSigma$ are known~\citep{li:2016inverse, yu:2020learning}, while the parameters $r(\bu), c(\bu)$, and $\Delta c(\bu)$ are unavailable for test set users. 
However, we have collected the true parameters $r(\bhu_i)$, $c(\bhu_i)$, and $\Delta c(\bhu_i)$ for users in our training and validation sets. 
Not only is this necessary to evaluate model quality in training by evaluating whether training decisions satisfy the context-dependent constraints, but we can use this information to construct a look-up-table-based feasibility oracle such as the one introduced in Example~\ref{ex:oracle_lt}. The oracle is:
\begin{align} \label{eq:portfolio_oracle}
    \Psi(\bx, \bhu_i) = \begin{cases}
        0 & \text{if~} \bx \notin \Xrel \\
        0 & \text{if~} \bx \notin \Xfeas(\bb_r^\tpose \bhu_i, \bb_c^\tpose \bhu_i + 1, \bb_\Delta^\tpose \bhu_i + 1) \\
        1 & \text{otherwise}
    \end{cases}, & \qquad\qquad\forall \bhu_i \in \datasetparam
\end{align}

To construct our initial $\dataset^{(1)}$ and $\datasetinfeas^{(1)}$, we generate 10 feasible and 10 infeasible portfolios per user $\bhu_i \in \datasetparam$ (for a total of $100,000$ data points) by solving random approximations and relaxations of $\PP(\bhu_i)$. We then implement two feed-forward neural networks as our generator and classifier. We train these models with IPMAN for $K=150$ iterations. Finally, we consider a single dual regularization parameter $\lambda = 5 \times 10^{-4}$, which was selected via cross-validation using our training and validation sets.
We implement IPMAN using PyTorch with an Nvidia GTX 1080 GPU. At test time, our generator can generate a decision in approximately $130$ ms.

Our baseline models consist of predict-then-optimize frameworks that estimate $r(\bu)$, $c(\bu)$, and $\Delta c(\bu)$ from $\bu$ and then solve $\PP(\bu)$.
Rather than building a specific prediction model, we instead assume that the baselines have perfect predictions of these parameters in the experiments in Section~\ref{sec:portfolio_exp1}, and then evaluate the effect of noisy estimators in Section~\ref{sec:portfolio_exp2}. 
Furthermore, we implement three different optimization algorithms as baselines: a Mixed-Integer Quadratic Program (MIQP), a low-latency version MIQP ($0.2$ s) where we set the time out to $200$ ms to match IPMAN, and a constrained regression reformulation (MIQP-R) that replaces the risk constraint with a Lagrangian relaxation and a quadratic regularization~\citep{bertsimas:2022scalable}. The MIQP-R baseline represents a conventional approach for addressing complicating constraints by leveraging custom structure in the optimization problem. 
All baselines are solved using Gurobi $8.1.1$ with a time out of $100$ s for MIQP and MIQP-R.

\subsubsection{Experiment 1: Quickly Generating High-quality Decisions.}
\label{sec:portfolio_exp1}

We evaluate our generator over the set of solvable $\PP(\bhu_i)$ by analyzing the average optimality gap and the fraction of generated solutions that satisfy the constraints. Feasibility is defined as meeting all constraints to within a relative tolerance of $5\%$. Note that \citet{kim:2019gan} use approximately the same tolerance level for their related work in generating optimal portfolios.

\begin{table}[t]
\centering
\caption{Summary statistics comparing IPMAN with optimizer baselines on the held-out testing set. 
The Optimizer runtime windows are stratified by the MIQP baseline.
The optimality gap is measured over only feasible decisions and with respect to the MIQP solutions. 
The IPMAN runtime is approximately 0.13 s for all problems. 
$^*$The `All points' test statistics evaluate instances where MIQP had a runtime less than 100 s (i.e., 41.1\% of all instances). 
}
\small
\begin{tabular}{l p{0.3\linewidth} cccc} \toprule
                                    & & All points$^*$ & \multicolumn{3}{c}{MIQP optimizer runtime windows}    \\ \cmidrule(l){4-6}
                                    & &         & $0.1-1$ s & $1-10$ s & $10-100$ s  \\ 
                                    \midrule
\multirow{2}{*}{MIQP}
& Fraction of instances (\%)          & 58.9    & 6.3     & 26.8   & 25.8    \\ \cmidrule(l){2-6}
& Average model run time (s)       & 25.9    & 0.7     & 5.6    & 53.2        \\
& Fraction of feasible decisions (\%)      & 100    & 100    & 100   & 100     \\ 
\midrule
\multirow{3}{*}{MIQP (0.2 s)}
& Average model run time (s)                           & 0.2     & 0.2     & 0.2    & 0.2            \\
& Fraction of feasible decisions (\%)      & 100     & 100     & 100    & 100      \\ 
& Optimality gap (\%)                                     & 40.9    & 10.5    & 36.0   & 53.6      \\ 
\midrule
\multirow{3}{*}{MIQP-R}
& Average model run time (s)                           & 18.4    & 4.2     & 10.1   & 30.6         \\
& Fraction of feasible decisions (\%)      & 88.8    & 100     & 96.5   & 78.3    \\ 
& Optimality gap (\%)                                     & 4.9     & 11.5    & 2.1    & 6.4    \\ 
\midrule
\multirow{3}{*}{IPMAN}
& Average model run time (s)                           & $<$0.2    & $<$0.2    & $<$0.2    & $<$0.2             \\
& Fraction of feasible decisions (\%)      & 97.6    & 100     & 99.6    & 95.0       \\ 
& Optimality gap (\%)                                     & 17.4    & 25.9    & 22.2    & 9.9     \\ 
\bottomrule         
\end{tabular}
\label{tab:portfolio_test}
\end{table}

Table~\ref{tab:portfolio_test} compares IPMAN with optimization models stratified by optimizer runtime. 
Generating a decision with IPMAN requires, on average, 130 ms, whereas the baseline optimizer MIQP requires over $100$ s for $41.1\%$ of the test set instances. Furthermore, for problems that require less than $100$ s for MIQP,  IPMAN-generated decisions are consistently feasible and achieve low optimality gap, making our generative model a high-quality approximator of the optimizer in orders of magnitude less time. 
The low-latency MIQP, which has similar decision-generation time as IPMAN, achieves an average optimality gap of $40.9$\%, which is more than two times a large as IPMAN. 
We also compare against MIQP-R and observe that IPMAN-generated decisions are still generated in orders of magnitude less time and are feasible on average $8$ percentage points more often; however, the MIQP-R baseline has lower optimality gap, conditional on it finding a feasible solution. 
We conclude that our method can deliver decisions that are feasible nearly all of the time but with better objective function value than an optimization model under similar latency requirements.

Finally, recall that these experiments compare against a \emph{perfect} predict-then-optimize baseline that always knows the true parameter values of the optimization model. A realistic predict-then-optimize framework would inevitably introduce estimation errors in the parameters. We next show that prediction errors in constraint parameters lead to large optimization errors whereas IPMAN can generate feasible decisions even in the face of noisy feasibility estimates. 

\subsubsection{Experiment 2: Predicting Optimal Solutions under Noisy Information.} 
\label{sec:portfolio_exp2}

\begin{table}[t]
\centering
\caption{Summary statistics with mean and standard deviation over five trials comparing IPMAN with the MIQP (0.2 s) baseline on the testing set. Optimality gap is evaluated over only feasible decisions. The best approach for each noise level is shaded.}
\small
\begin{tabular}
{l p{0.17\linewidth} p{0.15\linewidth} p{0.17\linewidth} p{0.15\linewidth} }
\toprule
Noise $\sigma$
        & \multicolumn{2}{c}{Fraction of feasible decisions (\%)} & \multicolumn{2}{c}{Optimality gap (\%)} \\
        \cmidrule(l){2-3} \cmidrule(l){4-5}
        & MIQP (0.2 s) & IPMAN & MIQP (0.2 s) & IPMAN \\ 
        \midrule
0.0025  &          $92.5 \pm 16.1$ & \graycell$94.9 \pm 3.3 $ & $41.0 \pm 2.0 $ & \graycell$39.3 \pm 22.4$ \\
0.0075  &          $60.7 \pm 44.3$ & \graycell$70.5 \pm 14.6$ & $41.0 \pm 4.9 $ & \graycell$28.4 \pm 5.1 $ \\
0.0125  &          $52.6 \pm 34.3$ & \graycell$65.4 \pm 19.3$ & $43.3 \pm 1.4 $ & \graycell$31.5 \pm 4.8 $ \\
0.0175  &          $40.3 \pm 48.7$ & \graycell$45.9 \pm 21.5$ & $46.5 \pm 7.8 $ & \graycell$29.4 \pm 6.7 $ \\    
0.0225  & \graycell$55.8 \pm 43.6$ &          $45.9 \pm 32.7$ & $55.3 \pm 9.5 $ & \graycell$37.5 \pm 18.1$ \\
0.0275  & \graycell$31.5 \pm 42.6$ &          $12.9 \pm 11.0$ & $58.0 \pm 24.0$ & \graycell$32.8 \pm 6.2 $ \\ 
\bottomrule
\end{tabular}
\label{tab:portfolio_noise}
\end{table}

In practice, there may be a discrepancy between user features and problem parameters. As a result, predict-then-optimize models would introduce estimation errors and IPMAN would synthesize noisy labeled data. 
We explore this issue by considering a noisy predict-then-optimize baseline versus a noisy oracle for IPMAN. 
Since the parameter data has a linear relationship with the user's features ($\bb_r$, $\bb_c$, and $\bb_\Delta$), we replace $\Psi(\bx, \bhu_i)$ in~\eqref{eq:portfolio_oracle} with a black-box oracle that uses a noisy linear model $\bb_r + \bepsilon_r$, $\bb_c + \bepsilon_c$, and $\bb_\Delta + \bepsilon_\Delta$ where $\bepsilon_r, \bepsilon_c, \bepsilon_\Delta \sim \set{N}(0, \sigma)$. 
The predict-then-optimize baseline uses the same noisy linear model to estimate constraint parameters for test set users.

We perform five trials with progressively increasing the noise variance $\sigma$ from $0.0025$ to $0.0275$. 
Note that at the highest level (i.e., $\sigma=0.0275$), the noise variance is almost half the average risk $\field{E}[\bb_r^\tpose \bu] = 0.066$ (see Section~\ref{sec:ec_portfolio} for further details). This implies that as the noise variance increases, we may find instances of optimization problems that are infeasible for the baseline model and for the oracle in IPMAN. 
In Table~\ref{tab:portfolio_noise}, we compare IPMAN against the noisy predict-then-optimize MIQP ($0.2$ s). 
Both methods have equivalent run times, which allows fair comparisons of the fraction of feasible decisions and average optimality gap on the test set.

When the noise level is modest (i.e., $\sigma \leq 0.0175$), IPMAN can consistently generate both feasible and optimal decisions. In contrast, MIQP ($0.2$ s) has a larger average optimality gap while the average fraction of feasible decisions is lower; both metrics can be up to $12$ percentage points worse than IPMAN. Furthermore, the variance in the feasibility rate between instances is several times larger for  MIQP ($0.2$ s) as compared to IPMAN. For large noise levels (i.e., $\sigma \geq 0.0225$), although the average feasibility rate for MIQP ($0.2$ s) is higher than IPMAN, the variance between instances is so large so as to render the model unpredictable. The findings suggest that, as compared to a competitive predict-then-optimize baseline, IPMAN generates optimal solutions that are more robust to noise while the variance remains consistent between instances.

The intuition behind these results is that due to the mislabeled decision data, the classifier may incorrectly have support over new infeasible regions for many $\bhu_i \in \datasetparam$. 
As a consequence, $\GBP(\datasetparam, \lambda, B^*)$ may train the generator on larger relaxations of the true feasible sets, leading to lower feasibility rates as the noise increases. 
On the other hand, the initial data set contains a diverse set of correctly labeled decisions. Moreover, the barrier function term in~\eqref{eq:ipman_step2a} acts as a regularizer controlled by $\lambda$ that helps the generator learn to produce more robust solutions. Thus, IPMAN can consistently generate decisions with lower optimality gaps even when sacrificing feasibility. In contrast, MIQP ($0.2$ s) must find an optimal solution with respect to incorrect parameters $r(\bu), c(\bu), \Delta c(\bu)$ obtained from a noisy estimator. These estimation errors are not discernible to the optimization model, meaning MIQP ($0.2$ s) may not find a solution that is also feasible under the true problem parameters, even with small amounts of noise. Finally, the latency requirement means that MIQP ($0.2$ s) struggles to find solutions with small optimality gaps.

\subsubsection{Implementation Guidelines.}

In these experiments, we simulate investor behavior with an artificial data set of user contexts. 
In a real-world implementation, one would first deploy a baseline model (which could be an alternate ML model or a predict-then-optimize framework) over a trial period to a group of 
users. For each user in the group, one would collect their context features, recommend investments with the baseline, and observe their adherence to the recommendations in order to construct our decision data sets. At the end of the trial period, each user could be surveyed, and their true risk tolerance levels and cardinality requirements would be recorded. One could then use IPMAN to train a model and validate it by using these recorded preferences. This validation process would also serve as the feasibility oracle in the active learning algorithm.

\subsection{Personalized Cancer Treatment: 
A Comparison with Deep Learning}
\label{sec:rt}

In radiation therapy treatment planning, solving an optimization problem to produce a personalized treatment that clinicians are likely to find acceptable is challenging and time-consuming. Increasingly, machine learning models are being deployed to quickly generate personalized dose distributions that satisfy the relevant clinical constraints. The current state-of-the-art uses deep neural networks to map a patient's contoured computed tomography (CT) image to a 3-D dose distribution via supervised learning~\citep{Kearney:2018aa, mahmood:2018gancer, Babier:2019aa}. 
Below, we highlight the setup and key results. Details on the clinical problem, the data and algorithm, including data augmentation, and full results are provided in Section~\ref{sec:ec_RT}.

\subsubsection{Data and Model.}

We use a data set of clinically approved treatments for 217 patients with head-and-neck cancer. 
Each patient's CT image is discretized into a tensor whose elements are voxels (3-D volumetric pixels) of the patient's geometry.
Thus, for each patient $i$, we have a 3-D dose distribution $\bhx_i \in \field{R}^{128\times128\times128}$ (representing the treatment decision) that was approved by the oncologist and a contoured CT image $\bhu_i\in \field{R}^{128\times128\times128\times8}$ (representing the context vector) that the oncologist would have used to determine whether the dose was acceptable for that patient. 
We randomly split the data into 100, 67, and 50 patients for training, validation, and hold-out testing, respectively. We then apply a data augmentation step to update the 100 doses into approximately 5,000 doses (i.e., 50 per patient) binned into feasible and infeasible training sets. 

Each patient possesses up to seven structures that each correspond to a clinic-mandated constraint.
There are four organs-at-risk (OARs)---Right Parotid, Left Parotid, Larynx, Mandible---that each possess an upper bound constraint on the dose to the structure. 
There are also three targets, called planning target volumes (PTVs)---PTV70, PTV63, and PTV56 where the number indicates the amount of radiation that at least 90\% of the structure should receive (i.e., the 90\% Value-at-Risk). Let $\set{O}$ and $\set{T}$ index the OARs and PTVs, respectively, and let $\set{R}:= \set{O} \cup \set{T}$ index all structures of interest. For each structure $r \in \set{R}$, let $\set{V}_r$ index the voxel set, which are the elements of $\bx$ and $\bu$ that correspond to structure $r$. 
Thus, the context-dependent clinical constraints for each structure are written as $g_r(\bx) \leq h_r(\bu)$ (see column 2 of Table~\ref{tab:clinCrit}),
where $h_r(\bu) \in \{\overline{h}_r, \infty\}$ is equal to the clinic-specific bound $\overline{h}_r$ if an oncologist were to require this constraint to be satisfied for patient with context $\bu$ for it to be approved, and equal to $\infty$ (i.e., no constraint) otherwise.  
Finally, our non-convex radiation therapy optimization problem minimizes the sum of average doses measured in gray (Gy) delivered to the OARs subject to patient-specific clinical constraints and a fixed set of known safety constraints on the average dose (see~\citealp{Babier:2018a}):
\begin{subequations}\label{eq:RT_problem}
\begin{align*}
    \RT(\bu):\quad\minimize{\bx}\quad  & \frac{1}{|\set{O}|} \sum_{o \in \set{O}} \frac{1}{|\set{V}_o|} \sum_{v \in \set{V}_o} x_v  \\ 
    \subjectto \quad  &  \bx \in \Xfeas_r(\bu) := \left\{ \bx \;\Bigg|\; 
        g_r (\bx) \leq h_r(\bu), \quad \underline{x}_r \leq \frac{1}{|\set{V}_r|} \sum_{v \in \set{V}_r} x_v \leq \overline{x}_r \right\} & \quad \forall r \in \set{R}. 
\end{align*}
\end{subequations}
Each $\Xfeas_r(\bu)$ comprises the patient-specific clinic constraint and upper/lower bounds ($\underline{x}_r$, $\overline{x}_r$) on the average dose to the structure if $h_r(\bu) = \infty$. These bounds are sufficiently large so that each $\Xfeas_r(\bu)$ is compact. 
Let $\Xfeas(\bu) := \cup_{r\in\set{R}} \Xfeas_r(\bu)$ and $\Xrel := \{ \bx \;|\; \underline{x}_r  \leq \frac{1}{|\set{V}_r|} \sum_{v \in \set{V}_r} x_v \leq \overline{x}_r, \; \forall r \in \set{R} \}$.

For each patient $i$ in our data set, we can evaluate the clinically approved dose $\bhx_i$ (the actual dose that was delivered to the patient) and determine which of the clinical constraints are satisfied. When we use IPMAN to train a generator to produce a dose $F(\bhu_i)$, we can determine whether $F(\bhu_i)$ is feasible if it satisfies all of the same clinical constraints as the clinically approved dose $\bhx_i$. 
As a result, the oracle $\Feasfunc$ used to label generated decisions as feasible or infeasible is a look-up table that uses on-off constraints such as in Example~\ref{ex:oracle_onoff}.

We implement IPMAN using two convolutional neural network (CNN) models. Our generator and classifier are modified from the generative adversarial network (GAN) of~\citet{Babier:2019aa}. 
We train four generators $F^{(j,k)}(\bu)$ for $\lambda_j \in \{256, 64, 16, 4\}$ for $K=11$ iterations with our training set. We compare against two state-of-the-art baseline models from the clinical automated planning literature: a U-net CNN~\citep{Nguyen:2019aa} and a GAN~\citep{Babier:2019aa}. 
The CNN is trained by minimizing the mean-squared error of predicted doses from clinical doses. 
The GAN is composed of two networks that are trained adversarially via minimax. 

\subsubsection{Experiment 1: Generating Feasible and Optimal Dose Distributions.}
\label{sec:rt_learning}

Table~\ref{tab:clinCrit} shows the percentage of dose distributions that are feasible with respect to the clinical constraints, as well as their average objective function value for IPMAN and the two baseline models on the hold-out test set. We further evaluate each model on the percentage of dose distributions that individually satisfy each of the relevant clinical constraints (first seven rows) to within a $1$ Gy relaxation, which is within acceptable limits~\citep{low1998technique}.

\begin{table}[t]
    \centering
    \caption{The percentage of dose distributions on the hold-out test set that satisfy each contextual constraint to 1 Gy relaxation. The best performing models on the summary statistics are highlighted. $\mathrm{VaR}_{90}$ refers to the Value-at-Risk at $90$\%. }
    \small
    \begin{tabular}{l c c c c c c c c} \toprule
        Structure $r$ & Constraints (Gy) &  \multicolumn{2}{c}{Baselines} & \multicolumn{4}{c}{IPMAN  ($\lambda$)} \\ \cmidrule(l){3-4} \cmidrule(l){5-8}
        ~& $g_r(\bx) \leq \overline{h}_r$ & GAN & CNN & 256 & 64 & 16 & 4  \\ \midrule
        Right Parotid & $\mathrm{mean}\{ x_v \;|\; v \in \set{V}_r \} \leq26$ & $85.7$ & $85.7$ & $86.2$ & $90.0$ & $93.3$ & $100$ \\
        Left Parotid  & $\mathrm{mean}\{ x_v \;|\; v \in \set{V}_r \} \leq26$ & $70.0$ & $60.0$ & $70.0$ & $90.0$ & $90.0$ & $100$ \\
        Larynx        & $\mathrm{mean}\{ x_v \;|\; v \in \set{V}_r \} \leq45$ & $93.3$ & $83.3$ & $89.7$ & $89.7$ & $93.3$ & $100$ \\
        Mandible      & $\mathrm{max}\{ x_v \;|\; v \in \set{V}_r \} \leq73.5$ & $100$ & $100$ & $100$ & $100$ & $100$ & $100$ \\
        PTV70         & $\mathrm{VaR}_{90}\{ x_v \;|\; v \in \set{V}_r \} \geq70$ & $97.6$  & $97.6$ & $97.6$  & $97.6$  & $95.2$ & $92.8$ \\
        PTV63         & $\mathrm{VaR}_{90}\{ x_v \;|\; v \in \set{V}_r \} \geq63$ & $96.3$  & $96.3$ & $96.3$  & $96.3$  & $96.3$ & $96.3$ \\ 
        PTV56         & $\mathrm{VaR}_{90}\{ x_v \;|\; v \in \set{V}_r \} \geq56$ & $100$  & $100$ & $100$  & $100$  & $100$ & $ 100$ \\
        \midrule
        \multicolumn{2}{l}{Fraction of feasible doses $\bx \in \left( \cap_{r} \Xfeas_r(\bu) \right)$ (\%)}  & $86.0$ & $78.0$ & $82.0$ & $88.0$ & $88.0$ & \graycell$94.0$ \\ \midrule
        \multicolumn{2}{l}{Average objective function value (Gy)}  & $40.3$ & $41.0$ & $41.0$ & $41.0$ & $40.0$ & \graycell$37.8$ \\
        \bottomrule
    \end{tabular}
    \label{tab:clinCrit}
\end{table}

In general, the IPMAN models for $\lambda \leq 64$ satisfy almost all of the relevant constraints at equal or higher rates than the baselines.
At $\lambda = 4$, IPMAN dominates both baselines in constraint satisfaction as well as the average objective function value. That is, this model predicts dose distributions that deliver lower dose to healthy tissue while better satisfying the relevant clinical constraints. Thus, it is possible for IPMAN-trained models to produce solutions that are feasible more often and exhibit lower objective function value on average, compared to existing state-of-the-art methods.

\subsubsection{Experiment 2: Adapting to the Policies of Different Institutions.}
\label{sec:rt_new_clinic}

Different clinics mandate different clinical constraints~\citep{Wu:2017aa}, meaning that a data set obtained from one clinic may not satisfy the institutional requirements at another. Further, new clinics may have low patient volumes making it difficult to obtain sufficient data to properly train generative models using conventional supervised learning techniques~\citep{sSize}.  
IPMAN can potentially address this issue by using training set data from a high-volume clinic and encoding the constraints specific to the new clinic into the oracle to adaptively learn to satisfy the new clinic's policies.

We train IPMAN to generate doses to satisfy the clinical constraints from~\citet{Geretschlager:2015aa}, who seek doses of 72 Gy, 66 Gy, and 54 Gy for the three targets in each patient, rather than the 70 Gy, 63 Gy, and 56 Gy, respectively, from our data set. 
For this experiment, we rename the target sites to PTV72, PTV66, and PTV54 and then update the clinical constraint bound $\overline{h}_r$ in our oracle. 
Although we do not know the exact preferences of oncologists at the new clinic in determining when a constraint is necessary, we assume that
oncologists at both clinics decide the necessity of a constraint in the same manner, i.e., patients for whom the PTV criteria is necessary in our original clinic would require the analogous dose bound at the new clinic.

\begin{table}[t]
    \centering
    \caption{The percentage of dose distributions on the hold-out test set that satisfy each contextual constraint of~\citet{Geretschlager:2015aa} to 1 Gy relaxation. The best performing models on the summary statistics are highlighted. $\mathrm{VaR}_{90}$ refers to the Value-at-Risk at $90$\%. 
    }
    \small
    \begin{tabular}{l c c c c c c} \toprule
        Structure & Constraint (Gy)  & Baseline & \multicolumn{4}{c}{IPMAN ($\lambda$)} \\  \cmidrule(l){4-7}
        ~& $g_r(\bx) \leq \overline{h}_r$ & GAN & 256 & 64 & 16 & 4  \\ \midrule
        
        Right Parotid   & $\mathrm{mean}\{ x_v \;|\; v \in \set{V}_r \} \leq26$     & $83.3$ & $85.7$ & $85.7$ & $100$ & $100$ \\
        Left Parotid    & $\mathrm{mean}\{ x_v \;|\; v \in \set{V}_r \} \leq26$     & $70.0$ & $50.0$ & $60.0$ & $100$ & $100$ \\
        Larynx          & $\mathrm{mean}\{ x_v \;|\; v \in \set{V}_r \} \leq45$     & $93.3$ & $86.7$ & $76.8$ & $100$ & $100$ \\
        Mandible        & $\mathrm{max}\{ x_v \;|\; v \in \set{V}_r \} \leq73.5$    & $100$  & $81.0$ & $90.5$  & $100$ & $100$ \\
        PTV72           & $\mathrm{VaR}_{90}\{ x_v \;|\; v \in \set{V}_r \} \geq72$ & $7.31$ & $95.2$  & $95.2$  & $14.3$ & $0$ \\
        PTV66           & $\mathrm{VaR}_{90}\{ x_v \;|\; v \in \set{V}_r \} \geq66$ & $77.8$ & $96.3$  & $96.3$  & $85.2$ & $0$ \\ 
        PTV54           & $\mathrm{VaR}_{90}\{ x_v \;|\; v \in \set{V}_r \} \geq54$ & $100$  & $100$  & $100$  & $96.0$ & $0$ \\       
        \midrule
        \multicolumn{2}{l}{Fraction of feasible doses $\bx \in \left( \cap_{r} \Xfeas_r(\bu) \right)$ (\%)}  & $18.3$ & $64.0$ & \graycell$66.0$ & $26.0$ & $0$ \\ \midrule
        \multicolumn{2}{l}{Average objective function value (Gy)}  & \graycell$40.3$ & $42.3$ & $42.3$ & $34.3$ & $10.0$ \\
        \bottomrule
    \end{tabular}
    \label{tab:clinCrit_followup}
\end{table}

Table~\ref{tab:clinCrit_followup} shows the performance on the hold-out test set for the generator. Here, the best model corresponds to $\lambda = 256$. In comparison to the previous experiment, a higher $\lambda$ performs better since it encourages more conservative training with respect to feasibility, which is aligned with our goal of satisfying a different set of clinical criteria. Our baseline model is the best baseline from the first experiment, namely the GAN model, which does not possess the active learning aspect of IPMAN. As a result, few plans ($18.3\%$) produced by the GAN baseline satisfy the new constraints. In particular, an unacceptably low $7.31$\% of the PTV72 and 77.8\% of the PTV66 criteria are satisfied. In contrast, IPMAN is able to learn to satisfy these new constraints at rates of 95.2\% and 96.3\%, respectively. 
As a result of learning higher doses to the PTVs, IPMAN solutions have slightly higher average objective function values. However, this trade-off would be tolerable since the primary goal of the treatment is to ensure sufficient dose to the targets.

\subsubsection{Implementation Guidelines}
These experiments use real clinical data to create a look-up table labeling oracle to guide our generators, i.e., doses that are generated during training must satisfy the same constraints as the clinically delivered ones. 
In a real implementation, oncologists would play the role of the oracle, as current practice already requires them to manually review all treatments plans and determine which ones are acceptable. 
In the second experiment, we make a mild assumption that clinicians at both clinics decide the importance of constraints in the same way. To correct for human variability between clinics, we could request clinicians at the second clinic to review our data set of treatments, report which ones they find clinically acceptable, and use the feasible decisions in the same manner as above to train IPMAN. 
Finally, we assume that all oncologists at the same clinic also have the same preferences. Here, we could use the same approach to create personalized data sets and models for each clinician. We emphasize that reviewing existing treatments (i.e., relabeling data) is significantly easier than collecting new personalized data.

\section{Discussion and Conclusion} 
\label{sec:conclusion}

This paper presents a new machine learning-based approach to solving contextual decision-making problems where a large universe of contexts affect the feasibility of decisions, a set of complicating constraints are controlled by the contexts, and there is need to quickly generate high-quality solutions. Our learning algorithm, IPMAN, consists of three components: (i) a generative model that outputs optimal decisions given an input context vector; (ii) a classification model that predicts whether a decision is feasible or infeasible and is used to train the generator to satisfy context-dependent constraints; and (iii) an active learning loop that uses the outputs of the generator
to iteratively refine both models during training. We prove that our approach admits in-sample and out-of-sample guarantees on the optimality of generated decisions and, in doing so, derive new results for interior point methods whose barrier functions are defined over a relaxation of the feasible set. We demonstrate the effectiveness of our approach by numerically studying portfolio optimization and personalized treatment design problems. As compared with predict-then-optimize baselines, IPMAN more quickly generates solutions that are of similar quality and are more robust to estimation errors in the constraint parameters. When compared with deep learning baselines, IPMAN generates better solutions that are just as fast while being robust to distribution shift.

Our methodology addresses several issues that affect the deployment of machine learning systems in environments where high-quality decisions are critical \citep[e.g.,][]{lamanna2018should}.

\textbf{The tradeoff between speed and quality.}
In many settings, safety issues may arise due to the sub-optimality of the generated decisions~\citep{bertsimas2020personalized}. At the same time, there often exists a trade-off between solution quality and decision-generation speed \citep[e.g.,][]{baeza2017quality}. 
This paper directly addresses this trade-off by developing a model that generalizes the classical $\epsilon$-optimality property of interior point methods while also permitting real-time decision generation. Not only are these guarantees necessary for large-data environments where a mapping between contexts and feasible decisions must be estimated \citep[e.g.,][]{li2021review}, but our approach learns to produce optimal solutions from past decisions that were not necessarily optimal.


\textbf{Better capturing the variation in decision-maker behavior induced by contexts.}
Automated decision-making systems can at times be inflexible. As a consequence, this may lead to a disconnect between solutions prescribed by a model and the real actions undertaken by decision-makers based on their understanding of the contexts that are guiding the instance~\citep{bendoly2011linking, bendoly2013real, deloitte2015, chan2018cherry}. 
For example, the parameters of context-dependent constraints may be sensitive to estimation errors \citep[e.g.,][]{hurley2015note}, which may cause downstream optimization models to produce drastically different solutions than what a decision-maker would implement.
To address these issues, IPMAN learns a \emph{model-free} representation of contextual feasible sets (i.e., an explicit formulation of the constraint set is not necessary) and learns to generate $\epsilon$-optimal decisions. As a result, decision-makers do not have to mathematically articulate every problem detail, yet the algorithm can still generate good solutions.
By tuning the learning parameters that control $\epsilon$, our approach naturally allows for decisions that are more likely to be feasible, which provides some degree of robustness to estimation errors.

\textbf{Including expert judgment in model development.}
The active learning component of IPMAN permits for the inclusion of expert judgement---which introduces information into the training process that is not necessarily contained in the data---when designing automated ML systems. Indeed, it is common to have decision-making problems possess valuable structural information that can be used as an aid in model development~\citep[e.g.,][]{wilder2019melding, fioretto2020predicting}. In our experiments, expert judgment comes in the form of using the structural information in the oracle to automatically evaluate the quality of decisions generated during the training process. Alternatively, one can employ a group of expert human decision-makers or platform users to perform this task. For instance, having medical practitioners evaluate the acceptability of generated treatment decisions based on their tacit preferences allows them to participate in the creation of the AI algorithm. An ancillary benefit of this human-in-the-training-loop approach is that by participating in the development of the model, practitioners may eventually be more comfortable in adopting the tool \citep{vollmer2020machine, faes2020clinician}.

\textbf{Additional issues and future extensions.} Structural information that is present may not always align with the preferences of the decision-maker. Consequently, automated labeling methods may produce synthetic data that is incorrectly classified with respect to whether the generated decision is feasible for that decision maker.
In our portfolio optimization experiments, we evaluate the effect that mislabeled data has on the quality of generated solutions. Although we show that our framework is robust to a modest degree of label noise, a potential extension may be to more formally integrate our learning approach with the deep learning literature that studies settings with noisy labels~\citep{natarajan:2013learning, li:2017learning, han:2019deep}. 
Further, our numerical experiments on cancer therapy explore the generation of treatments that meet a set of context-dependent constraints that correspond to a clinic policy that differs from the policy under which the initial data was generated. A natural extension is to investigate the extent to which known structural information in a decision-making problem can be used to guide a machine learning model to adapt to covariate or domain shift~\citep{ben:2010theory, redko:2020survey}. 
Finally, while IPMAN generates certifiably optimal decisions for a large number of contexts, subsequent research should investigate how training can be modified to produce personalized decision-making models. One such extension, for example, is to run an experiment with human decision-makers in the training loop 
to measure the concordance of the produced decisions against their preferences.




%
%
%



\bibliographystyle{informs2014} 
\bibliography{refs} 




\ECSwitch


\ECHead{Electronic Companion}

\section{Proofs of main results}
\label{sec:ec_proofs}

\begin{proof}{Proof of Theorem~\ref{thm:optimal_barrier}.}
    For notational simplicity, we use $\bxl$ and $\bx^*$ in place of $\bxl(\bu)$ and $\bx^*(\bu)$, respectively. We first prove that an optimal solution exists for any $\lambda > 0$. From the Weierstrauss Theorem, an optimal solution to $\BP(\bu, \barr, \lambda)$ exists if there is a sub-level set of the objective that is non-empty, bounded, and closed. Let $\bbx$ be a point in the interior of $\Xfeas(\bu)$; by Assumption~\ref{ass:structure_of_op}, $\bbx$ exists. 
    Furthermore, $\barr(\bx, \bu) > 0$ for all $\bx \in \Xfeas(\bu)$ implies that $\bbx$ has a finite value when input into the objective for $\BP(\bu, \barr, \lambda)$. 
    Moreover, the objective for $\BP(\bu, \barr, \lambda)$ is only finite within $\left\{ \bx\;|\;\barr(\bx, \bu) > 0 \right\} \subseteq \neighbourhood{\Xfeas(\bu)}$. Consequently the sub-level set
    \begin{align*}
        \left\{ \bx \;\Big|\; \bc^\tpose \bx - \lambda \log \barr(\bx, \bu) \leq \bc^\tpose \bbx - \lambda \log \barr(\bbx, \bu) \right\} \subseteq \neighbourhood{\Xfeas(\bu)}
    \end{align*}
    is non-empty, bounded, and closed (since the objective is continuous) and $\bxl(\bu)$ exists.
    
    Suppose we choose $\epsilon = -\lambda \log{\barr(\bx^*, \bu))}$. By definition, $0 < \barr(\bx^*,\bu) < 1$, meaning $C := -\log{\barr(\bx^*,\bu)} > 0$ is a positive constant. Let $\bxl$ be an optimal solution to $\BP(\bu, \barr, \lambda)$. We now prove that $\bc^\tpose \bxl - \epsilon < \bc^\tpose \bx^*$: 
    \begin{align*}
            \bc^\tpose \bx^* + \epsilon &= \bc^\tpose \bx^* - \lambda \log{\barr\left(\bx^*,\bu\right)} \\
            &\geq \bc^\tpose \bxl - \lambda \log{\barr\left(\bxl,\bu\right)} \\
            &> \bc^\tpose \bxl.
    \end{align*}
    The first inequality follows from the optimality of $\bxl$ for $\BP(\bu, \barr, \lambda)$ while the second inequality follows from $\log{\barr(\bxl, \bu)} < 0$, meaning that $-\lambda \log{\barr(\bxl,\bu)} > 0$. Moving $\epsilon$ to the right-hand-side gives the lower bound.

    The proof of $\bc^\tpose \bx^* < \bc^\tpose \bxl + \delta$ has two cases. If $\bxl \in \Xfeas(\bu)$, then by the optimality of $\bx^*$, we trivially satisfy $\bc^\tpose \bx^* \leq \bc^\tpose \bxl < \bc^\tpose \bxl + \delta$. If $\bxl \in \neighbourhood{\Xfeas(\bu)}\setminus\Xfeas(\bu)$, then let $\btx \in \argmin_{\bx \in \Xfeas(\bu)} \norm{\bxl - \bx}$ be the projection of $\bxl$ on $\Xfeas(\bu)$. Then,
    \begin{align*}
            \bc^\tpose \bx^* - \bc^\tpose \bxl &\leq \bc^\tpose \btx - \bc^\tpose \bxl \\
            & \leq \left| \bc^\tpose \btx - \bc^\tpose \bxl \right| \\
            & \leq \norm{\btx - \bxl} \\
            &< \delta. 
    \end{align*}
    The first inequality follows from the optimality of $\bx^*$ over $\btx$ for $\OP(\bu)$. The third inequality follows from the Cauchy-Schwartz inequality and $\norm{\bc} = 1$, while the fourth inequality follows from the fact that for $\bxl \in \neighbourhood{\Xfeas(\bu)}$, there exists $\bx \in \Xfeas(\bu)$ such that $\norm{\bx - \bxl} < \delta$ and that $\btx$ minimizes this distance. This proves the upper bound. 
\qedwhite\end{proof}
\begin{proof}{Proof of Lemma~\ref{lem:discriminator_infeas_obj}.}
    This result is an application of Urysohn's Smooth Lemma, which states that given two closed and disjoint sets $\set{A}$ and $\set{A}'$, there exists a continuous function $f(\cdot) \in [0,1]$ for which $f(\set{A}) = 1$ and $f(\set{A}') = 0$~\citep{engelking:77}. Letting $\set{A} = \support(\pfeas)$, and $\set{A}' = \support(\pinfeas)$ means there is a continuous function $B^*(\bx, \bu)$ that satisfies $B^*(\bx, \bu) = 1$ for all $(\bx, \bu) \in \support(\pfeas)$ and $B^*(\bx, \bu) = 0$ for all $(\bx, \bu) \in \support \pinfeas$. To prove that $B^*(\bx, \bu)$ is a maximum, note that every $B \in \set{B}$ satisfies $B(\bx, \bu) \in [0,1]$, meaning that $0$ is an upper bound on the optimal value. Substituting $B^*(\bx, \bu)$ into the objective of $\SFCP(\pfeas, \pinfeas)$ achieves this value.

    To prove Statement 2, consider a fixed $\bu$. First note that $B^*(\bx, \bu) = 1$ for all $\bx \in \Xfeas(\bu)$. Because $B^*(\bx, \bu)$ is continuous, we must have $\Xfeas(\bu) \subset \left\{ \bx \;|\; B^*(\bx, \bu) > 0 \right\}$. Then, note that $B^{\Xrel}(\bx) = 0$ for all $\bx \in \field{R}^n \setminus \Xrel$. Consequently,  
    \begin{align*}
        \{ \bx \;|\; B^*(\bx, \bu) B^{\Xrel}(\bx) > 0 \} \subseteq \left\{ \bx \;|\; B^{\Xrel}(\bx) > 0 \right\} = \Xrel \subseteq \neighbourhoodsub{\Xfeas(\bu)}{d_H(\Xfeas(\bu), \Xrel)}.
    \end{align*}
    Finally, although $B^*(\bx, \bu) \in [0,1]$ rather than in $[0,1)$ as per the definition of a $\delta$-barrier, this can be rectified by simply scaling the classifier by a multiplicative factor. 
\qedwhite\end{proof}
\begin{proof}{Proof of Theorem~\ref{thm:delta_epsilon_insample}.}
    Using the same argument as in Theorem~\ref{thm:optimal_barrier}, as long as $B^*(\bx, \bu)$ is a continuous function, $\bxlj(\bhu_i)$ will exist. 
    For notational simplicity, we use $\bx^*$ and $\bxlj$ in place of $\bx^*(\bhu_i)$ and $\bxlj(\bhu_i)$, respectively. By the Triangle inequality,
    \begin{align*}
        \left| \bc^\tpose F^{(j)}(\bhu_i) - \bc^\tpose \bx^* \right| &\leq \left| \bc^\tpose F^{(j)}(\bhu_i) - \bc^\tpose \bxlj \right| +  \left| \bc^\tpose \bxlj - \bc^\tpose \bx^* \right|. 
    \end{align*}
    Even if $B^{*, \Xrel}(\bx, \bhu_i)$ is not a $\delta$-barrier for $\OP(\bhu_i)$, we will prove that $\bxl$ is an optimal solution to a barrier problem $\BP(\bhu_i, \barr, \lambda_j)$ for some $\delta$-barrier, and therefore, satisfies a $(\delta, \epsilon)$-optimality bound $| \bc^\tpose \bxlj - \bc^\tpose \bx^* | < \max(\delta, \epsilon)$ for some $\delta>0$ and $\epsilon >0$. We split the proof into two separate cases: when $B^{*, \Xrel}(\bx^*, \bhu_i) > 0$ and when $B^{*, \Xrel}(\bx^*, \bhu_i) = 0$.
    
    First, if $B^{*, \Xrel}(\bx^*, \bhu_i) > 0$, we construct a new constrained problem for which $\bx^*$ is an optimal solution and show that $B^{*, \Xrel}(\bx, \bhu_i)$ is a $\delta$-barrier for the new problem. 
    Let $0 < \varepsilon \leq B^{*, \Xrel}(\bx^*, \bhu_i)$ be any sufficiently small parameter value and consider the optimization problem with a smaller feasible set than $\Xfeas(\bhu_i)$, below:
    \begin{align}\label{eq:constructed_problem}
        \min ~\left\{ \bc^\tpose \bx \;\Big|\; \bx \in \Xfeas(\bhu_i),\; B^{*, \Xrel}(\bx, \bhu_i) \geq \varepsilon \right\}
    \end{align}
    and note that because $\bx^*$ is feasible for~\eqref{eq:constructed_problem}, it is optimal. Further, $\{ \bx \;|\; B^{*, \Xrel}(\bx, \bhu_i) > 0 \}$ is a superset of the feasible set of~\eqref{eq:constructed_problem}, meaning that $B^{*, \Xrel}(\bx, \bhu_i)$ is a $\delta$-barrier for the above problem for some $\delta > 0$. Then, from Theorem~\ref{thm:optimal_barrier}, $\bxlj$, which is an optimal solution to $\BP(\bhu_i, B^{*, \Xrel}, \lambda_j)$, is $(\delta, \epsilon)$-optimal for~\eqref{eq:constructed_problem} with $\epsilon = -\lambda \log{B^{*, \Xrel}(\bx^*, \bhu_i)} > 0$. 

    Second, if $B^{*, \Xrel}(\bx^*, \bhu_i) = 0$, then $\bx^*$ does not lie in the support of the classifier. Instead, we construct a ``test'' $\delta$-barrier $\btest(\bx)$ for $\Xfeas(\bhu_i)$ and show that $\bxlj$ is also an optimal solution to $\BP(\bhu_i, \btest, \lambda_j)$, meaning that it satisfies an alternative $(\delta, \epsilon)$-optimality bound. 
    Let $\bxrel \in \argmin_{\bx} \{ \bc^\tpose \bx \;|\; \bx \in \Xrel \}$ and let $\bar{B}$ be a constant defined as follows: 
    \begin{align}
        \bar{B} = B^{*, \Xrel}(\bxlj, \bhu_i) \min \bigg\{ 1 \;,\; \exp \left[ -\frac{1}{\lambda_j}\left( \bc^\tpose\bxlj - \bc^\tpose \bxrel \right)  \right] \bigg\}. \label{eq:delta_epsilon_insample_test_B}
    \end{align}
    Note that $\bar{B} \in (0, 1)$. We now define $\btest(\bx)$ as a continuous function that satisfies:
    \begin{align*}
        \btest(\bx) = 
        \begin{cases}
            B^{*, \Xrel}(\bx,\bhu_i)    & \forall \bx \in \{ \bx \;|\; B^{*, \Xrel}(\bx, \bhu_i) \geq \bar{B} \} \\
            \bar{B}                & \forall \bx \in \{ \bx \;|\; B^{*, \Xrel}(\bx, \bhu_i) < \bar{B}, \; \bx \in \Xfeas(\bhu_i) \} \\
            \leq \bar{B}           & \forall \bx \in \{ \bx \;|\; B^{*, \Xrel}(\bx, \bhu_i) < \bar{B}, \; \bx \in \Xrel \setminus \Xfeas(\bhu_i)  \} \\
            0                      & \forall \bx \in \field{R}^n \setminus \Xrel.
        \end{cases}
    \end{align*}
    The third condition is left as an inequality since we only need a continuous function $\btest(\bx)$ to satisfy an inequality in that range. We argue that such a function must exist. Because $B^{*,\Xrel}(\bx, \bhu_i)$ is continuous in $\bx$, the first region $\{ \bx \;|\; \btest(\bx) \geq \bar{B} \}$ is closed. Furthermore because the the support of $B^{*, \Xrel}(\bx, \bhu_i)$ is a subset of $\Xrel$, the first region is disjoint from $\field{R}^n \setminus \Xrel$. By Urysohn's Smooth Lemma, a continuous function within $[0,1]$ that is equal to $1$ for $\bx \in \{ \bx \;|\; \btest(\bx) \geq \bar{B} \}$ and $0$ for $\bx \in \field{R}^n \setminus \Xrel$ exists; scaling this function gives $\btest(\bx)$.
    
    The support of $\btest(\bx)$ is a superset of $\Xfeas(\bhu_i)$ and $\btest(\bx)$ is in the range $[0,1)$, meaning that it is a $\delta$-barrier for $\OP(\bhu_i)$ for some $\delta > 0$. It only remains to prove that $\bxlj$ is an optimal solution for $\BP(\bhu_i, \btest, \lambda_j)$, i.e.,
    \begin{align*}
        \bc^\tpose \bxlj - \lambda_j \log \btest(\bxlj) \leq \bc^\tpose \bx - \lambda_j \log \btest(\bx), \quad \forall \bx \in \{ \bx \;|\; \btest(\bx) > 0 \}.
    \end{align*}
    First, consider the region $\{ \bx \;|\; \btest(\bx) \geq \bar{B} \}$. From~\eqref{eq:delta_epsilon_insample_test_B}, $\bar{B} \leq B^{*, \Xrel}(\bxlj, \bhu_i)$ meaning $\btest(\bxlj) = B^{*, \Xrel}(\bxlj, \bhu_i)$. Because $\bxlj$ is optimal for $\BP(\bhu_i, B^{*, \Xrel}, \lambda_j)$ and this region is a subset of the support of $B^{*, \Xrel}(\bx, \bhu_i)$, we have
    \begin{align*}
        \bc^\tpose \bxlj - \lambda_j \log \btest(\bxlj) \leq  \bc^\tpose \bx - \lambda_j \log \btest(\bx) \quad \forall \bx \in \{ \bx \;|\; \btest(\bx) \geq \bar{B} \}.
    \end{align*}
    
    Now, consider the region $\{ \bx \;|\; \btest(\bx) < \bar{B} \}$. Replacing the minimum in~\eqref{eq:delta_epsilon_insample_test_B} with an inequality and taking the logarithm on both sides yields
    \begin{align*}
        -\log \bar{B} &\geq -\log B^{*, \Xrel}(\bxlj, \bhu_i) + \frac{1}{\lambda_j} \left( \bc^\tpose\bxlj - \bc^\tpose \bxrel \right).
    \end{align*}
    We further re-arrange this inequality to
    \begin{align}
        \bc^\tpose\bxlj - \lambda_j  \log B^{*, \Xrel}(\bxlj, \bhu_i) &\leq  \bc^\tpose \bxrel - \lambda_j  \log \bar{B}  \label{eq:proof_delta_epsilon_insample1} \\
        \bc^\tpose\bxlj - \lambda_j \log \btest(\bxlj)          &\leq \bc^\tpose \bx - \lambda_j \log \bar{B} & \forall \bx \in \{ \bx \;|\; \btest(\bx) < \bar{B} \} \label{eq:proof_delta_epsilon_insample2} \\
        \bc^\tpose\bxlj - \lambda_j \log \btest(\bxlj)          &\leq \bc^\tpose \bx - \lambda_j  \log \btest(\bx) & \forall \bx \in \{ \bx \;|\; \btest(\bx) < \bar{B} \} \label{eq:proof_delta_epsilon_insample3} 
    \end{align}
    We obtain~\eqref{eq:proof_delta_epsilon_insample2} by substituting $\btest(\bxlj) = B^{*, \Xrel}(\bxlj, \bhu_i)$ and noting $\bc^\tpose \bxrel \leq \bc^\tpose \bx$ for all $\bx \in \subset \Xrel$. We obtain~\eqref{eq:proof_delta_epsilon_insample3} because $\btest(\bx) < \bar{B}$. Thus, $\bxlj$ is an optimal solution to $\BP(\bhu_j, \btest, \lambda)$ and is $(\delta, \epsilon)$-optimal for $\OP(\bhu_i)$ where $\epsilon = -\lambda_j \log \bar{B}$.
\qedwhite\end{proof}
\begin{proof}{Proof of Corollary~\ref{cor:lb_on_delta}.}
    The $\delta$ term in the bound of~\eqref{eq:delta_epsilon_insample} is derived from the fact that an optimal solution $\bxlj(\bhu_i)$ to $\BP(\bhu_i, B^{*, \Xrel}, \lambda_j)$ is also an optimal solution to some barrier problem with a $\delta$-barrier. Just as the proof of Theorem~\ref{thm:delta_epsilon_insample} invoked two separate cases with two different bounds, i.e., $\delta$ takes a different value if $B^{*, \Xrel}(\bx^*, \bhu_i) > 0$ or if $B^{*, \Xrel}(\bx^*, \bhu_i) = 0$, we consider each case separately and show that $\delta$ can be bounded by the above in both cases.

    If $B^{*, \Xrel}(\bx^*, \bhu_i) > 0$, then the proof of Theorem~\ref{thm:delta_epsilon_insample} followed by constructing a new optimization problem~\eqref{eq:constructed_problem} with a smaller feasible set and showing that the product classifier is a $\delta$-barrier for that problem, where $\delta$ is equal to the Hausdorff distance between this feasible set and the support of the product classifier 
    \begin{align*}
        \delta = d_H\left( \left\{ \bx \;\big|\; \bx \in \Xfeas(\bhu_i),\; B^{*, \Xrel}(\bx, \bhu_i) \geq \varepsilon \right\},\; \left\{ \bx \;\big|\; B^{*, \Xrel}(\bx, \bhu_i) > 0 \right\} \right).
    \end{align*}
    Because we can choose any sufficiently small value for the $\varepsilon$ parameter, we select a value such that $B^{*, \Xrel}(\bhx, \bhu_i) \geq \varepsilon$ for all $(\bhx, \bhu_i) \in \dataset$. Furthermore, all $\bhx$ in this data set are feasible decisions to $\OP(\bhu_i)$. Thus, we have $\left\{ \bx \;\big|\; \bx \in \Xfeas(\bhu_i),\; B^{*, \Xrel}(\bx, \bhu_i) \geq \varepsilon \right\} \supset \left\{ \bhx \;|\; (\bhx, \bhu_i) \in \dataset \right\}$. Finally, $\left\{ \bx \;\big|\; B^{*, \Xrel}(\bx, \bhu_i) > 0 \right\} \subset \Xrel$. Substituting the subset and superset into the definition of the Hausdorff distance yields the upper bound on $\delta$ for this case.

    If $B^{*, \Xrel}(\bx^*, \bhu_i) = 0$, then the proof of Theorem~\ref{thm:delta_epsilon_insample} followed by constructing a new barrier function $\btest(\bx)$ for $\OP(\bhu_i)$. Then, $\delta = d_H\left( \Xfeas(\bhu_i), \; \left\{ \bx \;|\; \btest(\bx) > 0 \right\} \right)$. However, again note that $\Xfeas(\bhu_i) \supset \left\{ \bhx \;|\; (\bhx, \bhu_i) \in \dataset \right\}$ and that $\left\{ \bx \;|\; \btest(\bx) > 0 \right\} \subseteq \Xrel$. Substituting these two sets into the definition of the Hausdorff distance yields the same upper bound for this case.
\qedwhite\end{proof}

\begin{proof}{Proof of Proposition~\ref{propn:shrinking_optimal_set}.}
    When $\set{B}$ is unrestricted, Lemma~\ref{lem:discriminator_infeas_obj} states that for any $k$, the optimal value of $\FCP(\dataset^{(k)}, \datasetinfeas^{(k)})$ is $0$ and can be achieved when the optimal solution satisfies $B^{(k)}(\bhx, \bhu) = 1$ for all $(\bhx, \bhu) \in \dataset^{(k)}$ and $B^{(k)}(\bhx, \bhu) = 0$ for all $(\bbx, \bhu) \in \datasetinfeas^{(k)}$.

    In the $(k+1)^{\textrm{st}}$ iteration, $\dataset^{(k+1)} = \dataset^{(k)} \cup \set{Q}$ and $\datasetinfeas^{(k+1)} = \datasetinfeas^{(k)} \cup \bar{\set{Q}}$ where $\set{Q}$ and $\bar{\set{Q}}$ are defined as in Algorithm~\ref{alg:ipman_summary}. To show that $\set{B}^{(k+1)} \subset \set{B}^{(k)}$, we first prove $\set{B}^{(k+1)} \subseteq \set{B}^{(k)}$ and then present a counter-example which disproves the equivalence.

    The objective function of $\FCP(\dataset^{(k+1)}, \datasetinfeas^{(k+1)})$ is
    \begin{align*}
        &\frac{1}{|\dataset^{(k+1)}|} \sum_{(\bhx, \bhu) \in \dataset^{(k+1)}} \log{B(\bhx, \bhu)} + \frac{1}{|\datasetinfeas^{(k+1)}|} \sum_{(\bbx, \bhu) \in \datasetinfeas^{(k+1)}} \log{\big(1 - B(\bbx, \bhu)\big)} \\
        &\hspace{6em}= \frac{\alpha}{|\dataset^{(k)}|} \sum_{(\bhx, \bhu) \in \dataset^{(k)}} \log{B(\bhx, \bhu)} + \frac{1-\alpha}{|\set{Q}|} \sum_{(\bhx, \bhu) \in \set{Q}} \log{B(\bhx, \bhu)}
        \\&\hspace{6em}\quad+ \frac{\alpha'}{|\datasetinfeas^{(k)}|} \sum_{(\bbx, \bhu) \in \datasetinfeas^{(k)}} \log{\big(1 - B(\bbx, \bhu)\big)} + \frac{1-\alpha'}{|\bar{\set{Q}}|} \sum_{(\bbx, \bhu) \in \bar{\set{Q}}} \log{\big(1 - B(\bbx, \bhu)\big)},
    \end{align*}
    where $\alpha = |\dataset^{(k)}|/|\dataset^{(k+1)}|$ and $\alpha' = |\datasetinfeas^{(k)}|/|\datasetinfeas^{(k+1)}|$ are the mixture weights defining the ratio of existing to new points in each data set. Because the optimal value of $\FCP(\dataset^{(k+1)}, \datasetinfeas^{(k+1)})$ is $0$ and $B(\bx,\bu) \in [0, 1]$, each of the individual terms must be equal to $0$ for an optimal solution. However, the first and third terms define the objective function for $\FCP(\dataset^{(k)}, \datasetinfeas^{(k)})$. Thus, any optimal solution $B^{(k+1)}$ to $\FCP(\dataset^{(k+1)}, \datasetinfeas^{(k+1)})$ must also be optimal for $\FCP(\dataset^{(k)}, \datasetinfeas^{(k)})$ implying $\set{B}^{(k+1)} \subseteq \set{B}^{(k)}$.

    To prove the inclusion is strict, consider the sets $\dataset^{(k)} \cup \bar{\set{Q}}$ and $\datasetinfeas^{(k)}$. We can define a function $B^*(\bx,\bu)$ such that $B^*(\bhx, \bhu) = 1$ for all $(\bhx,\bhu) \in \dataset^{(k)} \cup \bar{\set{Q}}$ and $B^*(\bbx,\bhu) = 0$ for all $(\bbx,\bhu) \in \datasetinfeas^{(k)}$, i.e., $B^* \in \set{B}^{(k)}$. However, then $B^*(\bbx, \bhu) = 1$ for all $(\bbx, \bhu) \in \bar{\set{Q}}$ and $B^*(\bx, \bu)$ has an infinite objective function value for $\FCP(\dataset^{(k+1)}, \datasetinfeas^{(k+1)})$. Thus, $B^* \notin \set{B}^{(k+1)}$.
\qedwhite\end{proof}

\section{Structural properties of $(\delta, \epsilon)$-optimality for the barrier problem}
\label{sec:ipman_delta_algorithm}

Our learning problem simultaneously trains a classifier and a generative model to learn feasibility and predict optimal solutions respectively. Alternatively, if we are already given a $\delta$-barrier $\barr(\bx,\bu)$, we may consider directly optimizing $\BP(\bu, \barr, \lambda)$. We show how tuning the $\lambda$ parameter can yield feasible or infeasible solutions of different qualities.

Under a mild regularity assumption, for a sufficiently large $\lambda$, an optimal solution $\bxl(\bu)$ to $\BP(\bu, \barr, \lambda)$ is guaranteed to lie inside $\Xfeas(\bu)$. Once $\lambda$ is sufficiently small, the optimal solutions then enter $\neighbourhood{\Xfeas(\bu)} \setminus \Xfeas(\bu)$.
We first state this assumption before characterizing the trajectory of the sequence of points obtained via an IPM.
\begin{assumption}[Regularity of the $\delta$-barrier]\label{ass:D_feasible_infeasible}~%
        \begin{enumerate}
                \item There exist $\btx \in \interior(\Xfeas(\bu))$ such that $\barr(\btx,\bu) > \barr(\bx,\bu)$ for all $\bx \in \closure(\neighbourhood{\Xfeas(\bu)}\setminus\Xfeas(\bu))$.

                \item There exist $\btx' \in \neighbourhood{\Xfeas(\bu)}\setminus\Xfeas(\bu)$ such that $\bc^\tpose \btx' < \bc^\tpose \bx^*(\bu)$ and $0 < \barr(\btx',\bu) < \barr(\bx,\bu)$ for all $\bx \in \Xfeas(\bu)$.
        \end{enumerate}
\end{assumption}
The first statement implies that there exists a point inside $\Xfeas(\bu)$ for which $\barr(\bx,\bu)$ is greater than any point outside of $\Xfeas(\bu)$. Similarly, the second statement implies that there exists a point outside of $\Xfeas(\bu)$ for which $\barr(\bx,\bu)$ is lower than any point inside $\Xfeas(\bu)$. Intuitively, the barrier yields higher values for points inside $\Xfeas(\bu)$ rather than outside. Furthermore, the existence of $\btx'$ for which $\bc^\tpose \btx >  \bc^\tpose \bx^*(\bu) > \bc^\tpose \btx'$ is a direct consequence of the linear objective.
Figure~\ref{fig:delta_barrier2} shows an example of such points for a feasible set where the $\delta$-barrier is a canonical barrier for $\Xrel$. Given a barrier function satisfying Assumption~\ref{ass:D_feasible_infeasible}, $\lambda$ controls the feasibility of $\bxl(\bu)$ for $\OP(\bu)$.
\begin{lemma}
        \label{lem:barr_has_feasible_optimal}
        If Assumption~\ref{ass:D_feasible_infeasible} is satisfied, then there exists $\lambdau$ such that for all $\lambda \geq \lambdau$, the optimal solution to $\BP(\bu, \barr, \lambda)$ is feasible for $\OP(\bu)$, i.e., $\bxl(\bu) \in \Xfeas(\bu)$.
\end{lemma}
\begin{proof}{Proof of Lemma~\ref{lem:barr_has_feasible_optimal}.}
        Let $\bx^+ \in \argsup_{\bx} \left\{ \barr(\bx,\bu)\;|\;\bx \in \neighbourhood{\Xfeas(\bu)}\setminus\Xfeas(\bu) \right\}$ and $\bx^- \in \arginf_{\bx} \left\{ \bc^\tpose \bx \;|\;\barr(\bx,\bu) > 0 \right\}$. Then, for $\btx$ satisfying Assumption~\ref{ass:D_feasible_infeasible} Statement 1, we set
        \begin{align} \label{eq:lambda_barr_has_feasible_optimal}
                \lambdau = \frac{\bc^\tpose \btx - \bc^\tpose \bx^-}{\log{\barr(\btx,\bu)} - \log{\barr(\bx^+, \bu)}}.
        \end{align}
        From the optimality of $\bx^-$, we have $\bc^\tpose \btx > \bc^\tpose \bx^-$. Also, Assumption~\ref{ass:D_feasible_infeasible} implies that the denominator is positive, and therefore $\lambdau > 0$. Rearranging~\eqref{eq:lambda_barr_has_feasible_optimal} yields 
        \begin{align*}
                \bc^\tpose \btx - \lambdau\log{\barr(\btx, \bu)} &= \bc^\tpose \bx^- - \lambdau\log{\barr(\bx^+, \bu)}.
        \end{align*}
        By optimality of $\bx^+$ and $\bx^-$, we have $\bc^\tpose \bx \geq \bc^\tpose \bx^-$ and $\log{\barr(\bx,\bu)} \leq \log{\barr(\bx^+, \bu)}$ respectively, for all $\bx \in \neighbourhood{\Xfeas(\bu)}\setminus\Xfeas(\bu)$. Therefore, $\bc^\tpose \btx - \lambdau\log{\barr(\btx,\bu)} \leq \bc^\tpose \bx - \lambdau\log{\barr(\bx,\bu)}$ for all $\bx \in \neighbourhood{\Xfeas(\bu)}\setminus\Xfeas(\bu)$, concluding that the optimal solution to $\BP(\bu, \barr, \lambdau)$ must satisfy $\bx^{\lambdau}(\bu) \in \Xfeas(\bu)$.

        Now for any $\varepsilon > 0$, observe that
        \begin{align*}
                \bc^\tpose \btx - (\lambdau+ \varepsilon) \log{\barr(\btx,\bu)} &\leq \bc^\tpose \bx - \lambdau\log{\barr(\bx,\bu)} - \varepsilon \log{\barr(\btx,\bu)} \quad\quad \forall \bx \in \neighbourhood{\Xfeas(\bu)}\setminus\Xfeas(\bu) \\
                &< \bc^\tpose \bx - \lambdau\log{\barr(\bx,\bu)} - \varepsilon \log{\barr(\bx,\bu)} \quad\quad \forall \bx \in \neighbourhood{\Xfeas(\bu)}\setminus\Xfeas(\bu).
        \end{align*}
        The first line is obtained by adding $\varepsilon \log{\barr(\btx,\bu)}$ to both sides, and the second from $\barr(\btx,\bu) > \barr(\bx,\bu)$ for $\bx \in \neighbourhood{\Xfeas(\bu)}\setminus\Xfeas(\bu)$. Thus, $\BP(\bu, \barr, \lambdau+\varepsilon)$ yields feasible solutions to $\OP(\bu)$.
\qedwhite\end{proof}
\begin{lemma}
        \label{lem:barr_has_infeasible_optimal}
        If Assumption~\ref{ass:D_feasible_infeasible} is satisfied, then there exists $\lambdal$ such that for all $\lambda \leq \lambdal$, the optimal solution to $\BP(\bu, \barr, \lambda)$ is infeasible for $\OP(\bu)$, i.e., $\bxl(\bu) \in \neighbourhood{\Xfeas(\bu)}\setminus\Xfeas(\bu)$.
\end{lemma}
\begin{proof}{Proof of Lemma~\ref{lem:barr_has_infeasible_optimal}.}
        Let $\bx^\dagger \in \argmax_{\bx} \left\{ \barr(\bx,\bu)\;|\;\bx \in \Xfeas(\bu) \right\}$. Then, for $\btx'$ satisfying Assumption~\ref{ass:D_feasible_infeasible} Statement 2, let
        \begin{align} \label{eq:lambda_barr_has_infeasible_optimal}
                \lambdal = \frac{\bc^\tpose\bx^*(\bu) - \bc^\tpose\btx'}{\log{\barr(\bx^\dagger,\bu)} - \log{\barr(\btx',\bu)}}.
        \end{align}
        Assumption~\ref{ass:D_feasible_infeasible} Statement 2 ensures $\bc^\tpose\bx^*(\bu) > \bc^\tpose \btx'$ and $\log{\barr(\bx^\dagger,\bu)} > \log{\barr(\btx',\bu)}$. Therefore, $\lambdal> 0$.  Rearranging~\eqref{eq:lambda_barr_has_infeasible_optimal} gives us
        \begin{align*}
                \bc^\tpose \btx' - \lambdal\log{\barr(\btx',\bu)} &= \bc^\tpose\bx^*(\bu) - \lambdal\log{\barr(\bx^\dagger, \bu)}.
        \end{align*}
        By optimality of $\bx^*(\bu)$ and $\bx^\dagger$, we have $\bc^\tpose \bx \geq \bc^\tpose \bx^*(\bu)$ and $\log{\barr(\bx,\bu)} \leq \log{\barr(\bx^\dagger, \bu)}$ respectively, for all $\bx \in \Xfeas(\bu)$. Therefore $\bc^\tpose \btx' - \lambdal\log{\barr(\btx',\bu)} \leq \bc^\tpose \bx - \lambdal \log{\barr(\bx,\bu)}$ for all $\bx \in \Xfeas(\bu)$, concluding that the optimal solution to $\BP(\bu, \barr, \lambdal)$ must satisfy $\bx^{\lambdal}(\bu) \in \neighbourhood{\Xfeas(\bu)}\setminus\Xfeas(\bu)$.

        Now for any $\varepsilon > 0$, observe that
        \begin{align*}
                \bc^\tpose \btx' - (\lambdal- \varepsilon) \log{\barr(\btx',\bu)} &\leq \bc^\tpose \bx  - \lambdal\log{\barr(\bx,\bu)} + \varepsilon \log{\barr(\btx',\bu)} \quad\quad \forall \bx \in \Xfeas(\bu) \\
                & < \bc^\tpose\bx - \lambdal\log{\barr(\bx,\bu)} + \varepsilon \log{\barr(\bx,\bu)} \quad\quad \forall \bx \in \Xfeas(\bu).
        \end{align*}
        The first line is obtained by subtracting $\varepsilon \log{\barr(\btx', \bu)}$ to both sides, and the second from $\barr(\btx',\bu) < \barr(\bx,\bu)$ for all $\bx \in \Xfeas(\bu)$. Thus, $\BP(\bu, \barr, \lambdal- \varepsilon)$ yields infeasible solutions to $\OP(\bu)$.
\qedwhite\end{proof}

\begin{figure}[t]
    \centering
    \resizebox{0.4\textwidth}{!}{%
        \begin{tikzpicture}
                \path [draw, ultra thick] (4, 0) -- (1.26, 3.79) -- (-3.20, 2.39) -- (-3.28, -2.28) -- (1.13, -3.83) -- (4, 0);
                \draw [thick, dotted, fill=white, fill opacity=0, scale=0.8] plot [smooth cycle] coordinates{(4, 0) (1.26, 3.79) (-3.2, 2.39) (-3.28, -2.28) (1.13, -3.83)};
                \draw [thick, dotted, fill=white, fill opacity=0, scale=0.57] plot [smooth cycle] coordinates{(4, 0) (1.26, 3.79) (-3.2, 2.39) (-3.28, -2.28) (1.13, -3.83)};
                \draw [thick, dotted, fill=white, fill opacity=0, scale=0.2] plot [smooth cycle] coordinates{(4, 0) (1.26, 3.79) (-3.2, 2.39) (-3.28, -2.28) (1.13, -3.83)};
                \draw [ultra thick, dashed, fill=black!30, fill opacity=0.5] plot [smooth cycle] coordinates{(2.2, 0) (1.8, 1.5) (1.0, 2.5) (-1, 1) (-2.5, 0) (-1, -2) (1.5, -1.5)};
                \draw [->, ultra thick] (3.0, -2.6) -- (3.9, -3.5) node[midway, label=right:{$\bc$}] {};
                \draw (-2.5, 0) node[datapoint, label=right:$\bx^*(\bu)$] (xstar) {};
                \draw (-0.62,0.52) node[datapoint, label=right:$\btx$] (bbx) {};
                \draw (-2.64,1.8) node[datapoint, label=right:$\btx'$] (btx) {};

                \draw (3, 2.5) node (xrel) {$\Xrel$};
                \draw (1.7, 0) node (xfeas) {$\Xfeas(\bu)$};
        \end{tikzpicture}
    }
    \caption{The canonical barrier $B^{\Xrel}(\bx)$ where the dotted lines are level sets. $\bx^*(\bu)$ is optimal for $\OP(\bu)$ while $\btx$ and $\btx'$ satisfy Lemmas~\ref{lem:barr_has_feasible_optimal} and~\ref{lem:barr_has_infeasible_optimal} respectively.}
    \label{fig:delta_barrier2}
\end{figure}

Lemma~\ref{lem:barr_has_infeasible_optimal} and Assumption~\ref{ass:D_feasible_infeasible} explore the case where the barrier problem produces undesirable results. Otherwise, if $\bc^\tpose \btx' > \bc^\tpose \bx^*(\bu)$ and $\barr(\btx', \bu) \geq \barr(\bx, \bu)$ for all $\btx' \in \neighbourhood{\Xfeas(\bu)}\setminus\Xfeas(\bu)$ and $\bx \in \Xfeas(\bu)$, $\OP(\bu)$ could be solved by classical IPMs.

Lemmas~\ref{lem:barr_has_feasible_optimal} and~\ref{lem:barr_has_infeasible_optimal} state that when $\lambda$ is set sufficiently high (or low), the corresponding optimal solution $\bxl(\bu)$ is a certifiably feasible (or infeasible) solution to $\OP(\bu)$. Furthermore, there exists a trajectory, i.e., feasibility (or infeasibility) is guaranteed for all $\lambda$ sufficiently high (or low).
Assuming access to an oracle $\Feasfunc$, we con construct a simple IPM (see Algorithm~\ref{alg:delta_barrier_ipm}) to obtain optimal solutions to $\OP(\bu)$. We initialize with a large $\lambda_0$ that satisfies Lemma~\ref{lem:barr_has_feasible_optimal}. We define a decay rate $\nu < 1$ and a number of iterations $j \in {0, \dots, J}$. Then, for each $j$, we simply let $\lambda_j = \lambda_0 \nu^{j}$ and solve $\BP(\bu, \barr, \lambda_j)$ to obtain a new $(\delta, \epsilon)$-optimal solution in each iteration. At the end of each iteration, the oracle checks if the solution is still feasible, and terminates when the solution exits the feasible set. 

\begin{algorithm}[t]
    \caption{Interior Point Method with a $\delta$-barrier}
    \label{alg:delta_barrier_ipm}
    \begin{algorithmic}[1]
            \renewcommand{\algorithmicrequire}{\textbf{Input:}}
            \renewcommand{\algorithmicensure}{\textbf{Output:}}
            \REQUIRE $\delta$-barrier $\barr(\bx, \bu)$; Initial dual variable $\lambda_0$ and decay rate $\nu < 1$; Oracle $\Feasfunc$.
            \ENSURE Optimal solution $\bxl(\bu)$ to the barrier problem. 
            \FOR{$j = 0$ \TO $\Niter$}
                \STATE Solve $\BP(\bu, \barr, \lambda_{j})$ to obtain optimal solution $\bx^{\lambda_{j}}(\bu)$.
                \IF{$\Psi(\bx^{\lambda_{j}}(\bu), \bu) = 0$}
                    \RETURN Previous optimal solution $\bx^{\lambda_{j-1}}(\bu)$.
                \ENDIF
            \ENDFOR
    \end{algorithmic}
\end{algorithm}

Recall that we always have access to the canonical barrier $B^{\Xrel}(\bx)$ and therefore, we only consider any $\delta$-barrier where $\delta \leq d_H(\Xfeas(\bu), \Xrel)$. We prove the optimality bound for solutions obtained via Algorithm~\ref{alg:delta_barrier_ipm}.

\begin{proposition}
        \label{propn:initialization_delta_barrier_ipm}
        Consider a $\delta$-barrier where $\delta \leq d_H(\Xfeas(\bu), \Xrel)$. Suppose that $\btx_1, \btx_2 \in \Xfeas(\bu)$ and $\btx'_1, \btx'_2 \in \neighbourhood{\Xfeas(\bu)}\setminus\Xfeas(\bu)$ satisfy Statements 1 and 2 of Assumption~\ref{ass:D_feasible_infeasible}, respectively. Assume without loss of generality $\barr(\btx_1,\bu) > \barr(\btx_2,\bu)$ and $\bc^\tpose \btx'_1 > \bc^\tpose \btx'_2$. Let $\bxrel \in \argmin_{\bx} \left\{ \bc^\tpose \bx \;|\; \bx \in \Xrel \right\}$. For $\Niter > 0$ and $j \in \{0,\dots,J\}$, consider
        \begin{align*}
                \lambda_0 = \frac{\bc^\tpose \btx_1 - \bc^\tpose \bxrel}{\log{\barr(\btx_1, \bu)} - \log{\barr(\btx_2, \bu)}}, \quad \nu = {\left( \frac{\bc^\tpose \btx'_1 - \bc^\tpose \btx'_2}{-\lambda_0 \log{\barr(\btx'_1, \bu)}} \right)}^{1/\Niter}, \quad \lambda_j = \lambda_0 \nu^j
        \end{align*}
        Then, the following statements are true:
        \begin{enumerate}
                \item An optimal solution $\bx^{\lambda_0}(\bu)$ to $\BP(\bu, \barr, \lambda_0)$ is a feasible solution for $\OP(\bu)$.

                \item There exists $1 \leq j^* \leq \Niter$ such that for all $j < j^*$, an optimal solution $\bx^{\lambda_j}(\bu)$ to $\BP(\bu, \barr, \lambda_j)$ is feasible for $\OP(\bu)$ and for all $j \geq  j^*$, $\bx^{\lambda_{j}}(\bu)$ is infeasible for $\OP(\bu)$.

                \item For any $j < j^*$, an optimal solution $\bx^{\lambda_{j}}(\bu)$ is $(0, \epsilon_j)$-optimal for $\OP(\bu)$ where
                        \begin{align*}
                                \epsilon_j =  \left(\bc^\tpose \btx'_1 - \bc^\tpose \btx'_2 \right) \nu^{j-\Niter}.
                        \end{align*}
                        Further, for any $j \geq j^*$, $\bx^{\lambda_{j}}(\bu)$ is $(\delta, \epsilon_j)$-optimal for $\OP(\bu)$, where $\delta \leq d_H(\Xfeas(\bu), \Xrel)$.
        \end{enumerate}
\end{proposition}
\begin{proof}{Proof of Proposition~\ref{propn:initialization_delta_barrier_ipm}.}
        We first make several observations about the parameters. Note that because $\Xfeas(\bu) \subset \Xrel$ relaxes the feasible set, we have $\bc^\tpose \bxrel \leq \bc^\tpose \bx^*(\bu)$. Next for all $j \leq J$, $\lambda_j = \lambda_0 \nu^j$ and specifically $\lambda_{\Niter} = \lambda_0 \nu^J = -(\bc^\tpose \btx'_1 - \bc^\tpose \btx'_2)/\log{\barr(\btx'_1, \bu)}$.

        To prove the first statement, we show that $\lambda_0 > \lambdau$ where $\lambdau$ is defined as in~\eqref{eq:lambda_barr_has_feasible_optimal} and constructed using $\btx_1$. Note that $\bc^\tpose \bxrel \leq \bc^\tpose \bx^-$ and by Assumption~\ref{ass:D_feasible_infeasible}, $\log{\barr(\btx_2,\bu)} > \log{\barr(\bx^+,\bu)}$. We substitute $\bc^\tpose \bxrel$ and $\log{\barr(\btx_2,\bu)}$ in $\lambda_0$ and prove $\lambda_0 > \lambdau$. By Lemma~\ref{lem:barr_has_feasible_optimal}, Statement 1 must hold.

        We use a similar argument using $\btx'_1$ to show $\lambda_{\Niter} < \lambdal$ as defined in~\eqref{eq:lambda_barr_has_infeasible_optimal}. By Lemma~\ref{lem:barr_has_infeasible_optimal}, an optimal solution $\bx^{\lambda_{\Niter}}$ must be infeasible for $\OP(\bu)$. Given that $\lambda_j$ decreases every iteration and using the first statement, there must exist a cutoff point $1 \leq j^* \leq J$ for which $\lambda_{j^*} < \lambdal$ and $\lambda_{j^*-1} \geq \lambdal$. Therefore, Statement 2 must also hold.

        In order to prove the third statement, recall that $\delta \leq d_H(\Xfeas(\bhu), \Xrel)$ for all $j$. We first prove $(\Delta(\bu),\epsilon_j)$-optimality when $j=\Niter$, and then prove for $j < \Niter$. Let $\epsilon_{\Niter} = \bc^\tpose \btx'_1 - \bc^\tpose \btx'_2$. Note that
        \begin{align*}
                \lambda_\Niter = \frac{\bc^\tpose \btx'_1 - \bc^\tpose \btx'_2}{-\log{\barr(\btx'_1,\bu)}} = \frac{\epsilon_\Niter}{-\log{\barr(\btx'_1,\bu)}} < \frac{\epsilon_\Niter}{-\log{\barr(\bx^*,\bu)}}.
        \end{align*}
        The second equality follows from substituting the value of $\epsilon_\Niter$ and the inequality from $\barr(\btx'_1,\bu) < \barr(\bx^*(\bu),\bu)$ (i.e., Assumption~\ref{ass:D_feasible_infeasible}). We next show that $\bx^{\lambda_{\Niter}}$ satisfies  $(\Delta(\bu), \epsilon_\Niter)$-optimality,
        \begin{align*}
                \bc^\tpose \bx^*(\bu) + \epsilon_{\Niter} &> \bc^\tpose \bx^*(\bu) - \lambda_{\Niter} \log{\barr\left(\bx^*(\bu), \bu\right)} \\
                &\geq \bc^\tpose \bx^{\lambda_{\Niter}}(\bu) - \lambda_{\Niter} \log{\barr\left(\bx^{\lambda_{\Niter}}(\bu), \bu\right)} \\
                &> \bc^\tpose \bx^{\lambda_{\Niter}}(\bu).
        \end{align*}
        The first line follows from substituting the value of $\epsilon_{\Niter}$ and the second from the optimality of $\bx^{\lambda_{\Niter}}(\bu)$ for $\BP(\bu, \barr, \lambda_{\Niter})$. The third line follows from $-\lambda_\Niter \log{\barr(\bx^{\lambda_{\Niter}}(\bu),\bu)} > 0$.

        For each $j < J$, we have $\lambda_j = \lambda_{\Niter} \nu^{j - J}$. Then, we write $\epsilon_j = (\bc^\tpose \btx'_1 - \bc^\tpose\btx'_2) \nu^{j-J}$. The same steps used for the $j=J$ case are repeated to obtain $(\Delta(\bu), \epsilon_j)$-optimality certificates. Finally, note that from Statement 2, for all $j < j^*$, the optimal solutions $\bx^{\lambda_j}(\bu)$ are feasible for $\OP(\bu)$. By optimality of $\bx^*(\bu)$ for $\OP(\bu)$, we have $\delta = 0$ for all $j < j^*$.
\qedwhite\end{proof}

Proposition~\ref{propn:initialization_delta_barrier_ipm} first provides parameters $\lambda_0 > \tlambda$ and $\lambda_\Niter < \tlambda'$ for which the optimal solutions to $\BP(\bu, \barr, \lambda_0)$ and $\BP(\bu, \barr, \lambda_\Niter)$ lie inside and outside of $\Xfeas(\bu)$, respectively. Next, it shows that the sequence of $\lambda_j$ produces a sequence of optimal solutions $\{\bx^{\lambda_j}(\bu)\}$ that start within the feasible set $\Xfeas(\bu)$ and proceed to move outside. Finally, it derives a sequence of corresponding $\{ \epsilon_j\}$ such that the sequence of solutions are $(\Delta(\bu), \epsilon_j)$-optimal for $\OP(\bu)$.
This implies the final solution is $(0, (\bc^\tpose \btx'_1 - \bc^\tpose \btx'_2)\nu^{j^*-1-\Niter}))$-optimal for $\OP(\bu)$.

The above proposition summarizes an IPM for solving $\OP(\bu)$ when given a $\delta$-barrier $\barr(\bx, \bu)$ at least as good as the canonical barrier and an oracle $\Feasfunc$. The IPM behaves predictably; by initializing with large $\lambda$, we ensure that we obtain feasible solutions, but by decreasing $\lambda$, we know that the solution will ultimately be infeasible. An oracle could identify the point of termination immediately before the IPM leaves the feasible set. We can from here obtain a tight bound on the $(\delta, \epsilon)$-optimality of the final solution.

While direct optimization is desirable for its structural properties, this IPM approach is reliant on access to a $\delta$-barrier. On the other hand, IPMAN learns a classifier that approximates a $\delta$-barrier after several iterations. Therefore, unless we are given an a priori $\delta$-barrier (e.g., a canonical barrier for $\Xrel$), this IPM approach is not necessarily feasible from the onset. A potential fix would be to first train IPMAN until a $\delta$-barrier is obtained and then use the $\delta$-barrier IPM to solve subsequent problems. This ties to the second difference between the two approaches; IPMAN is ultimately a predictive model and is therefore subject to prediction error. On the other hand, prediction from a trained model is much faster than direct optimization. Therefore, in cases where the problem is large and an IPM would be difficult to solve or require numerous queries from an oracle, the predictive power of IPMAN yields more practical benefits.

\section{Proof of the generalization bound (Theorem~\ref{thm:prob_bound_generative_model})}
\label{sec:ec_proof_of_prob_bound}

The proof of the generalization bound uses a Generalization Lemma of~\citet{bertsimas:2014predictive} to bound the $(\delta, \epsilon)$-bound in terms of an empirical risk objective function error between $F^*(\bu)$ versus $\bxl(\bu)$ and Markov's inequality to translate this bound to a probabilistic $(\delta, \epsilon)$-optimality certificate. However, in order to use the lemma in this way, we first require an auxiliary result to relate $F^*$ with $\Rad(\set{F})$.

Assumption~\ref{ass:well_behaved_generator} states that the generative model $F^*$ is a composition; we project the classifier output to $\Xrel$ whenever $F^{(j, k)}(\bu) \notin \Xrel$. Although $F^{(j, k)} \in \set{F}$, the final model $F^*(\bu) := \Proj(F(\bu)) = \argmin_{\bx} \{ \norm{\bx - F(\bu)} \;|\; \bx \in \Xrel \}$ may not be a member of $\set{F}$. We first bound the Rademacher complexity of models composed from projection below.
\begin{lemma}
        \label{lem:bound_projected_rademacher_complexity}
        Let $\set{F} = \{ F: \set{U} \rightarrow \field{R}^n \}$ be a model class and $\Proj(\set{F}) = \{ \Proj(F) \;|\; F \in \set{F} \}$ be the class of models composed by a projection to a polyhedron $\Xrel$. Then for any $\datasetparam \sim \pparam$, $\hRad(\Proj(\set{F}), \datasetparam) \leq \sqrt{2n} \hRad(\set{F}, \datasetparam)$.
\end{lemma}
\begin{proof}{Proof of Lemma~\ref{lem:bound_projected_rademacher_complexity}.}
        We want to show for any fixed $\datasetparam$ that
        \begin{align}\label{eq:projected_rademacher}
                \EX_{\bsigma \sim p_{\bsigma}} \Bigg[ \frac{2}{\Nu} \sup_{F \in \set{F}}\sum_{i=1}^\Nu \bsigma_{i}^\tpose \Proj\big( F(\bhu_i) \big)\Bigg] \leq \sqrt{2n} \EX_{\bsigma \sim p_{\bsigma}} \Bigg[ \frac{2}{\Nu} \sup_{F \in \set{F}}\sum_{i=1}^\Nu \bsigma_{i}^\tpose F(\bhu_i) \Bigg].
        \end{align}
        By conditioning and iterating, it suffices to prove the following inequality for any function $\Xi(F): \set{F} \rightarrow \field{R}$,
        \begin{align}\label{eq:projected_rademacher_special_case}
            \EX_{\bsigma \sim p_{\bsigma}} \left[ \sup_{F \in \set{F}} \bsigma^\tpose \Proj(F) + \Xi(F) \right] \leq \EX_{\bsigma \sim p_{\bsigma}} \left[ \sup_{F \in \set{F}} \sqrt{2n} \bsigma^\tpose F + \Xi(F) \right].
        \end{align}
        We first prove inequality~\eqref{eq:projected_rademacher_special_case}, before returning to the main lemma.
        
        As $\bsigma \sim p_{\bsigma}$ is a random vector of i.i.d. Rademacher variables, it is supported over the (ordered) set $\{ (-1,\dots,-1,-1), (-1, \dots,-1, 1), \dots, (1,\dots,1, 1) \}$ all with equal probability. Let $\bhsigma_\ell$ denote the $\ell$-th element of this set. By iterating over all values, we expand the left-hand-side of~\eqref{eq:projected_rademacher_special_case} out to:
        \begin{align}
            &\EX_{\bsigma \sim p_{\bsigma}} \left[ \sup_{F \in \set{F}} \bsigma^\tpose \Proj(F) + \Xi(F) \right] = \frac{1}{2^n} \sum_{\ell=1}^{2^n} \left( \sup_{F \in \set{F}} \hat{\bsigma}_{\ell}^\tpose \Proj(F) + \Xi(F) \right) \label{eq:proj_rad_proof1}\\
            &\qquad \qquad \qquad = \frac{1}{2^n} \sum_{\ell=1}^{2^{n-1}} \left( \sup_{F \in \set{F}} \left\{ \hat{\bsigma}_{\ell}^\tpose \Proj(F) + \Xi(F) \right\} + \sup_{F \in \set{F}} \left\{-\hat{\bsigma}_{\ell}^\tpose \Proj(F) + \Xi(F) \right\} \right) \label{eq:proj_rad_proof2}\\
            &\qquad \qquad \qquad = \frac{1}{2^n} \sum_{\ell=1}^{2^{n-1}} \left( \sup_{F_1, F_2 \in \set{F}}\hat{\bsigma}_{\ell}^\tpose \left(\Proj(F_1) - \Proj(F_2) \right) + \Xi(F_1) + \Xi(F_2)  \right). \label{eq:proj_rad_proof3}
        \end{align}
        Equation~\eqref{eq:proj_rad_proof1} follows by letting $\hat{\bsigma}_{\ell}$ iterate over the support of the distribution. Equation~\eqref{eq:proj_rad_proof2} follows from the symmetry of the Rademacher distribution. That is, for every $\hat{\bsigma}_{\ell}$, there exists $-\hat{\bsigma}_{\ell}$ with equal probability, and we need to only characterize half of the elements in the support.~\eqref{eq:proj_rad_proof3} merges the suprema.
        
        By the Obtuse Angle Criterion, projection to a convex set is a non-expansive operation (i.e., $\norm{\Proj(F_1) - \Proj(F_2)} \leq \norm{ F_1 - F_2 }$). We use the Cauchy-Schwarz inequality and the non-expansiveness property (in~\eqref{eq:proj_rad_proof4} and~\eqref{eq:proj_rad_proof5} below, respectively) to remove the dependency on the projection operator:
        \begin{align}
            \text{RHS }\eqref{eq:proj_rad_proof3} & \leq \frac{1}{2^n} \sum_{\ell=1}^{2^{n-1}} \left( \sup_{F_1, F_2 \in \set{F}} \norm{\hat{\bsigma}_{\ell}} \norm{\Proj(F_1) - \Proj(F_2)} + \Xi(F_1) + \Xi(F_2)  \right) \label{eq:proj_rad_proof4} \\
            & \leq \frac{1}{2^n} \sum_{\ell=1}^{2^{n-1}} \left( \sup_{F_1, F_2 \in \set{F}} \norm{\hat{\bsigma}_{\ell}} \norm{F_1 - F_2} + \Xi(F_1) + \Xi(F_2)  \right) \label{eq:proj_rad_proof5} \\
            & \leq \frac{1}{2^n} \sum_{\ell=1}^{2^{n-1}} \left( \sup_{F_1, F_2 \in \set{F}} \sqrt{n} \norm{F_1 - F_2} + \Xi(F_1) + \Xi(F_2)  \right) \label{eq:proj_rad_proof6} \\
            & \leq \frac{1}{2} \left( \sup_{F_1, F_2 \in \set{F}} \sqrt{n} \norm{F_1 - F_2} + \Xi(F_1) + \Xi(F_2)  \right) \label{eq:proj_rad_proof7} \\
            & \leq \frac{1}{2} \left( \sqrt{n} \norm{F^*_1 - F^*_2} + \Xi(F^*_1) + \Xi(F^*_2)  \right) \label{eq:proj_rad_proof8}.
        \end{align}
        %
        Inequality~\eqref{eq:proj_rad_proof6} follows by noting $\norm{\bsigma} \leq \sqrt{n}$ for all $\bsigma \sim p_{\bsigma}$ and~\eqref{eq:proj_rad_proof7} from the fact that the dependency on $\hat{\bsigma}_{\ell}$ has been removed. We obtain~\eqref{eq:proj_rad_proof8} by letting $F^*_1$ and $F^*_2$ be the two values that attain the supremum.

        We use the Khintchine inequality to bound $\norm{F^*_1 - F^*_2} \leq \sqrt{2} \EX_{\bsigma \sim p_{\bsigma}} \left[ \left| \bsigma^\tpose (F^*_1 - F^*_2) \right| \right]$. We then rearrange the terms as follows:
        \begin{align}
            \text{RHS }\eqref{eq:proj_rad_proof8} & \leq \frac{1}{2} \left( \sqrt{2n} \EX_{\bsigma \sim p_{\bsigma}} \left[ \left| \bsigma^\tpose (F^*_1 - F^*_2) \right| \right] + \Xi(F^*_1) + \Xi(F^*_2)  \right) \label{eq:proj_rad_proof9} \\
            &= \frac{1}{2} \left(  \EX_{\bsigma \sim p_{\bsigma}} \left[ \sqrt{2n} \left| \bsigma^\tpose (F^*_1 - F^*_2) \right|+ \Xi(F^*_1) + \Xi(F^*_2) \right]  \right)   \label{eq:proj_rad_proof10} \\
            &\leq \frac{1}{2} \left(  \EX_{\bsigma \sim p_{\bsigma}} \left[ \sup_{F_1, F_2 \in \set{F}} \sqrt{2n}\left| \bsigma^\tpose (F_1 - F_2) \right|+ \Xi(F_1) + \Xi(F_2)  \right]  \right)  \label{eq:proj_rad_proof11} \\
            &= \frac{1}{2} \left(  \EX_{\bsigma \sim p_{\bsigma}} \left[ \sup_{F \in \set{F}} \left\{ \sqrt{2n} \bsigma^\tpose F + \Xi(F) \right\} + \sup_{F \in \set{F}} \left\{ -\sqrt{2n}\bsigma^\tpose F + \Xi(F) \right\} \right] \right)  \label{eq:proj_rad_proof13} \\
            &=  \EX_{\bsigma \sim p_{\bsigma}} \left[ \sup_{F \in \set{F}} \sqrt{2n} \bsigma^\tpose F + \Xi(F) \right]  \label{eq:proj_rad_proof14}. 
        \end{align}
        Inequality~\eqref{eq:proj_rad_proof10} brings all of the terms inside the expectation.~\eqref{eq:proj_rad_proof11} upper bounds by the supremum. Because $\Xi(F_1) + \Xi(F_2)$ is invariant under the exchange of $F_1$ and $F_2$, the supremum will be obtained when $\bsigma^\tpose (F_1 - F_2)$ is positive, meaning we can remove the absolute value and separate the supremum in~\eqref{eq:proj_rad_proof13}. Finally, the symmetry of the random variable $\bsigma$ implies that the two suprema are equal, thereby giving~\eqref{eq:proj_rad_proof14}.

        To complete the proof, we use a standard conditioning argument (see~\citet{maurer:2016vector}) to show~\eqref{eq:projected_rademacher} decomposes to~\eqref{eq:projected_rademacher_special_case}. For any $0 \leq m \leq \Nu$, we prove the following by induction:
        \begin{align*}
            \EX_{\bsigma \sim p_{\bsigma}} \left[ \sup_{F \in \set{F}} \sum_{i=1}^\Nu \bsigma_i^\tpose \Proj\big(F(\bhu_i)\big) \right] \leq \EX_{\bsigma \sim p_{\bsigma}} \left[ \sup_{F \in \set{F}} \sum_{i=1}^m \sqrt{2n} \bsigma_i^\tpose F(\bhu_i) + \sum_{i=m+1}^\Nu \bsigma_i^\tpose \Proj\big(F(\bhu_i)\big) \right].
        \end{align*}
        The case for $m=0$ is an identity. Now for fixed values of $\hat{\bsigma}_i, \forall i \neq m$, let
        \begin{align*}
            \Xi(F) = \sum_{i=1}^{m-1} \sqrt{2n} \hat{\bsigma}_i^\tpose F(\bhu_i) + \sum_{i=m+1}^\Nu \hat{\bsigma}_i^\tpose \Proj\big(F(\bhu_i)\big).
        \end{align*}
        Then, assuming the inequality holds for $m-1$, we show
        \begin{align*}
            &\EX_{\bsigma \sim p_{\bsigma}} \left[ \sup_{F \in \set{F}} \sum_{i=1}^\Nu \bsigma_i^\tpose \Proj\big(F(\bhu_i)\big) \right] \leq \EX_{\bsigma \sim p_{\bsigma}} \left[ \sup_{F \in \set{F}} \sum_{i=1}^{m-1} \sqrt{2n} \bsigma_i^\tpose F(\bhu_i) + \sum_{i=m}^\Nu \bsigma_i^\tpose \Proj\big(F(\bhu_i)\big) \right] \\
            &\qquad \qquad \qquad \qquad = \EX_{\bsigma \sim p_{\bsigma}} \left[ \EX_{\bsigma_m \sim p_{\bsigma}} \left[ \sup_{F \in \set{F}} \bsigma_m^\tpose \Proj\big(F(\bhu_m)\big) + \Xi(F) \;\bigg|\; \{ \hat{\bsigma}_i, \forall i \neq m \} \right] \right] \\
            &\qquad \qquad \qquad \qquad \leq \EX_{\bsigma \sim p_{\bsigma}} \left[ \EX_{\bsigma_m \sim p_{\bsigma}} \left[ \sup_{F \in \set{F}} \sqrt{2n} \bsigma_m^\tpose F(\bhu_m) + \Xi(F) \;\bigg|\; \{\hat{\bsigma}_i, \forall i \neq m\} \right] \right] \\
            &\qquad \qquad \qquad \qquad = \EX_{\bsigma \sim p_{\bsigma}} \left[ \sup_{F \in \set{F}} \sum_{i=1}^m \sqrt{2n} \bsigma_i^\tpose F(\bhu_i) + \sum_{i=m+1}^\Nu \bsigma_i^\tpose \Proj\big(F(\bhu_i)\big) \right].
        \end{align*}
        The second inequality comes from substituting~\eqref{eq:projected_rademacher_special_case}. When $m=\Nu$, the proof is complete.
\qedwhite\end{proof}

Lemma~\ref{lem:bound_projected_rademacher_complexity} can be seen as an extension of the main theorem of~\citet{maurer:2016vector} and is proved using a similar sequence of steps. There, the authors showed that composition of a Lipschitz scalar-valued vector function onto a vector-valued model class bounds the Rademacher complexity of the composed class by $\sqrt{2L}$. In the above, we compose the projection operator, a vector-valued function, to the vector-valued model class and bound the Rademacher complexity by $\sqrt{2n}$.
Although we only specifically consider the projection operator, the proof easily extends to any vector-valued function, so long as it is $L$-Lipschitz, whereupon we would reintroduce $L$ back into the bound.

Before proving Theorem~\ref{thm:prob_bound_generative_model}, we re-state the Generalization Lemma of~\citet{bertsimas:2014predictive}.
\begin{lemma}[\citet{bertsimas:2014predictive}]\label{lem:kallus_rademacher}
        Consider a function $z(\bx, \bu): \Xrel \times \set{U} \rightarrow \field{R}$ that is bounded and $L_\infty$-Lipschitz continuous in $\bx$ using the $\norm{\cdot}_\infty$ norm,
        \begin{align*}
                \sup_{\bx \in \Xrel, \bu \in \set{U}} z(\bx, \bu) \leq K, &\quad \sup_{\bx_1 \neq \bx_2 \in \Xrel, \bu \in \set{U}} \frac{z(\bx_1, \bu) - z(\bx_2, \bu)}{\norm{\bx_1 - \bx_2}_\infty} \leq L_\infty.
        \end{align*}
       For any $\beta > 0$, with probability at least $1 - \beta$ with respect to the sampling of $\datasetparam$,
        \begin{align*}
                \EX_{\bu \sim \pparam} \Big[ z\big(F(\bu), \bu\big) \Big] \leq \frac{1}{\Nu} \sum_{i=1}^\Nu z\big(F(\bhu_i), \bhu_i\big) + K \sqrt{\frac{\log(1/\beta)}{2\Nu}} + L_\infty \Rad\big(\Proj(\set{F})\big), \quad \forall F \in \Proj(\set{F}).
        \end{align*}
\end{lemma}

We are now ready to prove Theorem~\ref{thm:prob_bound_generative_model}.
\begin{proof}{Proof of Theorem~\ref{thm:prob_bound_generative_model}.}
        The proof follows by first applying Lemma~\ref{lem:kallus_rademacher}, before applying Markov's inequality. We let $z(\bx, \bu) = | \bc^\tpose \bx - \bc^\tpose \bxl(\bu) |$, as a function of $\bx \in \Xrel$ and $\bu \in \set{U}$, and show it is bounded from above
        \begin{align}
                \sup_{\bx \in \Xrel, \bu \in \set{U}} z(\bx, \bu) &= \sup_{\bx \in \Xrel, \bu \in \set{U}} \left| \bc^\tpose \bx - \bc^\tpose \bxl(\bu)) \right|  \\
                &\leq \max_{\bx \in \Xrel} \bc^\tpose \bx - \min_{\bx \in \Xrel} \bc^\tpose \bx = K.\label{eq:bounded_z_function}
        \end{align}
        Because $\Xrel$ is compact,~\eqref{eq:bounded_z_function} is bounded. We set $K$ to be equal to RHS~\eqref{eq:bounded_z_function}.

        We next show $L_\infty$-Lipschitz continuity,
        \begin{align}
                \sup_{\bx_1 \neq \bx_2 \in \Xrel, \bu \in \set{U}} \frac{z(\bx_1, \bu) - z(\bx_2, \bu)}{\norm{\bx_1 - \bx_2}_\infty} &= \sup_{\bx_1 \neq \bx_2 \in \Xrel, \bu \in \set{U}} \frac{\left| \bc^\tpose\bx_1 - \bc^\tpose\bxl(\bu) \right| -  \left| \bc^\tpose\bx_2 - \bc^\tpose\bxl(\bu)\right| }{\norm{\bx_1 - \bx_2}_\infty} \\
                &\leq \sup_{\bx_1 \neq \bx_2 \in \Xrel, \bu \in \set{U}} \frac{\left| \bc^\tpose\bx_1 - \bc^\tpose\bxl(\bu) - \bc^\tpose\bx_2 + \bc^\tpose\bxl(\bu)\right| }{\norm{\bx_1 - \bx_2}_\infty} \label{eq:lipschitz_z_function2}\\
                &= \sup_{\bx_1 \neq \bx_2 \in \Xrel} \frac{| \bc^\tpose\bx_1 - \bc^\tpose\bx_2|}{\norm{\bx_1 - \bx_2}_\infty} = L_\infty \label{eq:lipschitz_z_function3}
        \end{align}
        Inequality~\eqref{eq:lipschitz_z_function2} follows from the Reverse Triangle Inequality.~\eqref{eq:lipschitz_z_function3} follows from the fact that $f(\bx)$ is linear and therefore, Lipschitz continuous using the $\norm{\cdot}_\infty$ norm. We let $L_\infty$ be the Lipschitz constant of $f(\bx)$.

        Because $z(\bx, \bu)$ satisfies the bounded and Lipschitz continuity assumptions, we apply Lemma~\ref{lem:kallus_rademacher} to obtain 
        \begin{align*}
                \EX_{\bu \sim \pparam} \Big[ z\big(F(\bu), \bu\big) \Big] \leq \frac{1}{\Nu} \sum_{i=1}^\Nu z\big(F(\bhu_i), \bhu_i\big) + K \sqrt{\frac{\log(1/\beta)}{2\Nu}} + L_\infty \Rad\big(\Proj(\set{F})\big), \quad \forall F \in \Proj(\set{F}).
        \end{align*}
        Specifically, this bound holds for $F^* \in \Proj(\set{F})$. By Lemma~\ref{lem:bound_projected_rademacher_complexity}, we can bound $\Rad(\Proj(\set{F})) \leq \sqrt{2n} \Rad(\set{F})$.

        The remainder of the proof follows from Markov's inequality. For $\gamma > 0$,
        \begin{align*}
                \pparam \Big\{ z\big(F^*(\bu), \bu\big) > \gamma \Big\} &= \pparam \Big\{ \left| \bc^\tpose F^*(\bu) - \bc^\tpose \bxl(\bu) \right| > \gamma \Big\} \\
                &\leq \frac{\EX_{\bu \sim \pparam} \Big[ \left| \bc^\tpose F^*(\bhu_i) - \bc^\tpose \bxl(\bhu_i) \right| \Big] }{\gamma}.
        \end{align*}
        From the Law of Total Probability, we obtain
        \begin{align*}
                &\pparam \Big\{ \left| \bc^\tpose F^*(\bu) - \bc^\tpose \bxl(\bu) \right| \leq \gamma \Big\} = 1 - \pparam \Big\{ \left| \bc^\tpose F^*(\bu) - \bc^\tpose \bxl(\bu) \right| > \gamma  \Big\} \\
                &\qquad\geq 1 - \frac{\EX_{\bu \sim \pparam} \Big[ \left| \bc^\tpose F^*(\bhu_i) - \bc^\tpose \bxl(\bhu_i) \right| \Big] }{\gamma},  \\
                &\qquad\geq1 - \frac{\displaystyle\frac{1}{\Nu} \sum_{i=1}^\Nu \left| \bc^\tpose F^*(\bhu_i) - \bc^\tpose \bxl(\bhu_i) \right| + K \sqrt{\frac{\log(1/\beta)}{2\Nu}} + \sqrt{2n} L_\infty \Rad(\set{F})}{\gamma},
        \end{align*}
        with probability $1 - \beta$. The second and third line follow from Markov's inequality and substituting the bound from Lemma~\ref{lem:kallus_rademacher}, respectively. Given that we have a probabilistic bound for the error of $F^*(\bu)$ from $\bxl(\bu)$, we bound the error to $\bx^*(\bu)$. Recall that $\bxl(\bu)$ is $(\delta, \epsilon)$-optimal. There are two cases to consider. First, if $\bc^\tpose \bxl(\bu) \leq \bc^\tpose F(\bu))\leq \bc^\tpose \bxl(\bu) + \gamma$, then by substitution,
        \begin{align*}
                \bc^\tpose F^*(\bu) - \epsilon - \gamma < \bc^\tpose \bx^*(\bu) < \bc^\tpose F^*(\bu) + \delta.
        \end{align*}
        Alternatively, if $\bc^\tpose F^*(\bu) \leq \bc^\tpose \bxl(\bu) \leq \bc^\tpose F^*(\bu) + \gamma$, then by substitution,
        \begin{align*}
                \bc^\tpose F^*(\bu) - \epsilon < \bc^\tpose \bx^*(\bu) < \bc^\tpose F^*(\bu) + \delta + \gamma.
        \end{align*}
        Note that both of these events can be covered by adding and subtracting $\gamma$ to both the upper and lower bounds respectively. Then,
        \begin{align*}
        \pparam\Big\{ \bc^\tpose F^*(\bu) - \epsilon - \gamma < \bc^\tpose \bx^*(\bu) < \bc^\tpose F^*(\bu) + \delta + \gamma \Big\} \geq \pparam \Big\{ \left| \bc^\tpose F^*(\bu) - \bc^\tpose \bxl(\bu) \right| \leq \gamma \Big\},         \end{align*}
        completing the proof.
\qedwhite\end{proof}

\section{Supplementary content for automated portfolio optimization}
\label{sec:ec_portfolio}

We consider a personalized version of the non-convex portfolio optimization problem $\PP(\bu)$ of~\citet{kim:2019gan}, where the objective is to maximize the return subject to constraints on the portfolio risk and the cardinality (i.e., the total number of stocks one can invest). Each user possesses their own problem parameters and requires an optimal solution to a personalized problem.

We first translate the cardinality constraint into a form that is amenable for optimization solvers. 
\citet{kim:2019gan} assume that any stock that is allocated less than $0.001$ of the total portfolio does not affect the cardinality (i.e., $x_i \leq 0.001$ is equivalent to $x_i \approx 0$ for $i \in \{1,\dots,n\}$). We increase the bound to $0.005$ and reformulate $\PP(\bu)$ to a mixed-integer quadratic optimization problem:
\begin{subequations} \label{eq:personalized_portfolio_miqp}
\begin{align}
    \minimize{\bx, \by} \quad   & - \bmu^\tpose \bx \\
    \subjectto\quad             & \bx^\tpose \bSigma \bx \leq r(\bu) \\
                                & x_i \geq 0.005 y_i    &~& \forall i \in \{1, \dots, n\} \\
                                & x_i \leq y_i          &~& \forall i \in \{1, \dots, n\} \\ 
                                & \sum_{i=1}^n y_i \geq \lfloor  c(\bu) \rfloor \label{eq:personalized_portfolio_miqp5}\\
                                & \sum_{i=1}^n y_i \leq \lfloor  c(\bu) \rfloor + \lceil \Delta c (\bu) \rceil \label{eq:personalized_portfolio_miqp6}\\
                                & y_i \in \{0, 1\}      &~& \forall i \in \{1, \dots, n\}
\end{align}
\end{subequations}
Note that problem~\eqref{eq:personalized_portfolio_miqp} is non-convex only due to the inclusion of binary variables, $\by$. Thus, for any fixed $\by$ that satisfies the cardinality constraints~\eqref{eq:personalized_portfolio_miqp5} and~\eqref{eq:personalized_portfolio_miqp6}, the corresponding problem~\eqref{eq:personalized_portfolio_miqp} becomes a convex quadratic program. 

\subsection{Data set generation}

Our portfolio optimization experiments use exchange data collected in 2020. To construct this data set, we randomly selected 2500 stocks listed on three exchanges (Nasdaq, NYSE, NYSE American). For the randomly selected set, we recorded the relative daily change in price of each stock over 20 consecutive trading days between August 1 and August 30, 
and normalized all changes by the relative change in the Standard and Poor's 500 (S\&P 500). Next, we used a greedy heuristic to select 100 stocks from our pool of 2500 with the goal of maximizing the variance of the elements in the corresponding covariance matrix while limiting the correlation between them.
Table~\ref{tab:tickers} lists the 100 stocks used in our simulations.

\begin{table}[t]
\centering
\caption{The tickers and their exchanges that were sampled to generate the data for our portfolio optimization experiments. For each ticker, we collected data for 20 consecutive trading days between August 1 and August 30.}
\label{tab:tickers}
\small
\resizebox{\linewidth}{!}{
\begin{tabular}{llllll}
\toprule
Ticker & Exchange      & Company Name                     & Ticker & Exchange      & Company Name                    \\ \midrule
AAMC   & NYSE A. & Altisource Asset Management Corp.  & KZIA   & Nasdaq      & Kazia Therapeutics Limited      \\
AESE   & Nasdaq      & Allied Esports Entertainment, Inc.  & LEJU   & NYSE          & Leju Holdings Limited           \\
AIHS   & Nasdaq      & Senmiao Technology Limited       & LINC   & Nasdaq      & Lincoln Educational Services Corp. \\
AINC   & NYSE A.  & Ashford Inc.  & LINK   & Nasdaq      & Interlink Electronics, Inc.     \\
ALYA   & Nasdaq      & Alithya Group, Inc.               & LMPX   & Nasdaq      & LMP Automotive Holdings, Inc.   \\
ARLO   & NYSE          & Arlo Technologies, Inc.          & LPTH   & Nasdaq      & LightPath Technologies, Inc.    \\
ASPS   & Nasdaq      & Altisource Portfolio Solutions   & LWAY   & Nasdaq      & Lifeway Foods, Inc.             \\
AVGR   & Nasdaq      & Avinger, Inc.                    & MDIA   & Nasdaq      & Mediaco Holding Inc.            \\
BDR    & NYSE A. & Blonder Tongue Laboratories, Inc.  & MDLY   & NYSE          & Medley Management, Inc.         \\
BLNK   & Nasdaq      & Blink Charging Co.  & MICT   & Nasdaq      & MICT, Inc.                      \\
BOXL   & Nasdaq      & Boxlight Corporation             & MOXC   & Nasdaq      & Moxian (BVI) Inc.                \\
BPTH   & Nasdaq      & Bio-Path Holdings, Inc.          & MYMD   & Nasdaq      & MyMD Pharmaceuticals, Inc.      \\
CELH   & Nasdaq      & Celsius Holdings, Inc.           & NXTD   & Nasdaq      & Nxt-ID, Inc.                          \\
CMMB   & Nasdaq      & Chemomab Therapeutics Ltd.       & OCGN   & Nasdaq      & Ocugen, Inc.                    \\
CNDT   & Nasdaq      & Conduent Incorporated            & ONE    & NYSE          & OneSmart International Education Group Ltd. \\
CNFR   & Nasdaq      & Conifer Holdings, Inc.           & ONTX   & Nasdaq      & Onconova Therapeutics, Inc.     \\
COCP   & Nasdaq      & Cocrystal Pharma, Inc.           & OPGN   & Nasdaq      & OpGen, Inc.                     \\
COHN   & NYSE A. & Cohen \& Company Inc.            & OPTN   & Nasdaq      & OptiNose, Inc.                  \\
CTIB   & Nasdaq      & Yunhong CTI Ltd.                 & OSPN   & Nasdaq      & OneSpan Inc.                    \\
DAC    & NYSE          & Danaos Corporation               & OSUR   & Nasdaq      & OraSure Technologies, Inc.      \\
DBVT   & Nasdaq      & DBV Technologies S.A.            & PBPB   & Nasdaq      & Potbelly Corp.            \\
DTSS   & Nasdaq      & Datasea Inc.                     & PHCF   & Nasdaq      & Puhui Wealth Investment Management Co. \\
DTST   & Nasdaq      & Data Storage Corporation         & PIXY   & Nasdaq      & ShiftPixy, Inc.                 \\
EBON   & Nasdaq      & Ebang International Holdings, Inc.  & PLCE   & Nasdaq      & The Children's Place, Inc.    \\
EFOI   & Nasdaq      & Energy Focus, Inc.               & PPSI   & Nasdaq      & Pioneer Power Solutions, Inc.   \\
ELYS   & Nasdaq      & Elys Game Technology, Corp.      & PRPO   & Nasdaq      & Precipio, Inc.                  \\
EVK    & Nasdaq      & Ever-Glory International Group, Inc.  & PTE    & Nasdaq      & PolarityTE, Inc.                \\
EYEG   & Nasdaq      & Eyegate Pharmaceuticals, Inc.    & PVAC   & Nasdaq      & Penn Virginia Corp.             \\
EYPT   & Nasdaq      & EyePoint Pharmaceuticals, Inc.   & REDU   & Nasdaq      & RISE Education Cayman Ltd       \\
FAMI   & Nasdaq      & Farmmi, Inc.  & REFR   & Nasdaq      & Research Frontiers Incorporated \\
FINV   & NYSE          & FinVolution Group                & RIOT   & Nasdaq      & Riot Blockchain, Inc            \\
FRGI   & Nasdaq      & Fiesta Restaurant Group, Inc.    & RMED   & NYSE A. & Ra Medical Systems, Inc.        \\
FSLY   & NYSE          & Fastly, Inc.                     & SGOC   & Nasdaq      & SGOCO Group Ltd                 \\
FTK    & NYSE          & Flotek Industries, Inc.          & SINT   & Nasdaq      & SiNtx Technologies, Inc.        \\
FTSI   & NYSE A. & FTS International, Inc.          & SNDL   & Nasdaq      & Sundial Growers Inc.            \\
GBOX   & Nasdaq      & Greenbox POS                     & SONN   & Nasdaq      & Sonnet BioTherapeutics Holdings, Inc. \\
GEVO   & Nasdaq      & Gevo, Inc.                       & SRNE   & Nasdaq      & Sorrento Therapeutics, Inc.     \\
GLBS   & Nasdaq      & Globus Maritime Limited          & SSNT   & Nasdaq      & SilverSun Technologies, Inc.    \\
GLDG   & NYSE A. & GoldMining, Inc.                  & SSTI   & Nasdaq      & ShotSpotter, Inc.               \\
GRIL   & Nasdaq      & Muscle Maker, Inc                & TA     & Nasdaq      & TravelCenters of America Inc.   \\
GRIN   & Nasdaq      & Grindrod Shipping Holdings Ltd.  & TBLTW  & Nasdaq      & ToughBuilt Industries, Inc.     \\
HBP    & Nasdaq      & Huttig Building Products, Inc.   & TOMZ   & Nasdaq      & TOMI Environmental Solutions, I \\
HJLIW  & Nasdaq      & Hancock Jaffe Laboratories, Inc. & TRIB   & Nasdaq      & Trinity Biotech plc             \\
HPKEW  & Nasdaq      & HighPeak Energy, Inc.            & TRVI   & Nasdaq      & Trevi Therapeutics, Inc.        \\
HTGM   & Nasdaq      & HTG Molecular Diagnostics, Inc.  & TRVN   & Nasdaq      & Trevena, Inc.                   \\
IMBI   & Nasdaq      & iMedia Brands, Inc.              & VIVE   & Nasdaq      & Viveve Medical, Inc.            \\
JMIA   & NYSE          & Jumia Technologies AG            & VRCA   & Nasdaq      & Verrica Pharmaceuticals Inc.    \\
JOB    & NYSE A. & GEE Group Inc.                   & WLL    & NYSE          & Whiting Petroleum Corporation   \\
KIRK   & Nasdaq      & Kirkland's, Inc.                 & WORX   & Nasdaq      & SCWorx Corp.                    \\
KULR   & NYSE A. & KULR Technology Group, Inc.      & XNET   & Nasdaq      & Xunlei Limited                 \\
\bottomrule
\end{tabular}
}
\end{table}



\begin{figure}[t]
    \centering
    \includegraphics[width=0.95\linewidth]{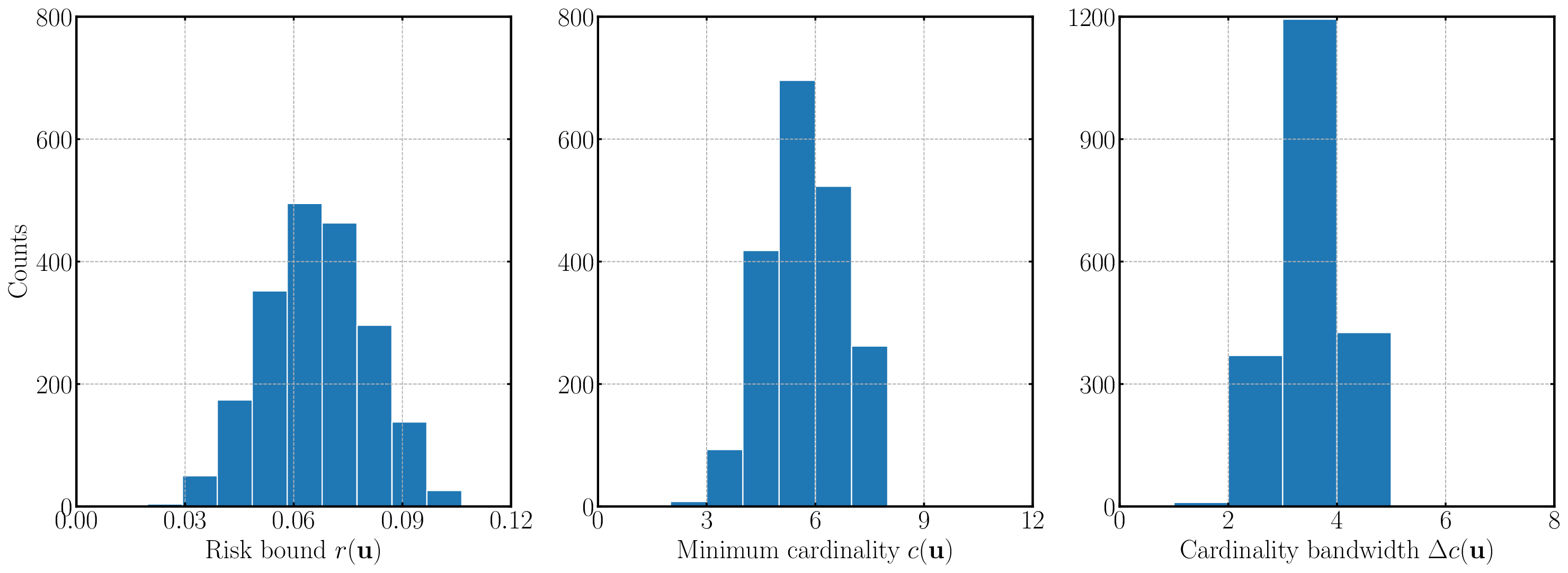}
    \caption{Histogram of parameter values of $\PP(\bu)$ for the test set in the two experiments. The average user in the test set has a risk tolerance of 0.066, minimum 5 investments, and up to 8 investments.}
    \label{fig:portfolio_params}
\end{figure}

The histograms of the final parameter distributions are given in Figure~\ref{fig:portfolio_params}. 
We seek parameter settings for $r(\bu)$, $c(\bu)$, and $\Delta c (\bu)$ such that $\PP(\bu)$ is difficult to solve within the latency requirements of an online application. The non-convex problem typically becomes more challenging for smaller values of these parameters. We randomly generate three linear functions $\bb_r, \bb_c, \bb_\Delta \in \field{R}^p$ where each element $b_i \sim U[0, 1]$ is uniformly sampled and i.i.d. We then normalize each of these models such that $\norm{\bb_r} = 1/20$, $\norm{\bb_c} = 5$, and $\norm{\bb_\Delta} = 3$. Then, the ``true models'' that map the context vectors to each parameter are $r(\bu) = \bb_r^\tpose \bu$, $c(\bu) = \bb_c^\tpose \bu + 1$, and $\Delta(\bu) = \bb_\Delta^\tpose \bc + 1$. We add $1$ in the latter cases to prevent these parameters from being equal to $0$. 
Under these settings, the average test set user has a risk tolerance of $\field{E}[r(\bu)] = 0.066$, a minimum number of investments $\field{E}[c(\bu)] = 5$, and a minimum bandwidth $\field{E}[\Delta c(\bu)] = 3$.


We randomly generate a data set of $14,000$ users split between training, validation, and testing. For each user, we randomly generate a context vector $\bhu_i$. Then, $\datasetparam$ is the collection of $10,000$ context vectors for the training set users. We then randomly generate $\dataset$ and $\datasetinfeas$ of $100,000$ points each by solving relaxations of problem~\eqref{eq:personalized_portfolio_miqp}. To generate feasible decisions for each user, we randomly sample a vector $\by$ that satisfies constraints~\eqref{eq:personalized_portfolio_miqp5} and~\eqref{eq:personalized_portfolio_miqp6}, fix this variable, and then solve problem~\eqref{eq:personalized_portfolio_miqp}. To generate infeasible decisions for each user, we randomly sample a vector $\by$ that fails to satisfy the cardinality constraints, i.e., $\sum_{i=1}^n y_i > \lfloor  c(\bu) \rfloor + \lceil \Delta c (\bu) \rceil$ or $\sum_{i=1}^n y_i < \lfloor  c(\bu) \rfloor$. We randomly generate 10 feasible and 10 infeasible decisions per user in the training set.

\subsection{Implementation of baselines}

We consider three different predict-then-optimization baselines. For each user, these baselines have access to estimates of the parameters $(r(\bu), c(\bu), \Delta c(\bu))$ for any given $\bu \in \set{U}$, which they use to solve their corresponding optimization problem. The first and second baselines are direct implementation of problem~\eqref{eq:personalized_portfolio_miqp}, which we refer to as the Mixed Integer Quadratic Program (MIQP) and MIQP ($0.2$ s) baselines. We implement both baselines using Gurobi 8.1~\citep{gurobi} and set the maximum run time of the solver to $100$ s and $0.2$ s, respectively.

The third baseline implements an approach introduced in~\citet{bertsimas:2022scalable} to reformulate the context-dependent constraints into a constrained regression problem, which we refer to as MIQP-R. In their paper,~\citet{bertsimas:2022scalable} consider a variation of our portfolio optimization problem
\begin{subequations}\label{eq:personalized_portfolio_bertsimas}
\begin{align*}
    \min_{\bx}\quad  & -\bmu^\tpose \bx + \frac{\sigma}{2} \bx^\tpose \bSigma \bx  + \frac{1}{2\gamma} \norm{\bx}_2^2 \\ 
    \st \quad  
    & \lfloor c(\bu) \rfloor \leq \card(\bx) \leq \lfloor c(\bu) \rfloor + \lceil \Delta c(\bu) \rceil \\
    & \bx^\tpose \bone \leq 1, \quad \bx \geq \bzero  
\end{align*}
\end{subequations}
where $\sigma, \gamma > 0$ are fixed coefficients.
Compared to our formulation~\eqref{eq:personalized_portfolio_miqp}, the above problem now minimizes the risk as an additional term in the objective function instead of enforcing a constraint. Furthermore, this variation includes a regularization term $\norm{\bx}_2^2$. 
The authors show that after normalizing the risk matrix $\Sigma \gets \sigma \Sigma$, this problem is equivalent to the following constrained regression problem
\begin{subequations}\label{eq:personalized_portfolio_bertsimas_constrained}
\begin{align}
    \min_{\bx}\quad  & \bd^\tpose \bx + \frac{1}{2\gamma} \norm{\bx}_2^2 + \frac{1}{2} \norm{\bX \bx - \by}_2^2    \\ 
    \st \quad  
    & \lfloor c(\bu) \rfloor \leq \card(\bx) \leq \lfloor c(\bu) \rfloor + \lceil \Delta c(\bu) \rceil \\
    & \bx^\tpose \bone \leq 1, \quad \bx \geq \bzero  
\end{align}
\end{subequations}
where $\bX \in \field{R}^{r \times n}$ corresponds to the Cholesky decomposition of the risk matrix, i.e., $\bSigma = \bX^\tpose \bX$. Furthermore, $\bd := (\bX^\tpose (\bX \bX^\tpose)^{-1} - \bI) \bmu$ and $\by := (\bX \bX^\tpose)^{-1} \bmu$ are projections of $\bmu$ to the null space and span of $\bX$, respectively. The constrained regression version is notably an easier problem to solve than the original variant.

Finally, in order to create our baseline, we observe that our problem $\PP(\bu)$ requires the portfolio for each user to satisfy a personalized risk tolerance. 
As a result, for each user, we define a personalized coefficient $\gamma \gets \gamma / r(\bu)$ in order to penalize the regularization term (or equivalently the regression and projection terms that are directly tied to the risk coefficient. Thus, our final implementation of the MIQP-R baseline follows problem~\eqref{eq:personalized_portfolio_miqp} and can be written as 
\begin{align*}
    \min_{\bx, \by}\quad  & \bd^\tpose \bx + \frac{r(\bu)}{2\gamma} \norm{\bx}_2^2 + \frac{1}{2} \norm{\bX \bx - \by}_2^2    \\ 
    \st \quad  
    & x_i \geq 0.005 y_i    &~& \forall i \in \{1, \dots, n\} \\
                                & x_i \leq y_i          &~& \forall i \in \{1, \dots, n\} \\ 
                                & \lfloor  c(\bu) \rfloor \leq \sum_{i=1}^n y_i \leq \lfloor  c(\bu) \rfloor + \lceil \Delta c (\bu) \rceil \\
                                & y_i \in \{0, 1\}      &~& \forall i \in \{1, \dots, n\}
\end{align*}
When designing this baseline, we found that setting different values for $\gamma$ and $\sigma$ changes the average optimality gap, fraction of feasible instances, and run time. As a result, we opted to minimize the run time without sacrificing feasibility or objective function value. After cross-validating on the training set, we set both $\gamma = 1$ and $\sigma = 1$.

\subsection{Neural network architecture} \label{section:nnarch}

\begin{table}[t]		
\centering
\caption{Left: Overview of the generator architecture. Right: Overview of the classifier architecture. \batchnorm~refers to batch normalization; \leakyrelu~and $\softmax$~refer to Leaky $\ReLU$ (0.2 slope) and Softmax activations, respectively.}
\small
\begin{tabular}{ccc}
\toprule
    Layer   & Input shape    & Activation \\\midrule
    1       & $10$     & \leakyrelu \\
    2--5    & $200$   & \bnlr \\
    6       & $200$   & $\softmax$ \\
    Output  & $100$   & --- \\
\bottomrule
\end{tabular}
\begin{tabular}{ccc}
\toprule
    Layer   & Input shape    & Activation \\\midrule
    1       & $110$ & \leakyrelu \\
    2--4    & $100$ & \leakyrelu \\
    5       & $100$ & $\mathrm{sigmoid}$ \\
    Output  & $1$   & --- \\
\bottomrule
\end{tabular}
\label{tab:gen_disc_pp}
\end{table}

The architectures for the generator and classifier are described in Tables~\ref{tab:gen_disc_pp}. Both the generator and classifier are fully-connected neural networks. The generator takes as input the context vector $\bu \in \field{R}^{10}$ and is composed of six layers. Each layer has a Leaky $\ReLU$ activation and the inner layers also incorporate batch normalization. The final layer of the generator is passed through a $\softmax$ activation function which ensures that $F(\bu)$ is a vector of positive elements which sum up to one. Consequently, the generator will always output a valid portfolio that satisfies the known polyhedral constraints and the core challenge is to learn to satisfy the constraints. The classifier takes as input $(\bx, \bu) \in \field{R}^{110}$ and is composed of five layers with Leaky $\ReLU$ activations between each layer. The final output is passed through a Sigmoid activation to ensure that $B(\bx, \bu) \in [0, 1]$.

\subsection{Implementation of the IPMAN algorithm}

Our training algorithm follows the procedure described in Algorithm~\ref{alg:ipman_summary}. All models are trained using the Adam optimizer with $(\beta_1, \beta_2) = (0.9, 0.999)$. We train the classifier with a batch size of $2000$ data points, a learning rate of $1 \times 10^{-2}$, and plateau-scheduling decay. We train the generator with a batch size of $1000$ data points, a learning rate of $5 \times 10^{-3}$, and plateau-scheduling delay. In each iteration of the IPMAN algorithm, we train the generator and classifier for $10$ epochs each.

Classical implementations of IPMs require a good initial point (i.e., lying within the feasible set) to construct a trajectory of points leading to an optimal solution. Similarly, IPMAN can be made more efficient by ensuring that the generative model is initialized to predict points that are likely to be feasible. This initialization can both reduce the training time and improve the stability of the algorithm. Consequently, we first pre-train both the generator and the classifier at the beginning of the algorithm~\citep{goodfellow:2016deep}.

The classifier is pre-trained using the initial data set of feasible $\dataset^{(0)}$ and infeasible decisions $\datasetinfeas^{(0)}$ with the same learning rate and settings as used in IPMAN, but for $200$ epochs. This initial pre-training ensures that the classifier is warm-started to a good barrier function from the offset of the IPMAN algorithm and, therefore, reduces the overall training time that is required. Note that without pre-training the classifier, we would have to run more iterations of IPMAN. Pre-training the classifier is equivalent to training the classifier in the first iteration of IPMAN for $200$ epochs and subsequently training for $10$ epochs per iteration in subsequent iterations.

The generator is pre-trained using a sub-set of $\dataset^{(0)}$ constructed by randomly selecting a single feasible decision $\bhx$ per context vector $\bhu$ in $\dataset^{(0)}$ (i.e., a total of $5000$ points). With this sub-set, we train the classifier to minimize mean-squared error, similar to prior approaches to predicting decisions~\citep{Larsen:2018aa}. We use the same training settings for pre-training the generator as we do in IPMAN and train for $200$ epochs.

\subsection{Ablation on model complexity}
\label{sec:ec_portfolio_model_size}

Although IPMAN is designed with neural networks that tend to be complex, black-box models, a natural question is whether simpler and potentially more intrepretable models can be used instead. Since the learning problems $\FCP(\dataset, \datasetinfeas)$ and $\GBP(\datasetparam, B^*, \lambda)$ use differentiable loss functions to train their respective models, $B(\bx, \bu)$ and $F(\bu)$, any model class trained with gradient-based methods can be used with IPMAN (e.g., logistic and linear regression, SVM). Thus, in this subsection, we ablate the effect of using simpler models as opposed to those introduced in \ref{section:nnarch}.

Note that of the two models, the classifier specifically requires the most predictive accuracy since its support is used to train the generator. 
That is, Lemma 1 and Proposition 1 assume the model class $\set{B}$ can learn complex mappings. We do not require any analogous assumption in Theorem 2 or Theorem 3 for $\set{F}$; a weaker generator simply means larger $(\delta, \epsilon)$-optimality guarantees. Further, the Rademacher complexity results can also be more easily bounded for simple models such as linear regressors (i.e., $F(\bu) := \bW \bu$ for weights $\bW$). Thus for applications requiring interpretability, an option is to use a simpler model class for the generator and let the classifier remain a black-box.

In our ablation study, we iteratively decrease the complexity of the generator by reducing the number of layers in the neural network from the baseline used in our experiments (six).
Table~\ref{tab:model_size_portfolio} summarizes the performance statistics in terms of (i) the fraction of the test set instances for which a feasible decision is generated; and (ii) the average optimality gap. The results indicate that as model complexity is decreased (i.e., fewer neural network layers), the generator's ability to produce feasible decisions is reduced. This suggests that there may be an accuracy/complexity tradeoff, which is a known trend in machine learning research \citep[e.g., see Figure 2.7 in][]{gareth2013introduction}.


\begin{table}[t]		
\centering
\caption{Summary statistics of IPMAN with models of different architectures. The optimality gap is measured over only feasible decisions. For each row, we reduce the number of internal layers of the generator that was defined in Table~\ref{tab:gen_disc_pp}.}
\small
\begin{tabular}{ccc}
\toprule
    Number of layers   & Fraction of feasible $F(\bu) \in \Xfeas(\bu)$ (\%)    & Optimality gap (\%) \\\midrule
    1       & $7.3$  & $20.0$ \\
    2       & $0$    & --- \\
    3       & $0$    & --- \\
    4       & $41.9$ & $32.3$ \\
    5       & $69.3$ & $37.2$ \\
    6       & $97.6$ & $17.4$ \\
\bottomrule
\end{tabular}
\label{tab:model_size_portfolio}
\end{table}


\section{Supplementary content for personalized cancer treatment}
\label{sec:ec_RT}


RT treatments are delivered by a linear accelerator (LINAC) that projects 
high-energy X-rays from different angles to treat cancerous targets inside of the patient. The patient's body is discretized into voxels (i.e., $5\text{mm}\times5\text{mm}\times2\text{mm}$ volumetric pixels) and the dose delivered to each of these voxels is used to assess the quality of a treatment. 
The automated design of an RT treatment plan involves first using a machine learning model to predict a dose that is clinically desirable for a given patient. This predicted dose is then transformed into a configuration of beamlet intensities via mathematical optimization; these beamlet intensities are the parameters of the LINAC that allow it to deliver a dose that is nearly identical to the predicted treatment~\citep{Babier:2018a}. We refer the reader to~\citep{kaderka:2021wide} for a recent paper on the implementation of RT automated planning.
In this work, we focus on constructing clinically desirable dose distributions following the same procedures as the prediction components in~\citet{mahmood:2018gancer, Babier:2019aa}

Each of the OARs and targets require polyhedral upper and lower bound constraints to the mean dose delivered to that structure. Furthermore, there exists a ``clinical criteria'' constraint for each OAR and target that must be satisfied at the discretion of an oncologist. That is, if the ground truth treatment plan for a given patient from the data set satisfies a constraint, then any generated plan for that patient must also satisfy that constraint. The constraint for each OAR is an upper bound on either the mean or maximum dose delivered to that structure, while the constraint for each target is a lower bound on the value-at-risk, i.e., minimum dose delivered to 90-th percentile of the target structure. The oracle $\Feasfunc$ is a look-up table that compares the dose generated by our model with the ground truth (i.e., what was actually delivered). In particular, for each structure, $\Feasfunc$ checks whether the input dose satisfies all the constraints (i.e., two polyhedral and one clinical constraint). We expand on the classification of feasibility in the sections below.

\subsection{Neural network architecture}
\label{sec:ec_RT_architecture}

We use a modified version of the generative adversarial network (GAN) of~\citet{Babier:2019aa}, where two networks learn to predict dose distributions. 
The models are described in Table~\ref{tab:gen_rt}.

The generator takes as input a tensor $\bu \in \field{R}^{128\times128\times128\times8}$, where the first three dimensions correspond to a voxel in the patient's geometry. The fourth dimension is a concatenation of the CT image greyscale and a one-hot encoded vector in $\{0, 1\}^7$ whose elements label whether the voxel belongs to one of the seven contoured structures. The generator then outputs a tensor $\bx \in \field{R}^{128\times128\times128}$ whose elements specify 
the dose to be delivered to each voxel of the patient.

The classifier is trained to predict whether a given dose distribution satisfies all of the constraints for each structure of the patient. This network takes as input the concatenated tensor $(\bx, \bu)$ and outputs a vector in $[0, 1]^7$, whose elements each indicate the classifier's belief of whether the given dose distribution has satisfied all of the constraints for each 
specific structure. 
Consequently, we treat learning feasibility as a multi-label classification problem and the classifier acts as seven separate classifiers each predicting feasibility with respect to an individual structure, but whose model parameters are shared with each other. 
For any structure, in order to classify a dose distribution as satisfying the relevant constraints, the classifier must: (i) determine from the CT image whether the patient requires a clinical constraint to be satisfied, and (ii) determine from the dose whether the clinical constraint is satisfied if this constraint is required for the patient. Overall, a dose distribution is feasible only if all constraints are satisfied.

\begin{table}[t]		
\centering
\caption{Top: Overview of the generator architecture. Bottom: Overview of the classifier architecture. \batchnorm~refers to batch normalization; \leakyrelu, \texttt{R}, and $\tanh$ refer to Leaky $\ReLU$ (0.2 slope), $\ReLU$, and Tanh activations, respectively; \avgpool~refers to a mean pool; and \dropout~refers to dropout.}
\small
\begin{tabular}{ccccc}
\toprule
    Layer   & Concatenate with  & Input shape    & Block & Activation \\\midrule
    1       & ---   & $128\times128\times128\times8$    & \convblk   & \bnlr \\
    2       & ---   & $64\times64\times64\times64$      & \convblk   & \bnlr \\
    3       & ---   & $32\times32\times32\times128$     & \convblk   & \bnlr \\
    4       & ---   & $16\times16\times16\times256$     & \convblk   & \bnlr \\
    5       & ---   & $8\times8\times8\times512$        & \convblk   & \bnlr \\
    6       & ---   & $4\times4\times4\times512$        & \convblk   & \bnlr \\
    7       & ---   & $2\times2\times2\times512$        & \deconvblk   & \leakyrelu \\
    8       & layer 5 output   & $4\times4\times4\times1024$        & \deconvblk   & \bnr \\
    9       & layer 4 output & $8\times8\times8\times1024$       & \deconvblk   & \bndr \\
    10      & layer 3 output & $16\times16\times16\times512$    & \deconvblk   & \bndr \\
    11      & layer 2 output & $32\times32\times32\times256$    & \deconvblk   & \bnr \\
    12      & layer 1 output & $64\times64\times64\times128$     & \deconvblk   & \avgpool-$\tanh$ \\
    Output  & ---   & $128\times128\times128\times1$    & ---    & --- \\
\bottomrule
\end{tabular}

\vspace{1em}

\begin{tabular}{cccc}
    \toprule
    Layer   & Input shape & Block & Activation \\ \midrule
    1       & $128\times128\times128\times9$    & \convblk  & \leakyrelu \\
    2       & $64\times64\times64\times64$      & \convblk  & \bnlr \\
    3       & $32\times32\times64\times128$     & \convblk  & \bnlr \\
    4       & $16\times16\times16\times256$     & \convblk  & \bnlr \\
    5       & $8\times8\times8\times512$        & \convblk  & $\mathrm{sigmoid}$ \\
    Output  & $7$                               & ---       & ---\\
    \bottomrule
\end{tabular}
\label{tab:gen_rt}
\end{table}

\subsection{Implementation of the IPMAN algorithm}
\label{sec:ec_RT_algorithm}

\begin{algorithm}[t]
    \caption{IPMAN}
    \label{alg:ipman}
    \begin{algorithmic}[1]
            \renewcommand{\algorithmicrequire}{\textbf{Input:}}
            \renewcommand{\algorithmicensure}{\textbf{Output:}}
            \REQUIRE Data sets of decisions $\dataset$, $\datasetinfeas$, and inputs $\datasetparam = \{ \bhu_i \}_{i=1}^\Nu$, Set of dual variables $\{ \lambda_j \}_{j=0}^M$, Number of iterations $K$, Number of epochs $E_B, E_F$, Subset sampling rate $s$
            \ENSURE  Final generative models $F^{(j,K)}$ for $j \in \{0, \dots, M\}$
            \STATE Pre-train generator using the steps from~\citet{Babier:2019aa}.
            \STATE Initialize generator $F^{(j, 0)} \gets F^*$ for $j \in \{0, \dots, M\}$, classifier $B$.
            \FOR{$k = 1$ \TO $K$}
                \STATE Sample subsets to train $\dataset^{(k)} = \sigma(\dataset; s)$,  $\datasetinfeas^{(k)} =  \sigma(\datasetinfeas; s |\dataset|/|\datasetinfeas|)$. 
                \FOR{$e = 0$ \TO $E_B$}	   
                    \STATE Update classifier $B^{(k)} \gets \Adam(\nabla L_B)$.
                \ENDFOR
                \FOR{$j = 0$ \TO $M$}
                    \FOR{$e = 0$ \TO $E_F$}
                        \STATE Update generator $F^{(j, k)} \gets \Adam(\nabla L_F )$.
                    \ENDFOR
                    \FORALL{$\bhu_i \in \datasetparam$}
                        \STATE Append $\dataset \gets \dataset \cup (F^{(j, k)}(\bhu_i), \bhu_i)$ if $\Psi(F^{(j, k)}(\bhu_i), \bhu_i) = 1$ else $\datasetinfeas \gets \datasetinfeas \cup (F^{(j, k)}(\bhu_i), \bhu_i)$.
                    \ENDFOR
                \ENDFOR
            \ENDFOR
            \RETURN $F^{(j,K)}$ for $j \in \{0, \dots, M\}$
    \end{algorithmic}
\end{algorithm}

We summarize the steps in Algorithm~\ref{alg:ipman}. All models are trained using the Adam optimizer with $(\beta_1, \beta_2) = (0.5, 0.999)$. We train the classifier with a learning rate of $1\times10^{-3}$ and the generator with a learning rate of $2\times10^{-5}$. 
Below, we remark on several modifications of the IPMAN algorithm. 

\subsubsection{Sampling an infeasible data set of decisions $\datasetinfeas$.}

Similar to the experiments on personalized portfolio optimization, we first pre-train the generator and subsequently apply transfer learning at the beginning of the algorithm~\citep{goodfellow:2016deep}. Here, pre-training amounts to training the generator as a GAN to learn to predict dose distributions from CT images. 
We follow the exact setup from~\citet{Babier:2019aa} for this pre-training step. 

Training IPMAN requires an initial data set of infeasible decisions $\datasetinfeas$. In practice, a data set of infeasible decisions would not be available a priori and, instead, is generated by sampling. However, the generator produces decisions $F(\bhu_i)$ during the pre-training. We save these decisions and label them afterwards as feasible or infeasible by evaluating whether these treatments satisfy the appropriate clinical constraints. With 100 patients, we generate a total of $5000$ dose distributions that are labeled as feasible or infeasible and then binned in the appropriate $\dataset$ or $\datasetinfeas$, respectively. 

\subsubsection{Learning multi-label feasibility with sub-sampled data sets.}

We make two modifications to the Feasibility Classification Problem in the main paper. 
First, training IPMAN for multiple iterations can produce a large quantity of generated data points, which causes training the classifier to quickly become prohibitively expensive. 
Consequently, we do not use smaller sampled subsets of $\dataset$ and $\datasetinfeas$ for training. Let $\sigma(\cdot;s)$ be a random sampling operator (without replacement) where $s$ is the fraction of points to sample. For example, $\sigma(\dataset; 0.5)$ denotes a randomly sampled subset of size $0.5 | \dataset |$. In our implementation, we set $s = 0.3$ and trained the classifier using $\dataset^{(k)} = \sigma(\dataset; 0.3)$ and $\datasetinfeas^{(k)} = \sigma(\datasetinfeas; 0.3|\dataset|/|\datasetinfeas|)$; this reduced the training time to 24 hours. 

Second, since we know that $\Xfeas(\bu) = \cup_{r \in \set{R}} \Xfeas_r(\bu)$ is comprised of seven different constraints, we treat feasibility as a multi-label classification problem. 
For any $(\bhu_i, \bhx_i)$ in $\dataset$ or $\datasetinfeas$, let $\psi_{i,r}$ be a label determining whether the dose distribution had satisfied the polyhedral and the context-dependent clinical constraints for structure $r$. That is, if the ground truth dose for $\bhu_i$ satisfied the clinical constraints, $\psi_{i, r} = 1$ if the polyhedral and clinical constraints were satisfied and zero otherwise. If the clinical dose did not satisfy the contextual constraint, then $\psi_{i, r} = 1$ if only the polyhedral constraints were satisfied; here, the clinical constraint is inactive for this patient. Then, let $[B(\bx,\bu)]_r$ denote the $r$-th element of the classifier output. The classifier problem is
\begin{align*} 
	\max_{B \in \set{B}}~& \left\{ L_{B} := \frac{1}{\Nx + \Nbx} \sum_{\substack{(\bhx_i, \bhu_i) \in \\ \dataset^{(k)} \cup \datasetinfeas^{(k)}}} \sum_{r \in \set{R}}  \left( \psi_{i, r} \log\big[B(\bhx_i, \bhu_i)\big]_r + (1-\psi_{i, r}) \log\Big( 1 - \big[B(\bhx_i, \bhu_i)\big]_r\Big) \right) \right\}.
\end{align*}
The above problem specializes to $\FCP(\dataset^{(k)}, \datasetinfeas^{(k)})$ in the single-class setting (i.e., $|\set{R}| = 1$). For a dose distribution to be classified feasible, $B(\bx, \bu)_r = 1$ for all $r \in \set{R}$. Separating constraint satisfaction along structures individually is equivalent to optimization with a barrier function for each constraint. Furthermore, the barriers are approximated by a neural network classifier with shared weights except in the last layer. As we describe below, the objective of the barrier optimization problem is obtained by summing all of the separate barriers, i.e., $f(\bx) - \lambda \sum_{r \in \set{R}} \log[B(\bx,\bu)]_r$. We minimize $L_B$ using the Adam optimizer for $E_B=10$ epochs in every iteration. 

\subsubsection{Regularized barrier optimization problem.}
\label{sec:ec_RT_algorithm_regularizer}

Style Transfer GANs, which inspire our models, often use an $l_1$-regularization term in training to ensure that generated data does not deviate too far from a real (i.e., clinically delivered dose) data point~\citep{isola:2017image}. 
We only use this regularization in the first set of experiments (Section~\ref{sec:rt_learning}) and remove it in the second set of experiments (Section~\ref{sec:rt_new_clinic}). 
Let $\bc$ denote the linear cost vector denoting the average of average doses to the OARs. Then, our modified generative barrier problem is 
\begin{align*} 
	\min_{F \in \set{F}}~& \left\{ L_{F} := \frac{1}{\Nu} \sum_{\bhu_i \in \datasetparam} \left( \frac{1}{\lambda_j} \bc^\tpose F(\bhu_i) + \lambda_{ST} \norm{F(\bhu_i) - \bhx_i}_1 -  \sum_{r \in \set{R}} \Big[ B^{(k)}\big(F(\bhu_i), \bhu_i\big) \Big]_r \right) \right\}.
\end{align*}
where $\lambda_{ST}$ is the regularization weight and $\bhx_i$ is the clinically delivered dose for patient $i$. 
We set $\lambda_{ST} =50$ for the first set of experiments and $\lambda_{ST} =0$ for the second set.
We minimize $L_F$ using the Adam optimizer for $E_F = 1$ epoch in every iteration. It is 
important to ensure that the classifier is 
trained close to optimality to ensure that it approximates a $\delta$-barrier. Furthermore, given the nature of training the classifier, it is often the case that the classifier's support is uneven and may have areas of local optimality that the generator may abuse. A standard practice in the GAN literature is to control the training duration of the two networks; we employ a similar strategy by training the generative model for a shorter duration than the classifier in order to ensure that the generator does not overfit and abuse local optima caused by the classifier.


\begin{figure}[t]
    \centering
    \includegraphics[width=0.5\linewidth]{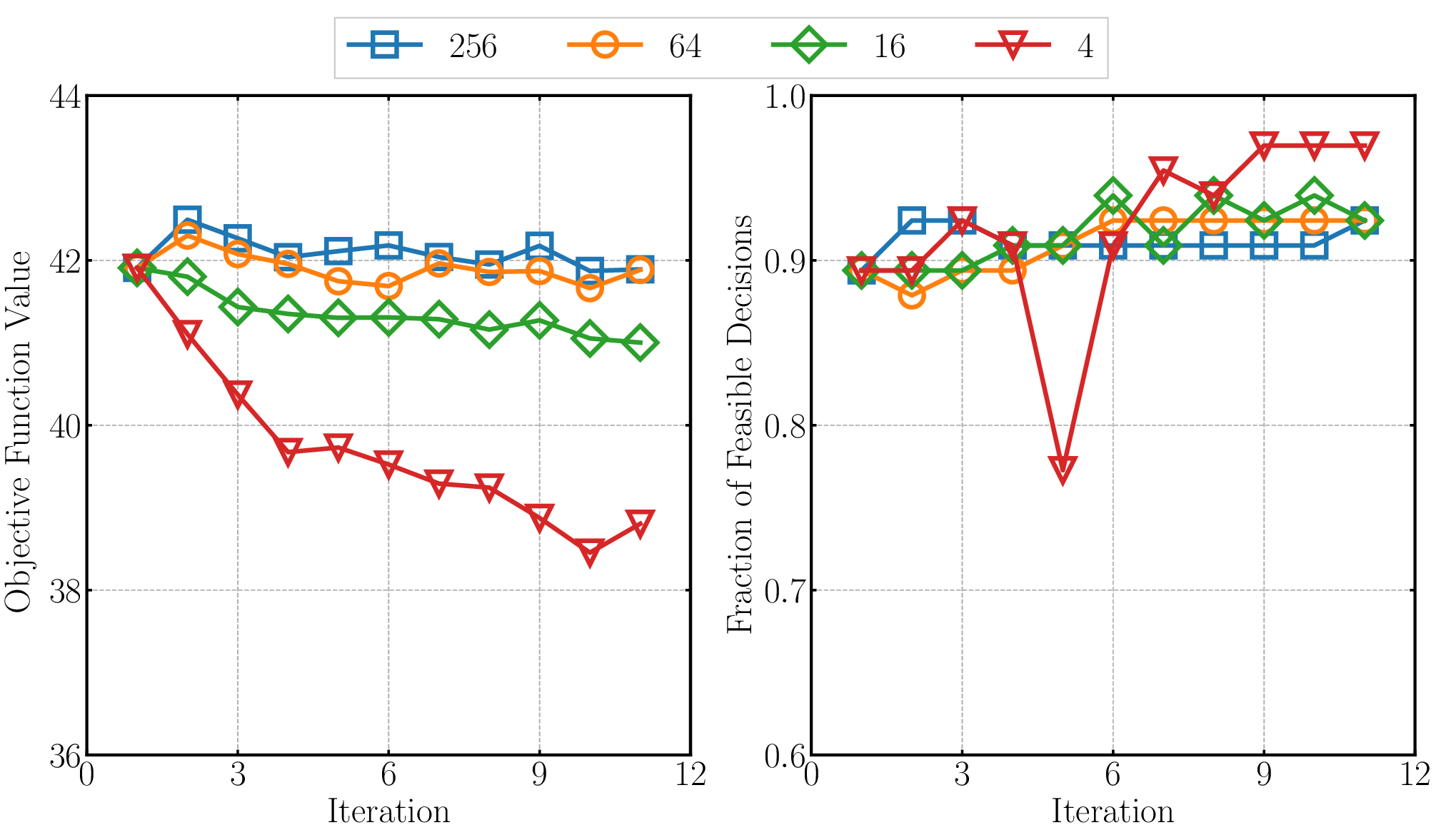}
    \includegraphics[width=0.9\linewidth]{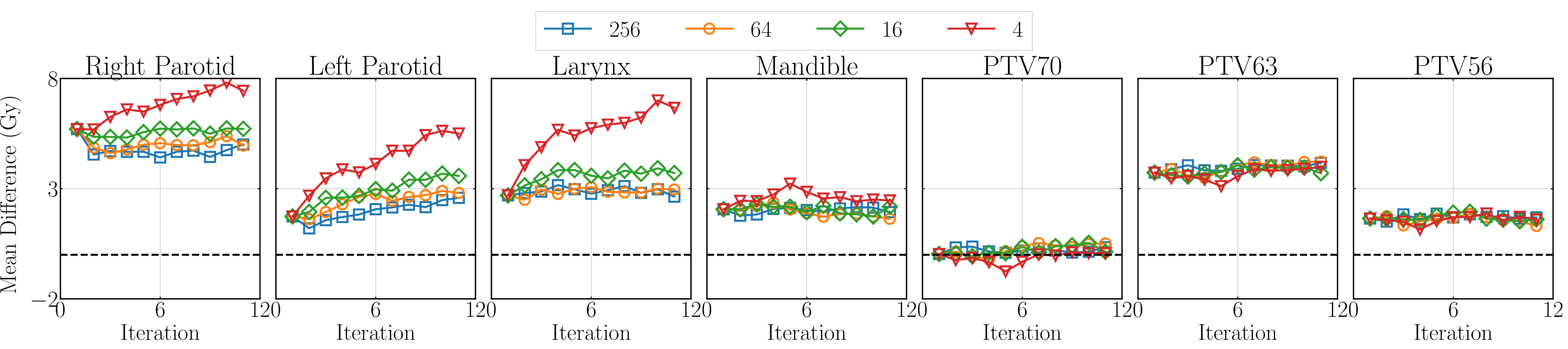}
    \caption{Statistics on the validation set when training on criteria from our institution.
    Top: Objective function and fraction of feasible plans. 
    Bottom: Average difference from the target $\overline{h}_r$. Above 0 Gy (dashed line) suggests plans satisfy the constraints on average.
    }
    \label{fig:best_model}
\end{figure}
\begin{figure}[t]
    \centering
    \includegraphics[width=0.5\linewidth]{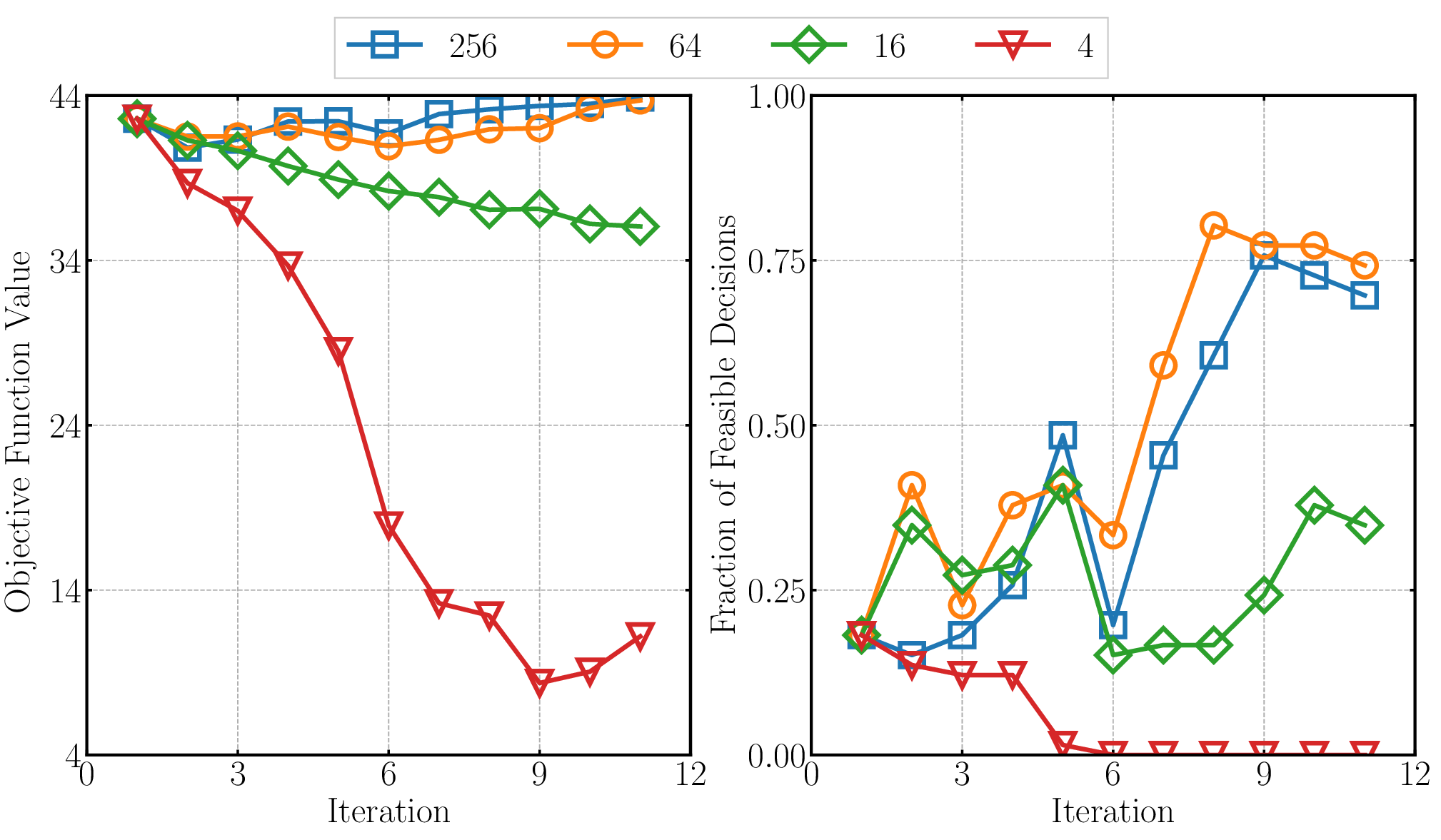}
    \includegraphics[width=0.99\linewidth]{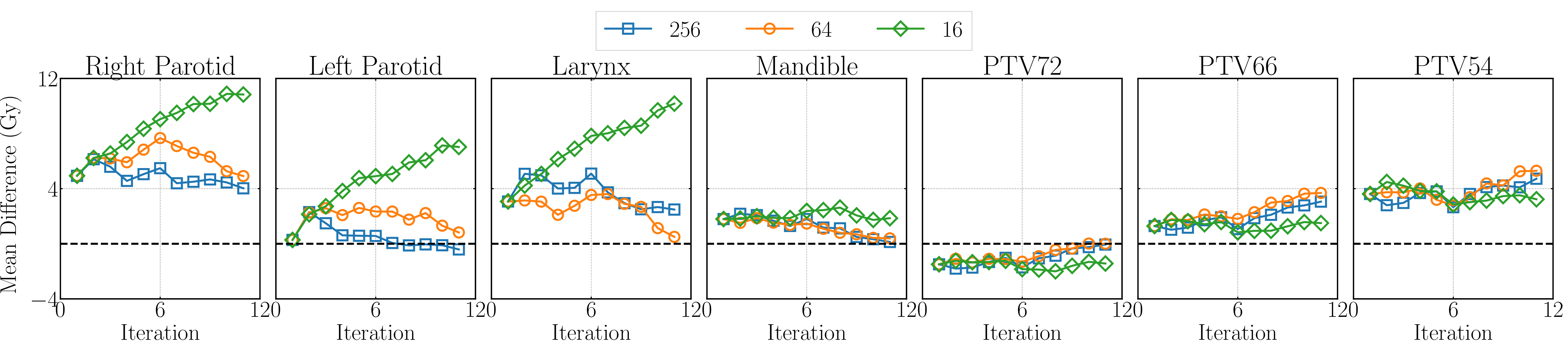}
    \caption{Statistics on the validation set when training on criteria from~\citet{Geretschlager:2015aa}.
    Top: Objective function and fraction of feasible plans.
    Bottom: Average difference from the target $\overline{h}_r$.  Above 0 Gy (dashed line) suggests plans satisfy the constraints on average.
    }
    \label{fig:follow_up}
\end{figure}
%

\subsection{Supplementary results for learning to predict dose distributions}
\label{sec:ec_rt_learning}

Figure~\ref{fig:best_model}(a) displays the average objective function value and the fraction of feasible dose distributions. 
Both the objective function value and constraint improve as a function of the number of iterations. For example, in the first iteration, $89\%$ of predictions in the validation set satisfy all of their clinical constraints, whereas this fraction increases to $97\%$ by iteration $11$. 
This suggests that IPMAN trains the model to generate fewer infeasible doses since the classifier is learning to produce a tighter characterization of the feasible set (see Proposition~\ref{propn:shrinking_optimal_set}).

Figure~\ref{fig:best_model}(b) shows the average difference from the boundary of the constraint $g_r(\bx) \leq \overline{h}_r$. If the difference is positive, doses on average satisfy the clinical criteria. We observe two important phenomena. First, the four leftmost plots are associated with OAR constraints. By minimizing the objective, the associated OAR constraints see progressively better adherence as expected. Note that the Mandible and PTV70 structures often overlap, meaning their constraints conflict with each other, preventing improvement for this organ. Second, the PTV constraints show small but sustained improvement as the number of training iterations increase. This is because they are solely associated with feasibility and are not part of the objective function. In particular, the PTV70 constraint is typically the hardest to satisfy in practice; IPMAN learns this difficulty and makes predictions that lie close to the boundary of the feasible set.

\subsection{Supplementary results for adapting to a different institution}
\label{sec:ec_rt_new_clinic}

Figure~\ref{fig:follow_up}(a) displays the objective function value and the fraction of plans that satisfied $\Xfeas(\bu)$ and $\Xrel$. 
The models trained for $\lambda \leq 16$ decrease in objective function value and constraint satisfaction as the algorithm progresses. In the early stages, the classifier (which is initially trained mainly using the clinical data and doses sampled from the same distribution) has not yet observed a sufficient and diverse number of feasible plans, preventing it from being a sufficient $\delta$-barrier. 

Generally at high $\lambda$, optimal solutions to the barrier problem are less aggressive in terms of minimizing the objective and instead lie well in the interior of the feasible set (see~\ref{sec:ipman_delta_algorithm}). 
As the classifier improves, particularly after iteration 6, the generators for $\lambda\geq 64$ learn to predict doses that satisfy $\Xfeas(\bu)$. In particular, the distance from the PTV72 boundary in Figure~\ref{fig:follow_up}(b) begins to increase from the 6th iteration and passes $0$ by the 10th iteration.
This result demonstrates that our model is learning a new constraint, the PTV72 constraint, which cannot be learned by naively using the training data. Recall that no clinical dose in the original data set reached 72 Gy on the PTV70. Thus, learning the PTV72 constraint is entirely attributable to IPMAN.

\end{document}